\documentclass[twoside,11pt]{article}

\usepackage{pgfplots,float}
\usepackage{placeins}

\pgfplotsset{compat=newest}
\usepackage{amsmath}

\usepackage[preprint]{jmlr2e}
\usepackage{lastpage}
\usepackage{amsmath}
\usepackage{graphicx}
\usepackage{subcaption}
\usepackage{subfiles}
\usepackage{algorithm, algpseudocode}
\usepackage{svg}
\usepackage{booktabs}


\jmlrheading{}{2026}{}{}{}{}{AliAhmad and Shukla}
\ShortHeadings{The \(n\)-Dominant Aligned Subspace}{AliAhmad and Shukla}
\firstpageno{1}

\begin{document}

\title{A Systematic Analysis of a Small Optimal Subspace for\\
Physics-Informed Neural Networks:
\texorpdfstring{\(n\)-DAS%
\thanks{DAS is the Dominant Aligned Subspace, the span of the Hessian
eigenvectors selected by their alignment with the gradient, and \(p\) is the
number of network parameters. We write \(n\)-DAS for an optimizer of order
\(n\) restricted to it: \(1\)-DAS for the first-order methods and
\(2\)-DAS for the second-order ones (Section~\ref{sec:projected-optimizers}).}
\ \(\subset\mathbb{R}^{p}\)}{n-DAS in R^p}}

\author{\name Ahmad AliAhmad \email ahmad\_aliahmad@brown.edu \\
        \addr Division of Applied Mathematics \\
        Brown University \\
        Providence, RI 02912, USA
        \AND
        \name Khemraj Shukla \email khemraj\_shukla@brown.edu \\
        \addr Division of Applied Mathematics \\
        Brown University \\
        Providence, RI 02912, USA
        }

\begin{NoHyper}\maketitle\end{NoHyper}

\begin{abstract}

The search for a low-dimensional subspace that governs neural network training has been a persistent goal in optimization. Yet subspaces defined by Hessian eigenvalue magnitude, such as the dominant/bulk decomposition underlying Dom-SGD and Bulk-SGD, have repeatedly failed to capture where optimization takes place, and adaptive Krylov-based alternatives built on energy-fraction thresholds show promise but still struggle at lower eigenmodes. We argue this failure is structural: the directions of highest curvature are not, in general, the directions along which training moves. We introduce the dominant aligned subspace (DAS), constructed from a full eigendecomposition of the Hessian but selected by gradient alignment rather than eigenvalue magnitude, as a tool for understanding optimization dynamics and a foundation for designing methods that exploit them. To expose this structure under the hardest conditions, we use physics-informed neural networks (PINNs) as a testbed: differential operators from the parabolic, elliptic, and hyperbolic classes induce severely ill-conditioned, anisotropic loss landscapes. Across Poisson, heat, wave, and viscous Burgers benchmarks DAS-restricted training with a line search matches full-space training even where the projected Hessian remains ill-conditioned, showing empirically that convergence does not require all Hessian eigendirections but only a low-dimensional, gradient-aligned subset. We extend the analysis to Gauss--Newton within the DAS and discuss what this structure implies for future subspace-aware optimizers.

\end{abstract}


\section{Introduction}

Neural networks are nonlinear function approximators constructed through compositions of affine transformations and nonlinear activation functions. In supervised, data-driven learning, their parameters are commonly trained by minimizing a mean-squared-error (MSE) objective. Given a dataset of input--target pairs \(\{(x_i,y_i)\}_{i=1}^N\), this objective can be written as
\[
\mathcal{L}(\theta)
=
\frac{1}{2N}\sum_{i=1}^{N}
\|f_\theta(x_i)-y_i\|^2
=
\frac{1}{2N}\|r(\theta)\|^2,
\]
where \(f_\theta\) denotes the neural network, \(\theta\in\mathbb{R}^{p}\) is the parameter vector, and \(r(\theta)\) is the vector of prediction residuals. Consequently, training a neural network with MSE loss is a nonlinear least-squares problem.

Despite this apparently simple formulation, the optimization problem is challenging. The residual map \(r(\theta)\) is highly nonlinear in the parameters, and modern neural networks often operate in extremely high-dimensional parameter spaces. The resulting loss landscape is nonconvex, anisotropic, and typically ill-conditioned, especially for PINNs \citep{wilamowski2001ill,wang2021pinnfail,rathore2024challenges}; its Hessian spectrum typically shows a few large outlier eigenvalues above a broad bulk \citep{sagun2017empirical,ghorbani2019investigation}. Its local geometry may contain directions of very different curvature, nearly flat manifolds associated with parameter redundancies, and sharp directions that can substantially constrain stable optimization\citep{li2018visualizing}. These geometric properties directly affect the optimization trajectory, convergence behavior, sensitivity to hyperparameters, and ultimately the quality of the learned solution.

In practice, neural networks are most often optimized using first-order methods, including gradient descent (GD), stochastic gradient descent (SGD)\citep{robbins1951stochastic}, and adaptive variants such as Adam\citep{kingma2015adam}. Given an iterate \(\theta_t\), the gradient descent method updates the parameters according to
\[
\theta_{t+1}
=
\theta_t-\eta\, g_t,
\]
where \(\eta>0\) is the learning rate and \(g_t\) is either the full gradient \(\nabla \mathcal{L}(\theta_t)\) or a stochastic mini-batch estimate. This iterative process can be interpreted dynamically: the optimizer moves through the loss landscape according to the local gradient field, while the landscape geometry determines which directions are amplified, suppressed, or destabilized by the update. Understanding these dynamics is therefore essential for explaining convergence, stability, implicit regularization, and the practical success of stochastic first-order optimization.

A widely used qualitative tool for investigating these questions is loss-landscape visualization, which examines the variation of the objective along selected low-dimensional parameter-space directions. Such visualizations have motivated a large literature on flatness and sharpness, Hessian spectra, gradient noise, mode connectivity, and the relationship between local curvature and generalization \citep{li2018visualizing,sagun2017empirical,ghorbani2019investigation,cohen2021gradient}. However, interpreting a low-dimensional projection of a high-dimensional optimization trajectory is inherently subtle. A projection may reveal apparent confinement, curvature, or alignment that depends strongly on the chosen subspace, the normalization of directions, the time at which the directions are computed, and the numerical accuracy with which these directions are represented.

A recent study by \citet{song2025does}, \emph{Does SGD Really Happen in Tiny Subspaces?}, investigated whether neural-network training dynamics are effectively confined to a small subspace spanned by leading Hessian eigendirections. Its central conclusion was that training does not, in general, occur solely within the span of a small number of dominant eigenvectors: although the optimization trajectory may exhibit substantial apparent alignment with leading spectral directions, important components of the dynamics remain outside such a low-dimensional subspace. This observation challenges overly simplified accounts in which optimization is governed exclusively by a handful of sharp curvature directions.

\paragraph{Related work.}
The observation that the gradient concentrates in a small number of Hessian eigendirections goes back to \citet{gur2018gradient}, who reported that gradient descent ``happens in a tiny subspace'' spanned by the top-\(k\) eigenvectors. \citet{song2025does} showed that this alignment can be spurious, in that the loss decrease is driven by the bulk complement; \citet{deng2026suspicious} analyze the same alignment through step-size conditions, and \citet{benarous2023spectral} through high-dimensional dynamics. A complementary line of work asks whether networks can be \emph{trained} in a fixed low-dimensional subspace: random subspaces of modest dimension suffice for many tasks \citep{li2018intrinsic,gressmann2020improving}, and so do subspaces spanned by past iterates \citep{li2022low}. Subspace second-order methods include Krylov subspace descent \citep{vinyals2012krylov}, randomized subspace Newton \citep{gower2019rsn}, practical Gauss--Newton for deep learning \citep{botev2017practical}, and the subspace Levenberg--Marquardt methods of \citet{hoang2026subspace,hoang2026adaptive}, whose convergence framework we adapt in Appendix~\ref{app:convergence}; the conditioning of the Gauss--Newton matrix in neural networks is characterized by \citet{zhao2024theoretical}. The edge-of-stability phenomenon \citep{cohen2021gradient,cohen2022adaptive} and analyses of its mechanism \citep{arora2022understanding,damian2023self,song2024edge} underlie the spurious-alignment collapse we observe; recent optimizers that treat sharp and flat directions differently \citep{zakarin2025accelerating,nicolas2025dome} and the analysis of mini-batch noise through dominant-subspace fluctuations by \citet{so2026minibatch} are closely related to the selection question studied here. On the PINN side, the difficulty of training \citep{raissi2019physics,karniadakis2021physics,wang2021pinnfail,rathore2024challenges} has motivated curvature-aware optimizers \citep{muller2023achieving,jnini2026curvature} and systematic optimizer comparisons \citep{wang2023expert,kiyani2025which}.

In this work, we revisit this conclusion through the lens of time-varying subspaces selected by an energy threshold. In particular, we argue that while optimization may not be concentrated in the top-curvature eigenvectors, the full bulk subspace is not needed for convergence either. We therefore rank Hessian eigenvectors by the fraction of gradient energy they carry.

Our analysis therefore distinguishes between aligned and non-aligned eigenvectors according to where the gradient at the current iterate lives within the full space. By re-examining the evidence with gradient-alignment diagnostics, we clarify which conclusions about training in low-dimensional subspaces are supported by the dynamics. In doing so we observe empirically that convergence does not require all Hessian eigendirections: a small, time-varying, gradient-aligned subset suffices.

\section{Motivation}

These findings motivate several directions explored in this work: (1) empirical evidence that first-order and Gauss--Newton optimizers do not need all Hessian eigendirections, but only a small, time-varying, gradient-aligned subset; (2) the formal definition of the dominant aligned subspace (DAS); (3) the observation that restricting updates to the DAS yields a condition number \(\kappa_{\mathcal{A}}(H)\) that is never larger than the full condition number \(\kappa_2(H)\) --- an improvement in conditioning \emph{relative} to the full space, even when \(\kappa_{\mathcal{A}}(H)\) remains large in absolute terms; (4) the development and evaluation of Gauss--Newton methods restricted to the DAS, enabling curvature-aware updates in low-dimensional subspaces; and (5) an investigation of how the spurious-alignment phenomenon can be overcome with line-search methods and with energy thresholds very close to one.

\section{Mathematical Preliminaries}
\label{sec:preliminaries}

\subsection{Edge of Stability}
\label{sec:eos}

Consider gradient descent applied to a twice-differentiable loss
\(\mathcal{L}:\mathbb{R}^p\to\mathbb{R}\):
\[
\theta_{t+1}
=
\theta_t
-
\eta \nabla \mathcal{L}(\theta_t),
\]
where \(\eta>0\) is the learning rate. Around a local minimizer, a quadratic
approximation of the loss has Hessian
\[
H(\theta)
=
\nabla^2\mathcal{L}(\theta).
\]
If \(v_i\) is an eigenvector of \(H(\theta)\) with eigenvalue
\(\lambda_i\), then the local gradient-descent dynamics along \(v_i\) are
governed by
\[
z_{t+1}
=
\left(1-\eta\lambda_i\right)z_t.
\]
Thus, contraction along this eigendirection requires
\[
\left|1-\eta\lambda_i\right|<1,
\]
or equivalently,
\[
0<\eta\lambda_i<2.
\]
For a positive-semidefinite Hessian, stability across all eigendirections
therefore requires
\begin{equation}
\lambda_{\max}\bigl(H(\theta)\bigr)
<
\frac{2}{\eta}.
\label{eq:classical-stability}
\end{equation}
The maximum Hessian eigenvalue
\(\lambda_{\max}(H)\) is commonly called the \emph{sharpness} of the loss
landscape.

\begin{figure}
    \centering
    \includegraphics{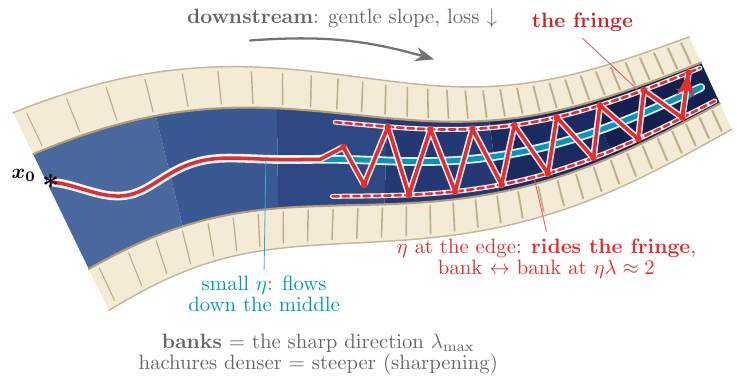}
    \caption{\textbf{The edge of stability, as a river (an illustration of the
idea).} The loss is a descending river: the many flat directions run
downstream (gentle slope --- the channel darkens as the loss falls), the
sharp direction $\lambda_{\max}$ is across the channel, and the banks
steepen as training progresses (denser hachures). A small learning rate
settles to the centre and glides. A large one glides too, until the banks
are steep enough that $\eta\lambda\approx2$: from there every step
overshoots the centre and lands on the opposite bank, so the trajectory
stops sinking and \emph{rides the fringe} --- slaloming bank-to-bank along
the dashed rails at the largest swing the walls can hold, while still
drifting downstream. Purely schematic.
}
    \label{fig:eos}
\end{figure}

Empirical studies have shown that neural-network training often operates near
the boundary in \eqref{eq:classical-stability}, a phenomenon known as the
\emph{edge of stability} (EoS) \citep{cohen2021gradient} (Figure~\ref{fig:eos}); see \citet{cohen2022adaptive} for adaptive methods and \citet{arora2022understanding,damian2023self,song2024edge} for analyses of the mechanism. In this regime,
the sharpness can fluctuate near, and sometimes temporarily exceed,
\(2/\eta\). Consequently, the training loss may increase over individual
iterations or short time intervals while still decreasing over longer
optimization horizons. This behavior is not captured by the classical local
quadratic analysis, which predicts monotone contraction only when the
stability condition holds at every step.

\subsection{Gradient--Hessian Alignment}
\label{sec:alignment}

Let
\[
H(\theta)
=
\nabla^2\mathcal{L}(\theta)
=
V\Lambda V^\top,
\qquad
V=[v_1,\ldots,v_p],
\]
where \(V\) is orthogonal and the eigenvalues are ordered as
\[
\lambda_1\geq\lambda_2\geq\cdots\geq\lambda_p.
\]
For the top-\(k\) Hessian eigenspace, define
\[
V_k
=
[v_1,\ldots,v_k].
\]
The alignment of the full gradient with this subspace is
\begin{equation}
\mathcal{A}_k(\theta)
=
\frac{
    \left\lVert
        V_k^\top\nabla\mathcal{L}(\theta)
    \right\rVert_2^2
}{
    \left\lVert
        \nabla\mathcal{L}(\theta)
    \right\rVert_2^2
}.
\label{eq:topk-alignment}
\end{equation}
The quantity \(\mathcal{A}_k(\theta)\in[0,1]\) measures the fraction of
gradient energy contained in the span of the \(k\) largest-curvature Hessian
eigendirections; it is the dominant-subspace alignment studied by
\citet{gur2018gradient} and \citet{song2025does}.

Our approach uses a directional form of alignment. Let \(g\) denote the
\emph{selecting gradient}, i.e., the gradient used to rank eigendirections:
the full-batch gradient \(\nabla\mathcal{L}_{\mathrm{full}}(\theta)\) in
the full-batch experiments of Section~\ref{sec:full-batch-optimization}, and
either the mini-batch gradient \(\nabla\mathcal{L}_{\mathrm{batch}}(\theta)\)
or the full-batch gradient in the mini-batch experiments of
Section~\ref{sec:mini-batch-optimization}. The Hessian eigendecomposition is
computed from the full-batch loss in the full-batch experiments and from the
loss of the current mini-batch in the mini-batch experiments, where the
Gauss--Newton runs use the minibatch Jacobian instead. For each eigenvector \(v_i\), define
\begin{equation}
a_i
=
\frac{
    \left(g^\top v_i\right)^2
}{
    \lVert g\rVert_2^2
},
\qquad
i=1,\ldots,p.
\label{eq:directional-energy}
\end{equation}
Because the Hessian eigenvectors are orthonormal,
\begin{equation}
\sum_{i=1}^{p}a_i
=
1.
\label{eq:directional-energy-partition}
\end{equation}
Hence, \(a_i\) is the fraction of the energy of \(g\) in the Hessian
eigendirection \(v_i\); we refer to the \(a_i\) as gradient-energy scores.

Song et al.\ \citep{song2025does} analyze stochastic optimization using a
spectral split between a dominant Hessian eigenspace and its bulk
complement. Let
\[
P_k(\theta_t)
=
\sum_{i=1}^{k}
v_i(\theta_t)v_i(\theta_t)^\top
\]
be the projector onto the span of the \(k\) leading Hessian eigenvectors at
\(\theta_t\), and define
\[
P_k^\perp(\theta_t)
=
I-P_k(\theta_t).
\]
For stochastic gradient \(g_t\) and learning rate \(\eta\), the dominant and
bulk updates are
\begin{align}
\theta_{t+1}^{\mathrm{Dom}}
&=
\theta_t
-
\eta P_k(\theta_t)g_t,
\label{eq:dom-sgd}
\\
\theta_{t+1}^{\mathrm{Bulk}}
&=
\theta_t
-
\eta P_k^\perp(\theta_t)g_t.
\label{eq:bulk-sgd}
\end{align}
The dominant/bulk decomposition ranks eigendirections by curvature magnitude.
It therefore implicitly assumes that the eigenvectors with the largest
eigenvalues are the directions that account for the optimization-relevant
components of the gradient.

This assumption can be unreliable in highly ill-conditioned landscapes, in
the presence of near-degenerate eigenvalues, or when numerical eigenvectors
are sensitive to perturbations. We therefore define the \emph{Dominant
Aligned Subspace} (DAS) by ranking Hessian eigenvectors according to their
gradient-energy scores \(a_i\), rather than their associated eigenvalues.

Let \(\pi\) be a permutation of \(\{1,\ldots,p\}\) such that
\[
a_{\pi(1)}
\geq
a_{\pi(2)}
\geq
\cdots
\geq
a_{\pi(p)}.
\]
For a prescribed \emph{energy threshold} \(\tau\in(0,1]\) --- the analogue
of an explained-variance threshold in principal component analysis
\citep{pearson1901lines,hotelling1933analysis} --- the \emph{DAS dimension}
is
\begin{equation}
k
=
\min
\left\{
k'\in\{1,\ldots,p\}
:
\sum_{j=1}^{k'}a_{\pi(j)}
\geq
\tau
\right\},
\label{eq:das-rank}
\end{equation}
and the DAS index set is \(\mathcal{A}=\{\pi(1),\ldots,\pi(k)\}\). The DAS
basis and associated orthogonal projector are
\begin{equation}
V(\theta)
=
\begin{bmatrix}
v_{\pi(1)} & \cdots & v_{\pi(k)}
\end{bmatrix},
\qquad
P(\theta)
=
V(\theta)V(\theta)^\top.
\label{eq:das-projector}
\end{equation}
Along an optimization trajectory \(\{\theta_t\}\) we write
\(V_t=V(\theta_t)\), \(P_t=P(\theta_t)\), \(\mathcal{A}_t\), and
\(k_t\). By construction,
\begin{equation}
\left\lVert
P(\theta)g
\right\rVert_2^2
=
\sum_{j=1}^{k}
\left(g^\top v_{\pi(j)}\right)^2
\geq
\tau\lVert g\rVert_2^2,
\label{eq:das-energy}
\end{equation}
and we call \(\lVert P(\theta)g\rVert_2^2/\lVert g\rVert_2^2\in[\tau,1]\)
the \emph{captured energy fraction}. Thus, the DAS is the smallest subspace
formed from Hessian eigendirections that retains at least a
\(\tau\)-fraction of the energy of the selecting gradient. In contrast to a
fixed top-\(k\) eigenspace, both its dimension and its spectral composition
adapt to the gradient at the current iterate. On a quadratic loss with a
fixed eigenbasis, restricting gradient descent to the DAS is block
coordinate descent in eigencoordinates with the greedy Gauss--Southwell
selection rule \citep{nutini2015coordinate,nutini2022let}; see
Remark~\ref{rem:gs}.

\subsection{Conditioning of the Loss Landscape}
\label{sec:conditioning}

The Hessian spectrum determines the local anisotropy of the loss landscape.
For a positive-definite Hessian, the spectral condition number is
\begin{equation}
\kappa_2\bigl(H(\theta)\bigr)
=
\frac{\lambda_{\max}\bigl(H(\theta)\bigr)}{\lambda_{\min}\bigl(H(\theta)\bigr)}.
\label{eq:condition-number}
\end{equation}
When \(\kappa_2(H)\approx 1\), the local quadratic approximation has nearly
spherical level sets; large \(\kappa_2(H)\) corresponds to elongated level
sets, steep along large-eigenvalue directions and nearly flat along small
ones. Hessians of modern neural networks typically exhibit a few large
outlier eigenvalues above a broad bulk of much smaller ones
\citep{sagun2017empirical,ghorbani2019investigation}, producing
highly anisotropic geometry and large condition numbers, particularly for
PINNs \citep{wang2021pinnfail,rathore2024challenges}. When \(H\) is indefinite or
singular, \eqref{eq:condition-number} is replaced by
\begin{equation}
\kappa_2(H)
=
\frac{\sigma_{\max}(H)}{\sigma_{\min}^{+}(H)},
\label{eq:singular-condition-number}
\end{equation}
where \(\sigma_{\min}^{+}(H)\) is the smallest nonzero singular value; for
symmetric \(H\), singular values equal absolute eigenvalues.

Ill-conditioning slows first-order optimization because a single learning rate
must serve both steep and flat directions: a rate stable in sharp
directions makes slow progress in flat ones, while a rate suited to flat
directions is unstable in sharp ones. This motivates preconditioning and
projected curvature-aware optimization. Yet \(\kappa_2(H)\) describes the
entire parameter space, including directions the optimizer may never
traverse. If training is confined to a low-dimensional subspace, the
relevant conditioning is that of \(H\) restricted to it.

\subsection{Numerical Instability}

Ill-conditioning also limits how accurately the curvature information itself
can be computed. A backward-stable eigensolver returns the exact spectrum of a
perturbed matrix \(H+E\) with \(\|E\|_2=\mathcal{O}\bigl(u\,\|H\|_2\bigr)\),
where \(u\) is the unit roundoff, so every eigenvalue carries an absolute error
of that order and eigenvalues with \(|\lambda_i|\lesssim u\,\|H\|_2\) are
indistinguishable from rounding noise
\citep{golub2013matrix,higham2002accuracy}; errors made in assembling the dense
Hessian only raise this level. The associated eigenvectors are affected even
more strongly, because their sensitivity scales inversely with the gap to
neighboring eigenvalues \citep{davis1970rotation,stewart1990matrix}; within a
cluster narrower than the noise level the computed vectors form an essentially
arbitrary basis of the cluster's invariant subspace. The singular value
decomposition of a Jacobian \(J\) obeys the same bounds on the scale of the
singular values: each \(\sigma_i\) is computed to within
\(\mathcal{O}\bigl(u\,\sigma_{\max}(J)\bigr)\), so the eigenvalues
\(\sigma_i^2\) of \(J^\top J\) are resolved down to
\(\mathcal{O}(u^2)\,\lambda_{\max}(J^\top J)\) and the null space is returned
exactly. Forming the Gram matrix \(J^\top J\) and passing it to an eigensolver
forfeits this: since \(\kappa_2(J^\top J)=\kappa_2(J)^2\), singular values
below \(\sqrt{u}\,\sigma_{\max}(J)\) are lost \citep{golub2013matrix}.

A selection of eigenvectors must therefore certify two things: that the
eigenvalue is distinguishable from zero, and that the eigenvector is well
determined. In double precision (\(u\approx 1.1\times 10^{-16}\)) we use the
tolerance \(\delta=800\,u\approx 9\times 10^{-14}\), one hundred times the
roundoff level \(8u\) assumed for the eigensolver. When the dense Hessian is
assembled, an eigenpair is retained only if
\(|\lambda_i|>\max(\delta,\tau)\,\|H\|_2\) and \(\lambda_i\) is separated from
both neighboring eigenvalues by more than \(\delta\,\|H\|_2\), which bounds the
rotation of the computed eigenvector by roughly \(10^{-2}\) radians. When the
eigenvectors are the right singular vectors of the Jacobian, \(J^\top J\) is
never formed and the same two tests are applied to the singular values:
\(\sigma_i>\max(\delta,\sqrt{\tau})\,\sigma_{\max}(J)\), with neighboring
singular values separated by more than \(\delta\,\sigma_{\max}(J)\). The floor
then corresponds to \(\sigma_i^2>\tau\,\lambda_{\max}(J^\top J)\). We use
\(\tau=10^{-14}\) in both cases; for the dense Hessian the floor is therefore
set by \(\delta\), whereas for the Jacobian it retains singular values down to
\(10^{-7}\,\sigma_{\max}(J)\). Directions that fail either test never enter the
subspace.

\subsection{DAS Condition Number}

Let \(V(\theta)\in\mathbb{R}^{p\times k}\) be the DAS basis of
\eqref{eq:das-projector}, whose columns are the \(k\) eigenvectors of
\(H=\nabla^2\mathcal{L}(\theta)\) selected by gradient alignment. The
projected Hessian is
\begin{equation}
H_{\mathcal{A}} = V(\theta)^\top H V(\theta),
\label{eq:projected-hessian}
\end{equation}
and the \emph{DAS condition number} is
\begin{equation}
\kappa_{\mathcal{A}}(H)
=
\frac{\lambda_{\max}(H_{\mathcal{A}})}{\lambda_{\min}(H_{\mathcal{A}})},
\label{eq:das-condition-number}
\end{equation}
with the singular-value ratio of \eqref{eq:singular-condition-number}
substituted when \(H_{\mathcal{A}}\) is not positive definite.

Since the columns of \(V(\theta)\) are orthonormal eigenvectors of
\(H\), \(H_{\mathcal{A}}=\operatorname{diag}(\lambda_i)_{i\in\mathcal{A}}\)
and \eqref{eq:das-condition-number} reduces to the ratio of the largest to
smallest selected eigenvalue: \(\kappa_{\mathcal{A}}(H)\) depends only on
which eigenvalues the alignment criterion admits, not on their magnitude
rank. For positive-definite \(H\), the Rayleigh bounds
\(\lambda_{\min}(H)\le\lambda_{\min}(H_{\mathcal{A}})\le
\lambda_{\max}(H_{\mathcal{A}})\le\lambda_{\max}(H)\) give
\(\kappa_{\mathcal{A}}(H)\le\kappa_2(H)\); restriction cannot worsen
conditioning, but need not improve it substantially. This is under the assumption that the model takes the same trajectory as the full space model which is not always the case. Thus
\(\kappa_{\mathcal{A}}(H)\) reflects only the geometry seen by the optimizer
within the DAS. Along a trajectory we write \(\kappa_{\mathcal{A}}(H_t)\).
In practice the selected eigenvalues may be indefinite or close to zero; we
therefore always evaluate \eqref{eq:das-condition-number} with absolute values
of the selected eigenvalues and replace the denominator by a numerical
cutoff when it falls below a prescribed tolerance.

\subsection{DAS-Restricted Gauss--Newton Step}
\label{sec:projected-optimizers}

The DAS-restricted methods below restrict their search directions to
\(\operatorname{range}(P_t)\), while evaluating the loss and the acceptance
criteria in the full space \(\mathbb{R}^p\). The first-order instance,
DAS-GD, simply replaces \(g_t\) by \(P_tg_t\) in the gradient-descent
update; it is stated in Section~\ref{sec:full-batch-first-order}. Here we
state the Gauss--Newton instance, DAS-GN. We write \(n\)-DAS for a
method of order \(n\) restricted to the DAS, where \(n\) is the order of
the update rule and not of the information used to build the subspace:
\(1\)-DAS denotes the first-order methods (DAS-GD, DAS-SGD), which still
use the Hessian to select their directions, and \(2\)-DAS the
second-order ones (DAS-Newton, DAS-GN, DAS-LM). Gauss Newton is a quasi second order optimizer but we can say second as an approximation. The subspace dimension is
written \(k_t\) throughout, so \(n\) never denotes it.

For a nonlinear least-squares objective
\[
\mathcal{L}(\theta)
=
\frac{1}{2}
\lVert r(\theta)\rVert_2^2,
\]
let \(J_t=\partial r(\theta_t)/\partial\theta\) be the residual Jacobian, so
that \(g_t=J_t^\top r(\theta_t)\) and \(J_t^\top J_t\) is the Gauss--Newton
curvature approximation \citep{botev2017practical,zhao2024theoretical}. Rather than computing a full-space Gauss--Newton
step and projecting it afterwards, we formulate the Gauss--Newton problem
directly in DAS coordinates. Writing
\[
\theta
=
\theta_t+V_ty,
\qquad
y\in\mathbb{R}^{k_t},
\]
the reduced Jacobian is \(J_tV_t\), and the damped DAS-restricted
Gauss--Newton step solves
\begin{equation}
\left(
V_t^\top J_t^\top J_t V_t
+
\mu_t I
\right)
y_t
=
-V_t^\top g_t,
\qquad
p_t
=
V_ty_t,
\label{eq:das-gauss-newton}
\end{equation}
where \(\mu_t\geq0\) is a Levenberg damping parameter
\citep[Chapter~10]{nocedal2006numerical}. When \(\mu_t=0\) and
\(V_t^\top J_t^\top J_tV_t\) is nonsingular, \eqref{eq:das-gauss-newton}
coincides with the projected pseudoinverse step
\(p_t=-P_t(P_tJ_t^\top J_tP_t)^\dagger P_tJ_t^\top r(\theta_t)\). The
resulting update is
\begin{equation}
\theta_{t+1}
=
\theta_t+\alpha_tp_t,
\label{eq:projected-gauss-newton-update}
\end{equation}
with the step length \(\alpha_t\) chosen by one of the globalization rules
below. The reduced system in \eqref{eq:das-gauss-newton} has dimension
\(k_t\times k_t\) rather than \(p\times p\), and its matrix
\((J_tV_t)^\top(J_tV_t)\) can be formed with \(k_t\) Jacobian--vector
products without materializing a dense \(p\times p\) Gauss--Newton matrix.

\subsection{Globalization in the Full Space}
\label{sec:globalization}

Convergence guarantees for these DAS-restricted methods are given in
Appendix~\ref{app:convergence}.

Although DAS-restricted methods restrict their directions to
\(\operatorname{range}(P_t)\), the full loss
\(\mathcal{L}(\theta)\) determines whether a proposed update is accepted.
Let \(p_t\) be a descent direction satisfying
\[
\nabla\mathcal{L}(\theta_t)^\top p_t < 0,
\]
and let \(\alpha_t>0\) denote the step length; we use the following
standard rules \citep{nocedal2006numerical}.

\paragraph{Wolfe line search.}
A step length \(\alpha_t>0\) satisfies the Wolfe conditions
\citep{wolfe1969convergence} if
\begin{align}
\mathcal{L}(\theta_t+\alpha_tp_t)
&\leq
\mathcal{L}(\theta_t)
+
c_1\alpha_t
\nabla\mathcal{L}(\theta_t)^\top p_t,
\label{eq:wolfe-armijo}
\\
\nabla\mathcal{L}(\theta_t+\alpha_tp_t)^\top p_t
&\geq
c_2
\nabla\mathcal{L}(\theta_t)^\top p_t,
\label{eq:wolfe-curvature}
\end{align}
where
\[
0<c_1<c_2<1.
\]
The first condition enforces sufficient decrease, while the second prevents
the accepted step from being excessively short.

\paragraph{Backtracking line search.}
Given an initial trial step length \(\alpha_t^{(0)}>0\), a contraction factor
\(\beta\in(0,1)\), and an Armijo parameter \(c_1\in(0,1)\)
\citep{armijo1966minimization}, backtracking
repeatedly replaces
\[
\alpha_t
\gets
\beta\alpha_t
\]
until
\begin{equation}
\mathcal{L}(\theta_t+\alpha_tp_t)
\leq
\mathcal{L}(\theta_t)
+
c_1\alpha_t
\nabla\mathcal{L}(\theta_t)^\top p_t.
\label{eq:backtracking-armijo}
\end{equation}

\section{Experiments}
\label{sec:experiments}

\subsection{Setup}
\label{sec:setup}

\paragraph{Benchmarks.}
Table~\ref{tab:collocation} lists the four PDE benchmarks, Poisson-2D,
heat-1D, wave-1D, and Burgers-1D, in the standard PINN formulation of
\citet{raissi2019physics}. Burgers-1D uses a larger network, with more
parameters, than the other three (Table~\ref{tab:collocation}) and is
referred to below as the scaled Burgers benchmark. Poisson-2D, heat-1D,
and wave-1D have closed-form
solutions; for Burgers-1D, with \(u(x,0)=-\sin(\pi x)\) and zero Dirichlet
data, the reference solution is the Cole--Hopf integral evaluated by Gauss--Hermite quadrature and
used for error measurement only. Each wave-1D initial point contributes two
residuals, for \(u(x,0)\) and \(u_t(x,0)\). Collocation points are drawn
uniformly at random once and held fixed; boundary and initial points are
never mini-batched.

All benchmarks use soft constraints, i.e., the unweighted sum of the
mean-squared interior residual, boundary mismatch, and initial mismatch,
except the scaled Burgers problem, which uses the hard-constraint ansatz
\[
u_\theta(x,t)
=
-\sin(\pi x)+t\,(1-x^{2})\,\mathcal{N}_\theta(x,t),
\]
so that its objective is the interior residual alone.

\begin{table}[!htbp]
\centering
\caption{
PDE benchmarks. Domains are the unit square for Poisson-2D,
\([0,1]\times[0,1]\) in \((x,t)\) for heat-1D and wave-1D, and
\([-1,1]\times[0,1]\) for Burgers-1D; \(\alpha=0.1\), \(c=1\),
\(\nu=0.01/\pi\).
}
\label{tab:collocation}
\small
\setlength{\tabcolsep}{5pt}
\begin{tabular}{@{}llcrrrr@{}}
\toprule
Benchmark & PDE & Exact solution & \(p\) & Int. & Bdry & Init \\
\midrule
Poisson-2D
& \(-\Delta u = 2\pi^{2}\sin\pi x\,\sin\pi y\)
& \(\sin\pi x\,\sin\pi y\)
& 1981 & 400 & 100  & --- \\
heat-1D
& \(u_t=\alpha u_{xx}\)
& \(\sin\pi x\;e^{-\pi^{2}\alpha t}\)
& 1981 & 80 & 24 & 24 \\
wave-1D
& \(u_{tt}=c^{2}u_{xx}\)
& \(\sin\pi x\,\cos\pi c t\)
& 1981 & 100 & 24 & 24 \\
Burgers-1D
& \(u_t+uu_x=\nu u_{xx}\)
& Cole--Hopf
& 6391 & 1500 & --- & --- \\
\bottomrule
\end{tabular}
\end{table}

\paragraph{Architecture.}
All networks are \(\tanh\) multilayer perceptrons in double precision.
Poisson-2D, heat-1D, and wave-1D use \(2\)--\(30\)--\(30\)--\(30\)--\(1\),
\(p=1981\); the scaled Burgers benchmark uses four hidden layers of width
45, \(p=6391\). DAS
dimensions should be read against these values of \(p\). The only exception
is the spurious-alignment study of Appendix~\ref{app:spurious}, which uses a
smaller \(2\)--\(20\)--\(20\)--\(20\)--\(1\) network with \(p=921\).

\paragraph{Curvature matrix.}
First-order experiments build the DAS from the eigenvectors of the dense
Hessian \(\nabla^{2}\mathcal{L}\); Gauss--Newton experiments from those of
\(J^{\top}J\), obtained as the right singular vectors of the Jacobian \(J\)
of the residuals at the collocation points of the current batch (all of
them at full batch), unless the basis is stated to come from
\(\nabla^{2}\mathcal{L}\). The basis and the step then use the same
Jacobian, so the reduced system \eqref{eq:das-gauss-newton} is diagonal,
also under mini-batching (Section~\ref{sec:mini-batch-gauss-newton}). The basis is
recomputed at every iteration. The line-search Gauss--Newton runs use the
fixed damping \(\mu_t=10^{-6}\); the Levenberg--Marquardt runs adapt it.

\paragraph{Energy threshold.}
First-order experiments use \(\tau=0.95\) or \(\tau=0.99\), as stated; the
Gauss--Newton experiments of Sections~\ref{sec:full-batch-gauss-newton}
and~\ref{sec:mini-batch-gauss-newton} use \(\tau=0.999\) and
\(\tau=0.9999\). The DAS dimension \(k_t\) is reported as the median and
\(10\)--\(90\%\) range, or as the median and mean, over the iterations
indicated.

\paragraph{Globalization.}
Fixed-step first-order runs use \(\eta=0.01\). Line searches start from
\(\alpha_t^{(0)}=1\): backtracking is Armijo halving with \(c_1=10^{-4}\)
and at most 30 trials; Wolfe is the strong bracket-and-zoom search of
\citet[Alg.~3.5--3.6]{nocedal2006numerical} with \(c_1=10^{-4}\),
\(c_2=0.9\), doubling to bracket and bisecting to zoom. Both are evaluated
on the full objective, so that an accepted step certifies a decrease of the
loss actually being minimized. Slopes can be measured with the projected
gradient \(P_tg_t\), from which the direction is built, or with the full
gradient \(g_t\), and the choice changes nothing: every
direction we use is confined to the DAS,
\(p_t\in\operatorname{range}(P_t)\), and for such a direction
\(g_t^\top p_t=(P_tg_t)^\top p_t\)
\eqref{eq:full-and-projected-derivative}, at \(\theta_t\) and at every
trial point. The sufficient-decrease and curvature tests therefore return
the same step length with either gradient. The identity does not hold for
methods whose direction leaves the subspace, such as momentum, Adam, or
quasi-Newton updates; none is used here. Under mini-batching, trial steps
are judged on the full-batch loss for the same reason: the direction comes
from a mini-batch gradient, and accepting it only when the full-batch loss
decreases keeps the training objective from increasing. The slope in the
sufficient-decrease test is then that of the mini-batch gradient along the
direction, and the curvature test at a trial point uses the full-batch
gradient there; in the mini-batch first-order runs of
Section~\ref{sec:mini-batch-first-order} both are measured with the
projected gradient. An exhausted
search then skips the step (\(\alpha_t=0\)); at full batch it takes the
nominal step.
The mini-batch Gauss--Newton experiments of
Section~\ref{sec:mini-batch-gauss-newton} are the exception: there the
conditions are evaluated on the mini-batch objective with the projected
gradient, and an exhausted search takes the nominal step.
Levenberg--Marquardt (LM) runs take the Gauss--Newton step with unit step
length and an adaptive damping \(\mu_t=c_t\lVert r_t\rVert_2^2\), bounded
below by \(10^{-12}\) times the squared Frobenius norm of the reduced
Jacobian. The first damping value is \(10^{-3}\) times the largest diagonal
entry of the reduced Gauss--Newton matrix. A trial step is accepted when the
gain ratio \eqref{eq:gain-ratio} exceeds \(10^{-4}\), after which \(c_t\) is
multiplied by \(\max\{1/3,\,1-(2\rho_t-1)^3\}\); otherwise \(c_t\) is
multiplied by a factor that starts at \(2\) and doubles with each rejection,
and the system is solved again with the same Jacobian, at most 30 times.
The update of \(c_t\) follows the gain-ratio rule of
\citet{nielsen1999damping}; see also \citet{madsen2004methods}.
Trust-region globalization is left for future work
\citep{hoang2026adaptive}.

\paragraph{Run lengths.}
\(10{,}000\) iterations and a single seed throughout, except the
\(20{,}000\)-iteration runs of Appendix~\ref{app:spurious}.

\paragraph{Metrics.}
The \emph{training loss} is the full-batch objective on the collocation set
of Table~\ref{tab:collocation}; the \emph{test loss} is the same objective
on an independent fixed set with four times as many points of each type.
Solution errors are relative \(L^{2}\), \(L^{\infty}\), \(L^{1}\), and
\(H^{1}\) errors against the reference solution on a uniform grid
(\(201^{2}\) Poisson-2D, \(401^{2}\) heat-1D and wave-1D, \(201^{2}\) scaled
Burgers-1D),
with the \(H^{1}\) seminorm by central differences; for soft-constraint
benchmarks we also report the root-mean-square boundary and initial
mismatch. Where stated, metrics are given as the ratio of the
DAS-restricted run to the full-space run with the same globalization. The
horizontal axis of every training curve is the iteration \(t\).
The relative \(L^{2}\) error is the primary criterion throughout: a PINN is
trained to recover the solution, and we call a restricted run successful
when it matches the solution error of the full-space run. Training and test
loss are reported as optimization diagnostics.

\paragraph{Legend key.}
``full'' is the unrestricted method; ``DAS-GD'', ``DAS-SGD'', ``DAS-GN''
the DAS-restricted methods; ``fixed step'', ``BT'', ``Wolfe'', ``LM'' the
globalization; ``basis \(J^{\top}J\)'' or ``basis
\(\nabla^{2}\mathcal{L}\)'' the curvature matrix; ``projB'' and ``projF''
the batch- and full-gradient-selected DAS of
Section~\ref{sec:mini-batch-optimization}. A trailing \(\tau\) overrides
the default above.

\subsection{Full-Batch Optimization}
\label{sec:full-batch-optimization}

We first study the DAS in a deterministic full-batch setting. This isolates
the effect of restricting the update to a gradient-aligned Hessian
eigenspace from the additional variability caused by stochastic gradients and
mini-batch sampling. The stochastic setting is considered separately in
Section~\ref{sec:mini-batch-optimization}.

At iteration \(t\), both the Hessian and the selecting gradient used to
construct the DAS are computed from the full training objective:
\[
H_t
=
\nabla^2\mathcal{L}_{\mathrm{full}}(\theta_t),
\qquad
g_t
=
\nabla\mathcal{L}_{\mathrm{full}}(\theta_t).
\]
We then construct the DAS basis \(V_t\) and projector \(P_t\) according to
\eqref{eq:das-rank}--\eqref{eq:das-projector}. In particular, the selected
subspace is the smallest collection of Hessian eigendirections that retains
at least a fraction \(\tau\) of the full gradient energy:
\[
\left\lVert
P_tg_t
\right\rVert_2^2
\geq
\tau
\left\lVert
g_t
\right\rVert_2^2.
\]

\subsubsection{First-Order Methods}
\label{sec:full-batch-first-order}

We begin with gradient descent (GD), because its update direction is directly
determined by the gradient and therefore provides the clearest setting for
examining the effect of the DAS restriction. The full-space and
DAS-restricted updates are, respectively,
\begin{equation}
\theta_{t+1}^{\mathrm{GD}}
=
\theta_t-\eta\,g_t,
\qquad
\theta_{t+1}^{\mathrm{DAS\text{-}GD}}
=
\theta_t-\eta\,P_tg_t.
\label{eq:full-batch-gd-updates}
\end{equation}
The DAS-GD update retains at least a \(\tau\)-fraction of the squared
gradient norm by construction, while removing components outside the
gradient-aligned Hessian eigenspace. Each update is globalized either with a
fixed learning rate \(\eta\) or with one of the backtracking or Wolfe line
searches of Section~\ref{sec:globalization}.

We do not include Adam \citep{kingma2015adam} as a primary first-order
baseline in this analysis. Adam uses coordinatewise first- and second-moment estimates, producing an
adaptive diagonal rescaling that changes the effective update direction
relative to the gradient. Consequently, the energy-threshold criterion used
to define the DAS does not directly characterize the energy retained in an
Adam update. A careful study of the DAS combined with adaptive
coordinatewise preconditioners is therefore left for future work.

\begin{figure}[!htbp]
    \centering
    \includegraphics[width=1\linewidth]{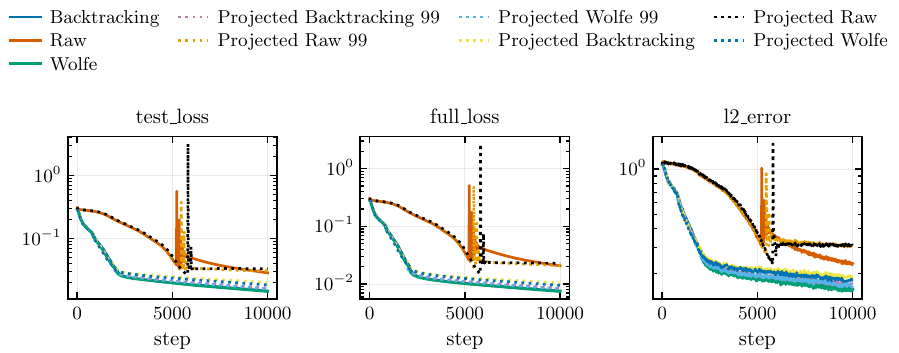}
    \caption{
    Full-batch first-order optimization of wave-1D (\(\tau=0.95 \text{ or  }\tau=0.99\)). Test
    loss, training loss, and relative \(L^2\) error versus iteration for
    full-space GD (solid) and DAS-GD (dotted) with a fixed learning rate
    (``Raw''), backtracking, and Wolfe line search. With a line search,
    DAS-GD matches its full-space counterpart. With a fixed learning rate both
    the full-space and the DAS-GD run show a loss spike near iteration 5000,
    after which the DAS-GD run makes little further progress; this is the
    spurious-alignment collapse analyzed, for longer runs and several values
    of \(\tau\), in Appendix~\ref{app:spurious}.
    }
    \label{fig:wave-first-order-loss}
\end{figure}

\begin{figure}[!htbp]
    \centering

    \begin{subfigure}[b]{\linewidth}
        \centering
        \includegraphics[width=0.45\linewidth]{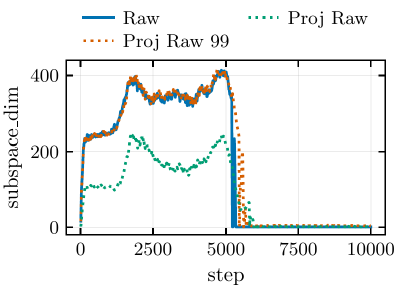}
        \hfill
        \includegraphics[width=0.45\linewidth]{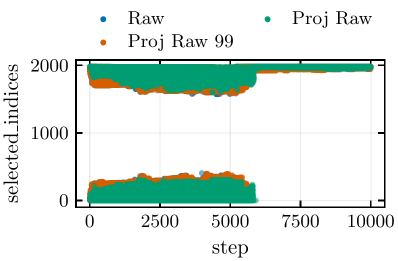}
        \label{fig:wave-size-raw}
    \end{subfigure}

    \vspace{0.60em}

    \begin{subfigure}[b]{\linewidth}
        \centering
        \includegraphics[width=0.45\linewidth]{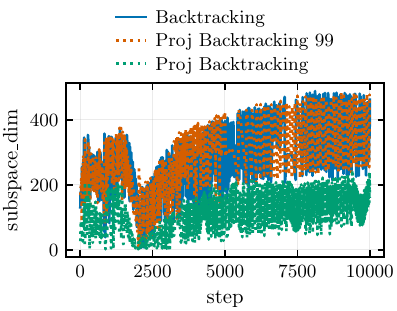}
        \hfill
        \includegraphics[width=0.45\linewidth]{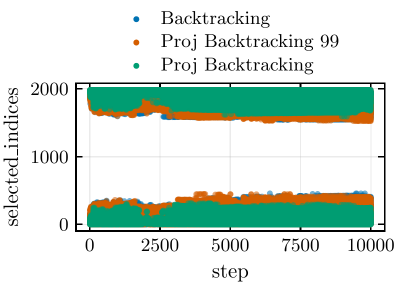}
        \label{fig:wave-size-bt}
    \end{subfigure}
    
    \vspace{0.60em}
    
    \begin{subfigure}[b]{\linewidth}
        \centering
        \includegraphics[width=0.45\linewidth]{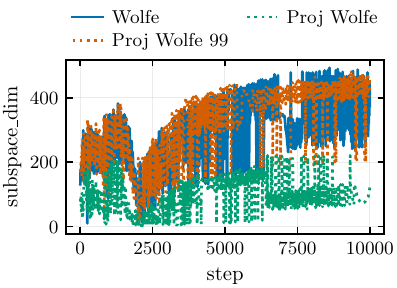}
        \hfill
        \includegraphics[width=0.45\linewidth]{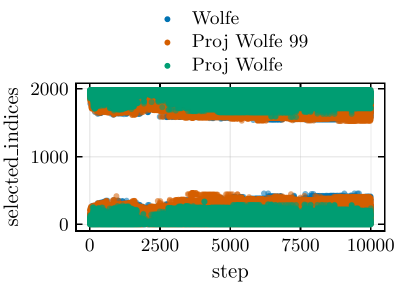}
        \label{fig:wave-size-wolfe}
    \end{subfigure}

    \caption{
    DAS structure along the wave-1D trajectories of
    Figure~\ref{fig:wave-first-order-loss}. Rows, top to bottom: fixed
    learning rate, backtracking, Wolfe. Left column: DAS
    dimension \(k_t\) versus iteration, computed along the full-space
    trajectory (solid) and along the DAS-GD trajectory (dotted). Right
    column: the indices of the selected eigenvectors at each iteration, with
    eigenvectors indexed by increasing eigenvalue (index \(0\) is the
    smallest eigenvalue, index \(p-1=1980\) the largest). The selected
    directions come from both ends of the spectrum --- the largest
    eigenvalues and the near-zero ones --- and the
    DAS-GD trajectory consistently needs fewer directions than the
    full-space trajectory. For the fixed-step run, only the top eigenvector
    remains selected after iteration \(\approx 5000\).
    }
    \label{fig:wave-solution-line-search-comparison}
\end{figure}

We additionally visualize the DAS dimension \(k_t\) and the Hessian
eigendirections selected by the DAS over the course of optimization
(Figure~\ref{fig:wave-solution-line-search-comparison}). The solid curves
in the left column are computed along the \emph{unrestricted} full-space
trajectories: at every iterate of ordinary GD, \(95\%\) of the gradient
energy lies in a subspace that is small compared with the \(p=1981\) Hessian
eigendirections, with a size that depends on the line-search method. This is the direct evidence that the gradient does not
live in the full parameter space but in a small, changing subspace of it;
the DAS-GD runs (dotted) then test whether discarding the remaining
\(5\%\) or \(1\%\) of the energy changes the trajectory. For the fixed-step
run, \(k_t\) collapses to a single direction --- the eigenvector with the
largest eigenvalue --- after the loss spike near iteration 5000, which is
the spurious-alignment phenomenon discussed by \citet{song2025does} (see
also \citealp{deng2026suspicious}); we analyze it in
Appendix~\ref{app:spurious}. The collapse is not seen for the
line-search variants, because their acceptance tests prevent the
overshooting that produces it.

\begin{figure}[!htbp]
    \centering
    \begin{subfigure}[b]{\linewidth}
        \centering
        \includegraphics[width=0.5\linewidth]{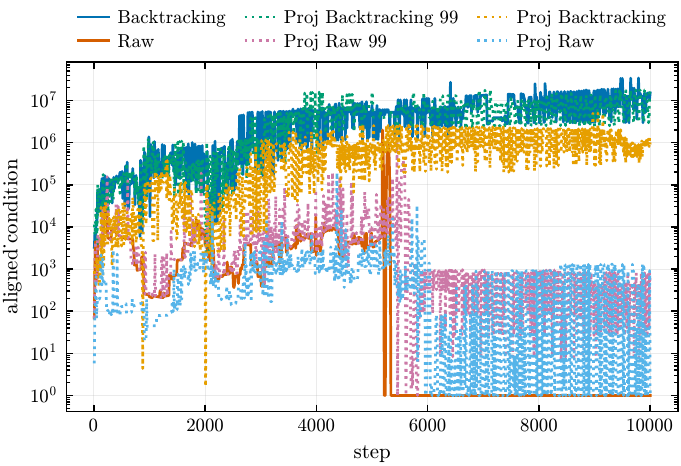}
        \hfill
        \includegraphics[width=0.5\linewidth]{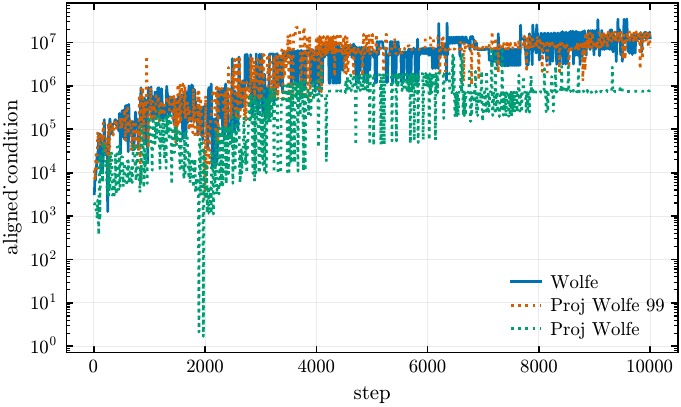}
    \end{subfigure}

    \caption{
    DAS condition number \(\kappa_{\mathcal{A}}(H_t)\) (labeled ``aligned
    condition'' on the vertical axis) versus iteration for the wave-1D
    runs of Figure~\ref{fig:wave-first-order-loss}. Top: fixed learning rate
    and backtracking; bottom: Wolfe; solid lines are the full-space
    trajectories and dotted lines the DAS-GD trajectories. Even
    inside the DAS the conditioning remains severe
    (\(\kappa_{\mathcal{A}}\sim10^{5}\)--\(10^{7}\) for the line-search
    runs). For the fixed-step run \(\kappa_{\mathcal{A}}\) drops to one
    after iteration \(\approx5000\) because the DAS has collapsed to very few directions around 4 in this case.
    }
    \label{fig:wave-AAC}
\end{figure}

The central quantity for interpreting DAS-restricted optimization is not
only the dimension of the selected subspace, but also the curvature
represented within it, measured by the DAS condition number
\(\kappa_{\mathcal{A}}(H_t)\) of \eqref{eq:das-condition-number}. A
DAS-restricted method can see a substantially smaller curvature range than
the full-space method if the DAS excludes directions associated with very
small-magnitude eigenvalues or otherwise poorly conditioned spectral
components. In this case we still include the negative eigenmodes as they are vital for first order optimizers and they carry real gradients\citep{alain2019negative}. This offers a possible explanation for cases in which a
lower-dimensional DAS-restricted trajectory achieves a smaller solution error
than a full-space trajectory. Such behavior should be interpreted as
empirical evidence that the selected subspace acts as a spectral filter,
rather than as a general guarantee that restriction improves optimization. Figure~\ref{fig:wave-AAC} reports \(\kappa_{\mathcal{A}}(H_t)\) for the
wave-1D experiment. 

Together with
Figure~\ref{fig:wave-solution-line-search-comparison}, this diagnostic shows
how the energy threshold controls a tradeoff between retaining more of the
gradient and admitting additional curvature variation into the restricted
update: a larger threshold (e.g., \(\tau=0.99\)) retains more instantaneous
gradient energy than a smaller one (\(\tau=0.95\)), but it can also require
more eigendirections and may include directions associated with less
favorable local conditioning.

\begin{figure}[!htbp]
    \centering
    \includegraphics[width=1\linewidth]{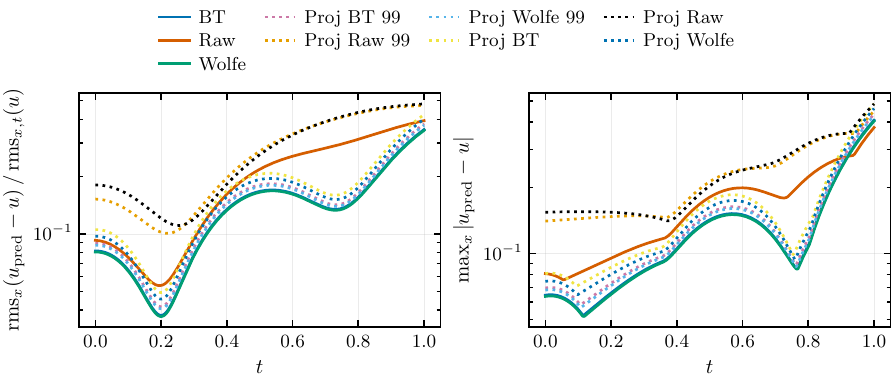}

    \caption{
    Wave-1D error as a function of time for the runs of
    Figure~\ref{fig:wave-fields}: relative root-mean-square error over
    \(x\) (left), maximum error over \(x\) (center), and root-mean-square
    PDE residual over \(x\) (right) at each time \(t\). Top: fixed-step and
    backtracking; bottom: Wolfe. Solid lines are full-space
    runs and dotted lines DAS-GD runs. The error grows with \(t\) for every
    method, and DAS-GD tracks its full-space counterpart closely.
    }
    \label{fig:wave-error-vs-t}
\end{figure}

\begin{figure}[!htbp]
    \centering
    \includegraphics[width=0.8\linewidth]{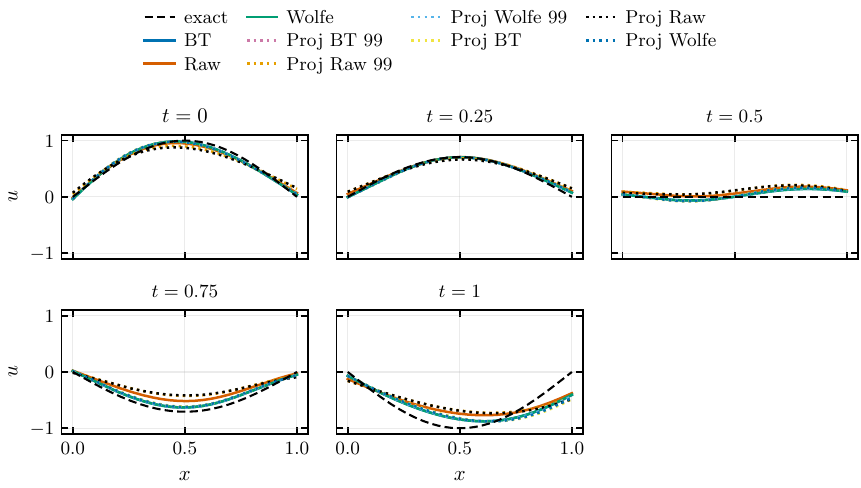}

    \caption{
    snapshots \(u(x,t_k)\) at fixed times \(t_k\) comparing the
    exact solution (dashed) with the full-space (solid) and DAS-GD (dotted)
    predictions.
    }
    \label{fig:wave-snapshots}
\end{figure}

\begin{figure}[!htbp]
    \centering
    \includegraphics[width=0.64\linewidth]{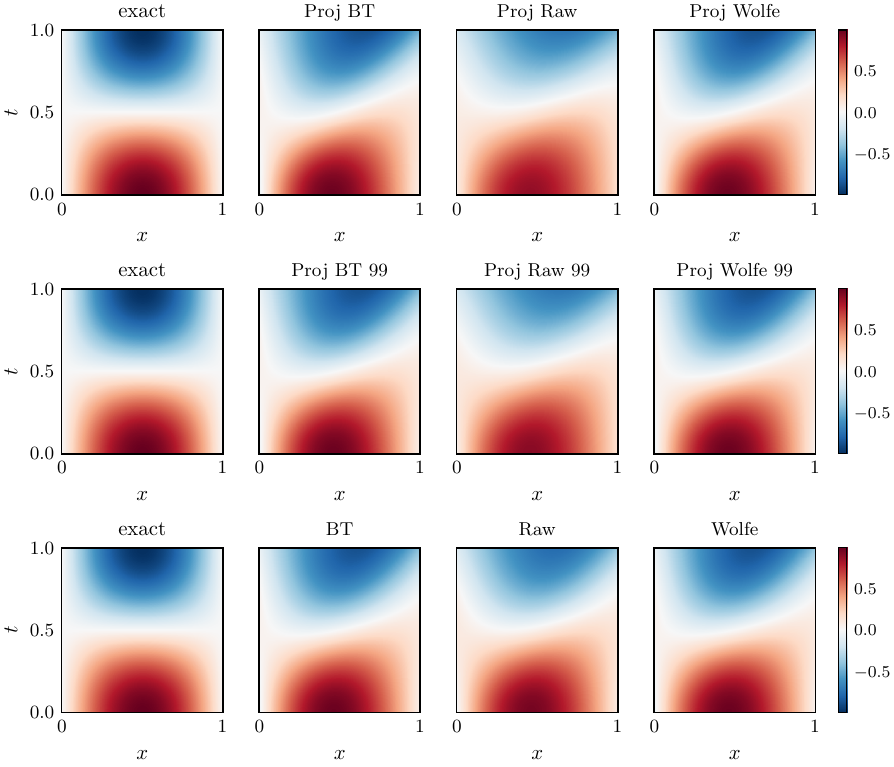}

    \vspace{0.4em}
    \includegraphics[width=0.64\linewidth]{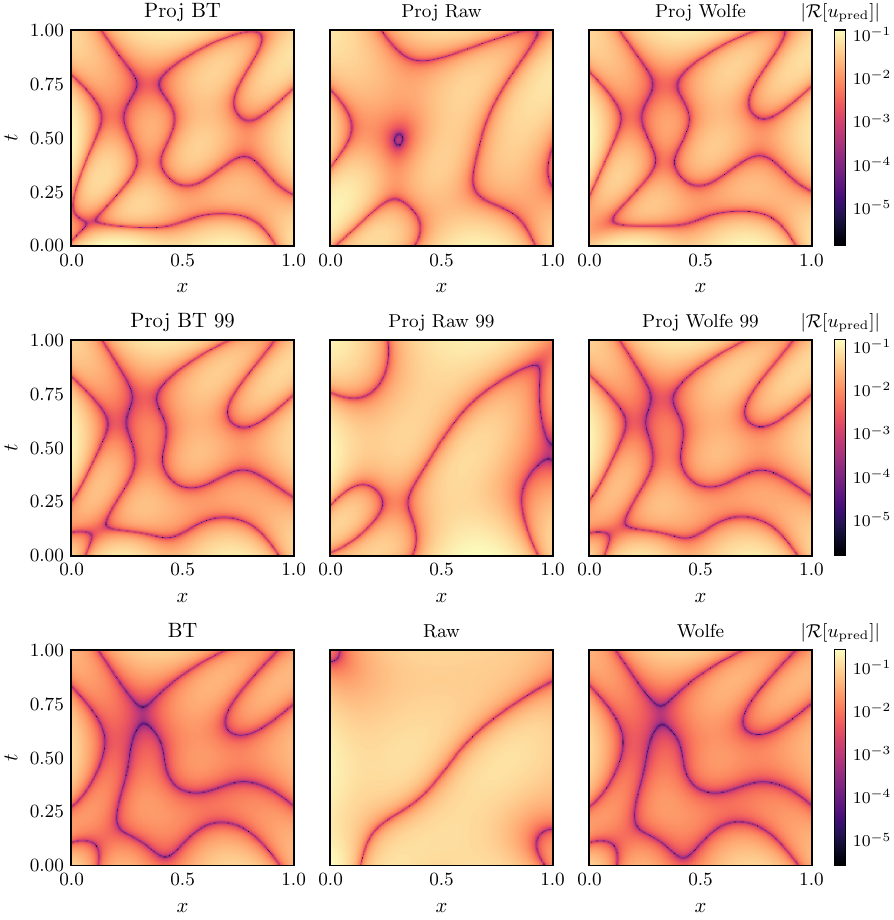}

    \caption{
    Wave-1D solution fields and PDE residuals after full-batch first-order
    training (\(\tau=0.95\)). Rows 1--2: the predicted field \(u(x,t)\) for
    the exact solution (leftmost panel) and for the full-space and DAS-GD
    runs named in the panel titles, for the fixed-step and backtracking
    runs (row 1) and the Wolfe runs (row 2). Rows 3--4:
    the corresponding pointwise PDE residual \(|\mathcal{R}[u_\theta]|\).
    }
    \label{fig:wave-fields}
\end{figure}

Figures~\ref{fig:wave-error-vs-t}--\ref{fig:wave-fields} compare the learned
solutions from DAS-GD and full-space GD at the solution level. The field
plots, residuals, pointwise errors, and snapshots complement the loss-based
metrics by showing whether the optimized PINN reproduces the wave dynamics
and by localizing the regions in which the approximation error remains
large.

\begin{table}[!htbp]
\centering
\footnotesize
\setlength{\tabcolsep}{4pt}
\begin{tabular}{lcccccccccc}
\toprule
 & Rel. & Rel. & Rel. & Rel. & Max &  & BC & IC & Res. & Res. \\
Method & $L^2$ & $L^\infty$ & $L^1$ & $H^1$ & err. & RMSE & err. & err. & RMS & max \\
\midrule
Proj BT & 0.1920 & 0.4844 & 0.1811 & 0.3388 & 0.4844 & 0.0961 & 0.4844 & 0.0811 & 0.0312 & 0.1365 \\
Proj Raw & 0.3094 & 0.4842 & 0.3040 & 0.4145 & 0.4842 & 0.1548 & 0.4842 & 0.1547 & 0.0338 & 0.1084 \\
Proj Wolfe & 0.1807 & 0.4615 & 0.1698 & 0.3196 & 0.4615 & 0.0904 & 0.4615 & 0.0747 & 0.0298 & 0.1165 \\
\midrule
Proj BT 99 & 0.1696 & 0.4356 & 0.1588 & 0.3001 & 0.4356 & 0.0848 & 0.4356 & 0.0696 & 0.0277 & 0.1210 \\
Proj Raw 99 & 0.3068 & 0.4498 & 0.3003 & 0.3968 & 0.4498 & 0.1535 & 0.4498 & 0.1409 & 0.0455 & 0.1525 \\
Proj Wolfe 99 & 0.1665 & 0.4292 & 0.1558 & 0.2950 & 0.4292 & 0.0833 & 0.4292 & 0.0681 & 0.0273 & 0.1214 \\
\midrule
BT & 0.1582 & 0.4079 & 0.1476 & 0.2792 & 0.4079 & 0.0791 & 0.4079 & 0.0645 & 0.0251 & 0.1085 \\
Raw & 0.2330 & 0.3777 & 0.2218 & 0.3156 & 0.3777 & 0.1165 & 0.3777 & 0.0810 & 0.0915 & 0.2708 \\
Wolfe & 0.1565 & 0.4038 & 0.1459 & 0.2763 & 0.4038 & 0.0783 & 0.4038 & 0.0637 & 0.0249 & 0.1088 \\
\bottomrule
\end{tabular}
\caption{Error norms for the 1D wave equation (MLP 2-30-30-30-1, $\tanh$) trained with full-batch GD.}
\label{tab:wave1d-sgd-error-norms}
\end{table}

\newpage

\FloatBarrier
\subsubsection{Gauss--Newton Methods}
\label{sec:full-batch-gauss-newton}

We next apply the DAS to curvature-aware optimization using the
DAS-restricted Gauss--Newton step \eqref{eq:das-gauss-newton}. At
iteration \(t\), the DAS basis \(V_t\in\mathbb{R}^{p\times k_t}\) is rebuilt
from the current full-batch Gauss--Newton matrix \(J_t^\top J_t\) and
gradient, and satisfies
\begin{equation}
\left\lVert
V_t^\top g_t
\right\rVert_2^2
=
\left\lVert
P_tg_t
\right\rVert_2^2
\geq
\tau
\left\lVert
g_t
\right\rVert_2^2
\label{eq:das-gradient-coverage}
\end{equation}
whenever the eigenvectors retained above the numerical floor of
Section~\ref{sec:conditioning} can reach this coverage; otherwise all
retained eigenvectors are used.
The step is computed in DAS coordinates, \(\theta=\theta_t+V_ty\), by
solving the damped reduced system \eqref{eq:das-gauss-newton} for
\(y_t\in\mathbb{R}^{k_t}\) and mapping back to \(p_t=V_ty_t\). The
resulting direction satisfies
\[
p_t
\in
\operatorname{span}(V_t),
\qquad
P_tp_t=p_t,
\]
so every trial point considered by a line search remains in the affine
subspace \(\theta_t+\operatorname{span}(V_t)\).

When
\(
V_t^\top J_t^\top J_tV_t+\mu_tI
\)
is positive definite and \(V_t^\top g_t\neq0\), the DAS-GN direction is a
strict descent direction for the full objective:
\begin{align}
g_t^\top p_t
&=
(V_t^\top g_t)^\top y_t
\nonumber\\
&=
-
(V_t^\top g_t)^\top
\left(
V_t^\top J_t^\top J_tV_t+\mu_tI
\right)^{-1}
(V_t^\top g_t)
<0.
\label{eq:reduced-gn-descent}
\end{align}
Moreover, because \(p_t=P_tp_t\),
\begin{equation}
g_t^\top p_t
=
\left(
P_tg_t
\right)^\top p_t .
\label{eq:full-and-projected-derivative}
\end{equation}
Therefore, evaluating the line-search directional derivative using the full
gradient or its DAS component gives the same scalar value.

The reduced system in \eqref{eq:das-gauss-newton} has dimension
\(k_t\times k_t\) rather than \(p\times p\). The basis is obtained from a
singular value decomposition of the Jacobian at every iteration, which is
the expected cost of identifying the subspace exactly. The experiments ask
whether a subspace of dimension \(k_t\ll p\) suffices to reach the solution,
and are not a run-time comparison; ideas on how to compute the DAS more
efficiently are discussed in
Section~\ref{sec:discussion-scalability}.

Equation~\eqref{eq:das-gradient-coverage} also rules out a spurious
stationary point caused only by the restriction. Since \(\tau>0\),
\[
P_tg_t=0
\quad\Longrightarrow\quad
g_t=0.
\]
Thus, if the DAS selection criterion is satisfied, a zero restricted
gradient implies a zero full gradient. This statement concerns first-order
stationarity; it does not imply that every stationary point is a local
minimum.

We use backtracking and Wolfe line searches for the DAS-GN experiments.
Both procedures operate without modification once the reduced direction has
been mapped back to the ambient parameter space: candidate iterates take the
form \(\theta_t+\alpha p_t\), and the sufficient-decrease and curvature
conditions are evaluated using the full objective and full gradient.

Levenberg--Marquardt (LM) and trust-region globalization raise a question
that a line search does not. Both adapt a scalar that persists across
iterations, the damping \(\mu_t\) or the trust radius, from the gain ratio
\begin{equation}
\rho_t
=
\frac{\mathcal{L}(\theta_t)-\mathcal{L}(\theta_t+p_t)}
     {m_t(0)-m_t(p_t)},
\label{eq:gain-ratio}
\end{equation}
which compares the decrease of the objective \(\mathcal{L}\) actually
obtained with the decrease predicted by the local Gauss--Newton model
\(m_t\). Because \(p_t=V_ty_t\), the predicted decrease is that of the
reduced model,
\[
m_t(0)-m_t(p_t)
=
-(V_t^\top g_t)^\top y_t
-\tfrac12\left\lVert J_tV_ty_t\right\rVert_2^2 ,
\]
so the ratio involves no curvature outside \(\operatorname{span}(V_t)\). The
update rule nevertheless presumes that consecutive models are comparable, so
that a damping value which produced a reliable step at iteration \(t\) is a
sensible starting point at iteration \(t+1\). In the full space this holds,
because \(m_t\) and \(m_{t+1}\) model the same objective in the same
coordinates. In the DAS the reduced model is rebuilt at every iteration on a
different basis \(V_t\) of different dimension \(k_t\), and its curvature
\(V_t^\top J_t^\top J_tV_t\) changes with it. One might expect a
damping value or radius calibrated on \(\operatorname{span}(V_t)\) to carry
little information about \(\operatorname{span}(V_{t+1})\), and LM to lose
more from the restriction than a line search, which relies on no such
memory: it tests the objective along the current direction only. We include
DAS-LM runs at both coverage levels to test this, here and in the mini-batch
setting of Section~\ref{sec:mini-batch-gauss-newton}, and do not use
trust-region globalization; a trust-region formulation that accounts for the
changing reduced geometry, in the spirit of \citet{hoang2026adaptive}, is
left for future work.

\begin{figure}[!htbp]
    \centering
    \includegraphics[width=\linewidth]{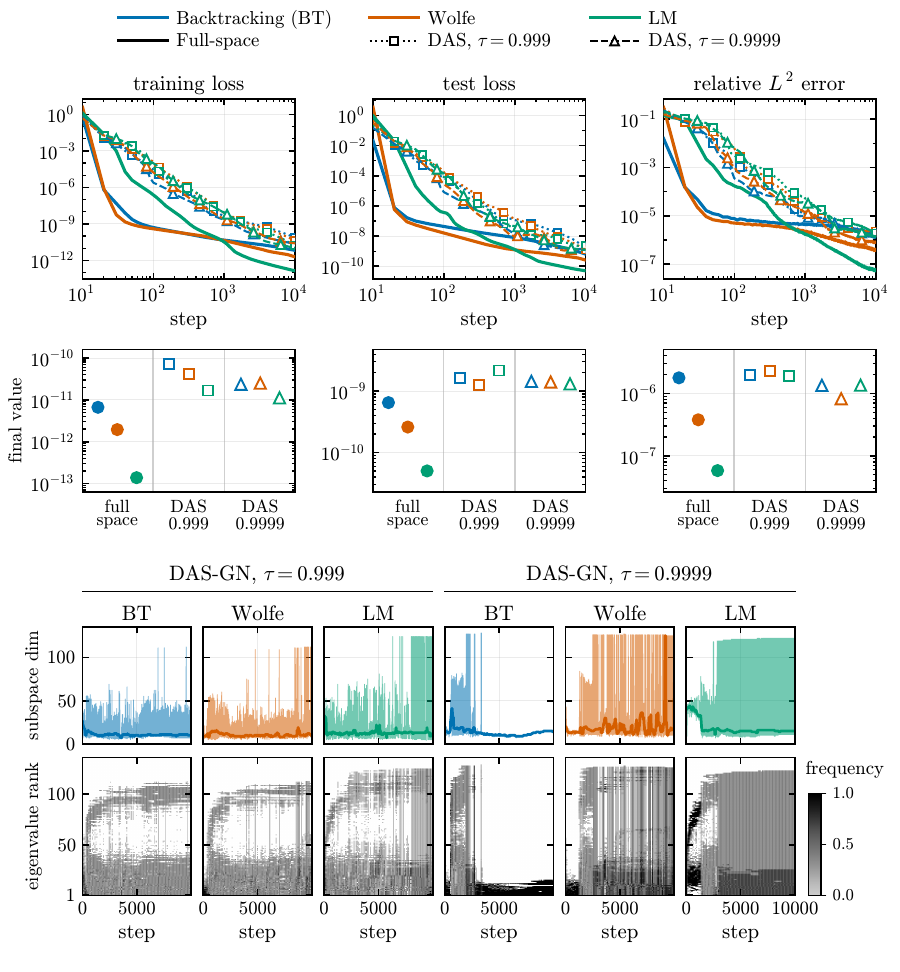}
    \caption{Full-batch Gauss--Newton optimization of Poisson-2D. First
    row: training loss, test loss, and relative \(L^2\) error versus
    iteration for full-space Gauss--Newton (solid) and DAS-GN with
    \(\tau=0.999\) (dotted, squares) and \(\tau=0.9999\) (dashed,
    triangles); color gives the globalization: backtracking (BT), Wolfe, or
    Levenberg--Marquardt (LM). Second row: the values of the same three
    quantities after \(10^4\) iterations, one marker per run. Third row:
    DAS dimension \(k_t\) (thin lines: logged values; thick lines: running
    median over 250 iterations). Fourth row: how often each eigenvector of
    the Gauss--Newton matrix is selected, per window of 100 iterations, with
    eigenvectors ranked by eigenvalue (rank 1 is the largest; white: never
    selected).}
    \label{fig:poisson2d_loss}
\end{figure}

\begin{figure}[!htbp]
    \centering
    \includegraphics[width=0.85\linewidth]{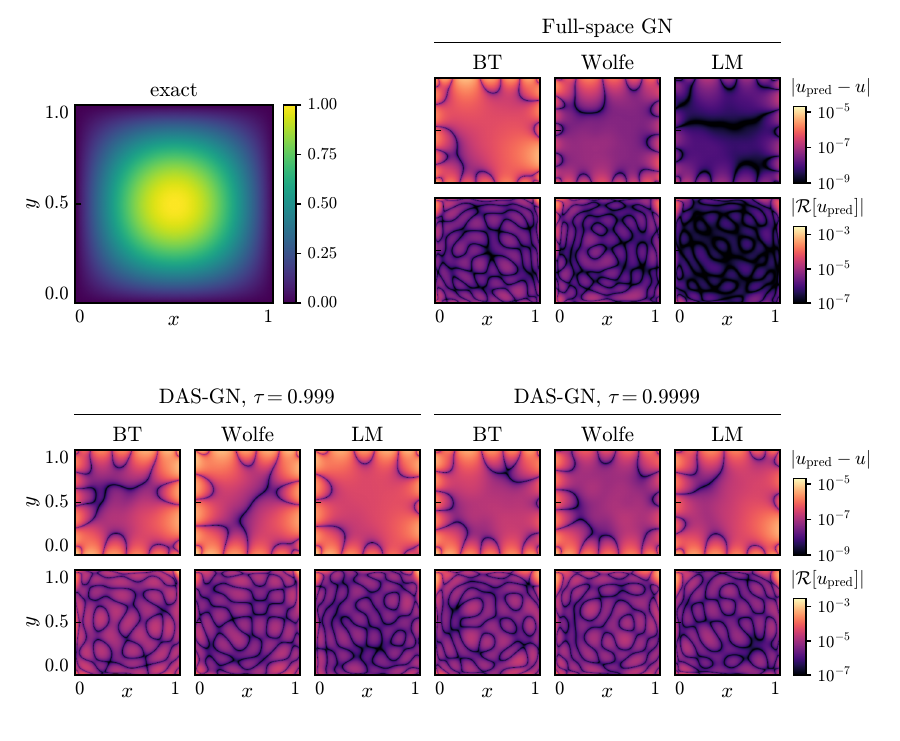}
    \caption{Poisson-2D solutions after full-batch Gauss--Newton training.
    Top left: the exact solution. Top right: full-space Gauss--Newton.
    Bottom: DAS-GN. In each block the upper row is the pointwise error
    \(|u_\theta-u|\) and the lower row the pointwise PDE residual
    \(|\mathcal{R}[u_\theta]|\), for the globalization named in the panel
    title; both color scales are logarithmic and shared by all runs, so
    panels are directly comparable. The learned fields are not shown: they
    are indistinguishable from the exact solution at this scale (relative
    \(L^2\) error below \(2.3\times10^{-6}\)).}
    \label{fig:poisson2d_sol}
\end{figure}

Figures~\ref{fig:poisson2d_loss} and~\ref{fig:poisson2d_sol} summarize the
full-batch Gauss--Newton experiments on the Poisson-2D benchmark for
coverage \(\tau=0.999\) and \(\tau=0.9999\). The first figure reports
optimization and solution-quality metrics together with the DAS dimension
and the selected eigendirections along the DAS trajectories; the second
shows the exact solution together with the pointwise errors and residuals
of all runs. With a line
search, DAS-GN reaches accuracy comparable to full-space Gauss--Newton: the
final relative \(L^2\) errors are \(0.8\times10^{-6}\)--\(2.3\times10^{-6}\),
against \(1.8\times10^{-6}\) (backtracking) and \(3.8\times10^{-7}\) (Wolfe)
in the full space. It does so with a small fraction of the directions: the
median \(k_t\) is \(10\)--\(16\), under one percent of the \(p=1981\)
parameters, and \(k_t\le 50\) in \(73\)--\(97\%\) of the logged iterations.
The spikes to \(k_t\approx110\)--\(130\) are iterations in which the
retained eigenvectors cannot reach the coverage \(\tau\), so that all of
them are used; this occurs in \(0.3\)--\(3.6\%\) of the logged iterations
for \(\tau=0.999\) and in \(2\)--\(24\%\) for \(\tau=0.9999\). The selected
directions come mostly from the leading \(\sim\!45\) eigenvectors and,
intermittently, from those with the smallest eigenvalues above the numerical
floor (Figure~\ref{fig:poisson2d_loss}, fourth row). As optimization
diagnostics, the restricted trajectories are longer and stop at a higher
training loss: the full-space line-search runs bring the training loss below
\(10^{-9}\) within \(70\) iterations, whereas DAS-GN needs
\(1300\)--\(2600\), and its final training loss is roughly \(4\)--\(22\)
times higher. This difference does not carry over to the solution, where the
errors above agree to within the spread of the two full-space runs.

Table~\ref{tab:poisson2d-gn-rank-savings} quantifies the reduction. Averaged over a
trajectory the DAS uses \(18\)--\(44\) directions (median \(10\)--\(16\)),
that is, \(0.9\)--\(2.2\%\) of the \(p=1981\) parameters, so on average more
than \(97\%\) of the directions are left untouched. The reduced
Gauss--Newton matrix \(V_t^\top J_t^\top J_tV_t\) then has on average
\(1{,}000\)--\(6{,}900\) times fewer entries than the \(p\times p\) matrix,
and a dense solve, whose cost grows cubically with the dimension, is on
average \(2\times10^{4}\)--\(3\times10^{5}\) times cheaper per iteration.
This outweighs the larger iteration count: summing \(k_t^3\) over the
iterations needed to bring the training loss below \(10^{-9}\) (estimated
from the logged values) gives \(4\times10^{3}\)--\(7\times10^{4}\) times
less solve work than the corresponding sum of \(p^3\) for full-space
Gauss--Newton with the same globalization. These figures describe the
reduced solve; the basis is built from a full decomposition at every
iteration, so they are a measure of the size of the subspace and not of run
time.

On this benchmark the LM runs are consistent with that expectation, and
they are the one case in which the DAS does not match the full-space
solution error. In the full space LM is the most accurate method, with a
relative \(L^2\) error of \(5.8\times10^{-8}\), \(6\)--\(31\) times lower
than that of the two line searches. In the DAS this advantage is lost: the
relative \(L^2\) errors of DAS-LM (\(1.9\times10^{-6}\) for \(\tau=0.999\)
and \(1.4\times10^{-6}\) for \(\tau=0.9999\)) lie within the range of the
DAS line searches and are \(24\)--\(33\) times higher than that of
full-space LM. DAS-LM neither stalls nor diverges, and it matches the
solution error of the full-space line searches; what it does not reproduce
is the additional accuracy that LM gains in the full space. The diagnostics
show the same pattern: DAS-LM reaches the lowest training loss among the DAS
runs (\(1.7\times10^{-11}\) and \(1.1\times10^{-11}\)), but only by a factor
of \(2\)--\(4\) over the DAS line searches, against a factor of
\(14\)--\(48\) in the full space (\(1.4\times10^{-13}\)), and its test
losses (\(2.2\times10^{-9}\) and \(1.3\times10^{-9}\)) are no lower than
theirs (\(1.2\times10^{-9}\)--\(1.6\times10^{-9}\)). Along the two
DAS-LM trajectories \(k_t\) ranges from \(4\) to \(124\) and changes by as
much as \(120\) between consecutive logged iterations, while the ratio of
the largest to the smallest selected eigenvalue ranges from about \(7\) to
\(10^{14}\); consecutive reduced models are far from comparable. The
mini-batch experiments of Section~\ref{sec:mini-batch-gauss-newton}, where
the reduced models change at least as much, do not show the same loss, so
the Poisson-2D result should not be read as a general property of DAS-LM.

\begin{table}[!htbp]
\centering
\small
\setlength{\tabcolsep}{4pt}
\begin{tabular}{lllcccccc}
\toprule
 & & & \multicolumn{2}{c}{\(k_t\)} & Training & Test & Relative & Iterations to \\
Method & Global. & \(\tau\) & median & mean & loss & loss & \(L^2\) error & loss \(<10^{-9}\) \\
\midrule
Full-space & BT & --- & --- & --- & \(6.6\times10^{-12}\) & \(6.5\times10^{-10}\) & \(1.8\times10^{-6}\) & 70 \\
Full-space & Wolfe & --- & --- & --- & \(1.9\times10^{-12}\) & \(2.6\times10^{-10}\) & \(3.8\times10^{-7}\) & 50 \\
Full-space & LM & --- & --- & --- & \(1.4\times10^{-13}\) & \(5.0\times10^{-11}\) & \(5.8\times10^{-8}\) & 330 \\
\midrule
DAS & BT & 0.999 & 10 & 18.5 & \(7.3\times10^{-11}\) & \(1.6\times10^{-9}\) & \(2.0\times10^{-6}\) & 2580 \\
DAS & Wolfe & 0.999 & 10 & 18.3 & \(4.2\times10^{-11}\) & \(1.2\times10^{-9}\) & \(2.3\times10^{-6}\) & 2270 \\
DAS & LM & 0.999 & 13 & 21.2 & \(1.7\times10^{-11}\) & \(2.2\times10^{-9}\) & \(1.9\times10^{-6}\) & 1930 \\
\midrule
DAS & BT & 0.9999 & 12 & 17.8 & \(2.4\times10^{-11}\) & \(1.4\times10^{-9}\) & \(1.3\times10^{-6}\) & 1300 \\
DAS & Wolfe & 0.9999 & 16 & 36.9 & \(2.5\times10^{-11}\) & \(1.4\times10^{-9}\) & \(8.2\times10^{-7}\) & 1600 \\
DAS & LM & 0.9999 & 16 & 43.9 & \(1.1\times10^{-11}\) & \(1.3\times10^{-9}\) & \(1.4\times10^{-6}\) & 1710 \\
\bottomrule
\end{tabular}
\caption{Poisson-2D, full-batch Gauss--Newton: final values after \(10^4\)
iterations, DAS dimension along the trajectory, and the number of iterations
needed to bring the training loss below \(10^{-9}\) (logged every ten
iterations). The
network has \(p=1981\) parameters.}
\label{tab:poisson2d-gn}
\end{table}

\begin{table}[!htbp]
\centering
\small
\setlength{\tabcolsep}{5pt}
\begin{tabular}{llcccccc}
\toprule
 & & & \multicolumn{3}{c}{Directions saved} & Matrix & Dense \\
\cmidrule(lr){4-6}
Global. & \(\tau\) & Mean \(k_t\) & mean & median & worst & entries saved & solve saved \\
\midrule
BT    & 0.999  & 18.5 & 96.3\% & 98.0\% & 77.6\% & 99.77\% & 99.980\% \\
Wolfe & 0.999  & 18.3 & 96.4\% & 98.0\% & 77.6\% & 99.72\% & 99.963\% \\
LM    & 0.999  & 21.2 & 95.8\% & 97.4\% & 75.2\% & 99.58\% & 99.928\% \\
\midrule
BT    & 0.9999 & 17.8 & 96.4\% & 97.6\% & 74.4\% & 99.70\% & 99.949\% \\
Wolfe & 0.9999 & 36.9 & 92.6\% & 96.8\% & 74.6\% & 98.76\% & 99.719\% \\
LM    & 0.9999 & 43.9 & 91.2\% & 96.8\% & 75.6\% & 98.42\% & 99.646\% \\
\midrule
\multicolumn{2}{l}{All runs pooled} & 26.1 & 94.8\% & 97.4\% & 74.4\% & 99.33\% & 99.864\% \\
\bottomrule
\end{tabular}
\caption{Savings of DAS-GN on Poisson-2D relative to the rank of the
Gauss--Newton matrix, \(\operatorname{rank}(J_t^\top J_t)=500\). Directions
saved is \(1-k_t/500\), at the mean, median, and largest \(k_t\) of each
trajectory; matrix entries and dense solve cost compare the trajectory
averages of \(k_t^2\) and \(k_t^3\) with \(500^2\) and \(500^3\).}
\label{tab:poisson2d-gn-rank-savings}
\end{table}

\FloatBarrier
\subsection{Mini-Batch Optimization}
\label{sec:mini-batch-optimization}

We next evaluate the dominant aligned subspace (DAS) under mini-batch
optimization. Unlike the full-batch setting, stochasticity now affects the
gradient used to compute the update direction and can also affect the
selection of the DAS basis (see \citealp{so2026minibatch} for an analysis of
mini-batch noise through fluctuations of the dominant subspace). All experiments use the collocation-point
configurations listed in Table~\ref{tab:collocation}; only interior points
are mini-batched. The DAS is built from mini-batch curvature: the
first-order experiments use the Hessian of the current mini-batch,
diagonalized at every iteration, and the Gauss--Newton experiments use the
minibatch Jacobian (Section~\ref{sec:mini-batch-gauss-newton}). Either
way the basis is recomputed at every iteration, and the mini-batch
experiments are therefore diagnostic and make no claim of computational
savings. At iteration \(t\), let
\[
H_t^{\mathrm{batch}}
=
\nabla^2\mathcal{L}_{\mathrm{batch}}(\theta_t)
\]
denote the Hessian of the mini-batch objective, computed on the same
mini-batch that gives \(g_t^{\mathrm{batch}}\), so the eigenbasis is redrawn
with every batch. Let
\[
g_t^{\mathrm{batch}}
=
\nabla\mathcal{L}_{\mathrm{batch}}(\theta_t)
\]
and
\[
g_t^{\mathrm{full}}
=
\nabla\mathcal{L}_{\mathrm{full}}(\theta_t)
\]
denote the mini-batch and full-batch gradients, respectively. We compare two
DAS basis-selection strategies.

The first strategy, the \emph{batch-selected DAS}, denoted \texttt{projB},
constructs the DAS basis using the mini-batch Hessian and the mini-batch
gradient as the selecting gradient:
\begin{equation}
P_t^{\mathrm{B}}
=
P_{\mathrm{DAS}}
\left(
H_t^{\mathrm{batch}},
g_t^{\mathrm{batch}}
\right).
\label{eq:mini-batch-projb}
\end{equation}
The second strategy, the \emph{full-gradient-selected DAS}, denoted
\texttt{projF}, constructs the DAS basis from the same mini-batch Hessian
but ranks its eigendirections using the full-batch gradient:
\begin{equation}
P_t^{\mathrm{F}}
=
P_{\mathrm{DAS}}
\left(
H_t^{\mathrm{batch}},
g_t^{\mathrm{full}}
\right).
\label{eq:mini-batch-projf}
\end{equation}
Here \(P_{\mathrm{DAS}}(H,g)\) denotes the projector of
\eqref{eq:das-projector} built from the eigenvectors of \(H\) and the
selecting gradient \(g\).

Both strategies use mini-batch quantities to compute the actual optimization
update. Thus, \texttt{projF} does not remove stochasticity from the update;
rather, it stabilizes the subspace-selection stage by ranking the
mini-batch Hessian eigendirections by their alignment with the gradient of
the full objective. \texttt{projB} forms its step from mini-batch
quantities only, while \texttt{projF} still needs the full-batch gradient
for selection.

\subsubsection{First-Order Methods}
\label{sec:mini-batch-first-order}

We first consider stochastic gradient descent. The full-space mini-batch SGD
update is
\begin{equation}
\theta_{t+1}
=
\theta_t
-
\eta\, g_t^{\mathrm{batch}},
\label{eq:mini-batch-sgd-update}
\end{equation}
where \(\eta\) is the learning rate. For DAS-SGD, the update is
\begin{equation}
\theta_{t+1}
=
\theta_t
-
\eta\,P_tg_t^{\mathrm{batch}},
\qquad
P_t
\in
\left\{
P_t^{\mathrm{B}},
P_t^{\mathrm{F}}
\right\}.
\label{eq:mini-batch-das-sgd-update}
\end{equation}

The two DAS-SGD variants differ only in how the DAS basis is selected. For
\texttt{projB}, mini-batch noise affects both the gradient update and the
selection of the Hessian eigendirections. For \texttt{projF}, mini-batch
noise remains in the update, but the selected eigenspace is determined using
the full-batch gradient. This comparison tests whether a more stable spectral
selection rule improves mini-batch first-order optimization. We run it on
heat-1D with \(\tau=0.95\), mini-batch fraction \(0.5\) of the interior
points, \(10^{4}\) iterations and a single seed. The fixed-step runs use
\(\eta=0.01\); the line searches start from \(\alpha^{(0)}=1\), judge trial
steps on the full-batch loss with slopes measured by the projected gradient
(Section~\ref{sec:setup}), and skip the step when the search is
exhausted. In the DAS runs the Hessian is that of the current mini-batch,
\(H_t^{\mathrm{batch}}\).

\begin{figure}[!htbp]
    \centering

    \begin{subfigure}[b]{1\linewidth}
        \centering
        \includegraphics[width=\linewidth]{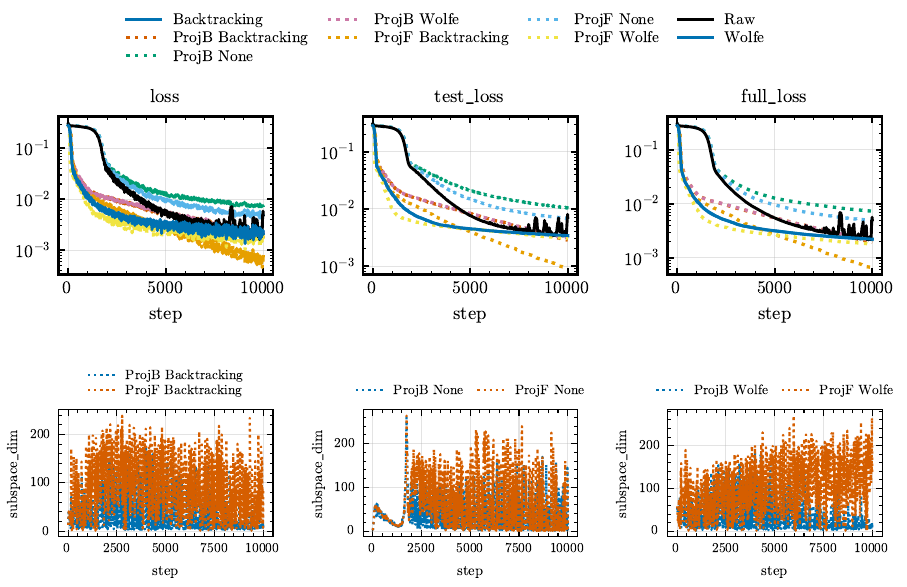}
    \end{subfigure}
    \caption{Mini-batch first-order optimization of heat-1D with the DAS
    built from the mini-batch Hessian (\(\tau=0.95\), mini-batch fraction
    \(0.5\) of the interior points). Top row:
    mini-batch loss, test loss, and full-batch training loss versus
    iteration for full-space SGD (solid) and DAS-SGD with the batch-selected
    (\texttt{projB}) and full-gradient-selected (\texttt{projF}) DAS
    (dotted), each with a fixed learning rate (``Raw'' in the full space,
    ``None'' in the DAS), backtracking, and Wolfe line search. Bottom row:
    DAS dimension \(k_t\) versus iteration for the backtracking,
    fixed-step, and Wolfe DAS-SGD runs.}
    \label{metric-heat}
\end{figure}

\newsavebox{\heatstackbox}

\begin{figure}[!htbp]
    \centering
    \includegraphics[width=0.75\textwidth]{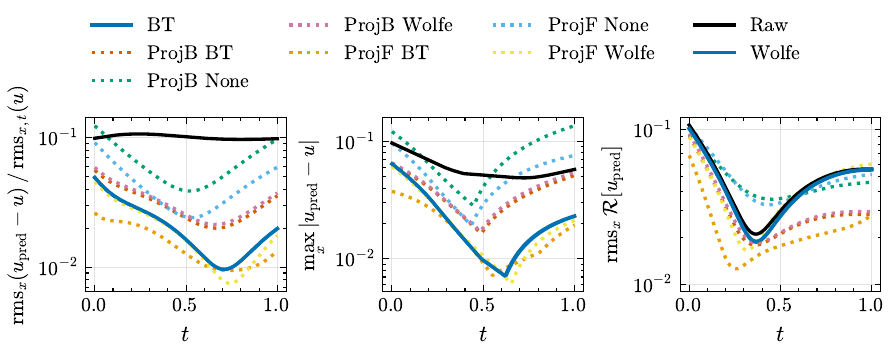}
    \caption{
    Relative root-mean-square error, maximum error, and root-mean-square
    residual over \(x\) as a function of time \(t\) for all runs.
    }
    \label{fig:heat-mb-first-order-sol}
\end{figure}

\begin{figure}[!htbp]
    \centering
    \sbox{\heatstackbox}{\includegraphics[width=0.49\linewidth]{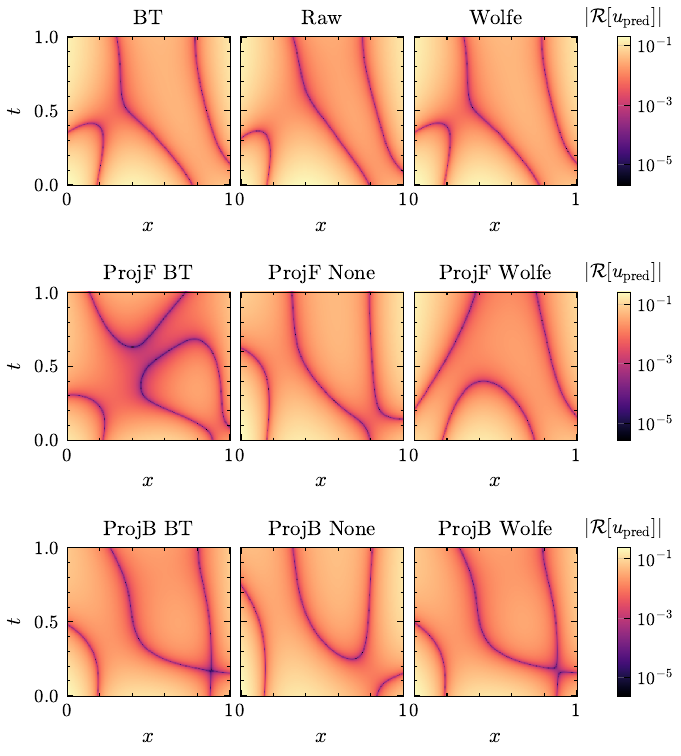}}

    \begin{subfigure}[b]{0.49\linewidth}
        \centering
        \includegraphics[width=\linewidth]{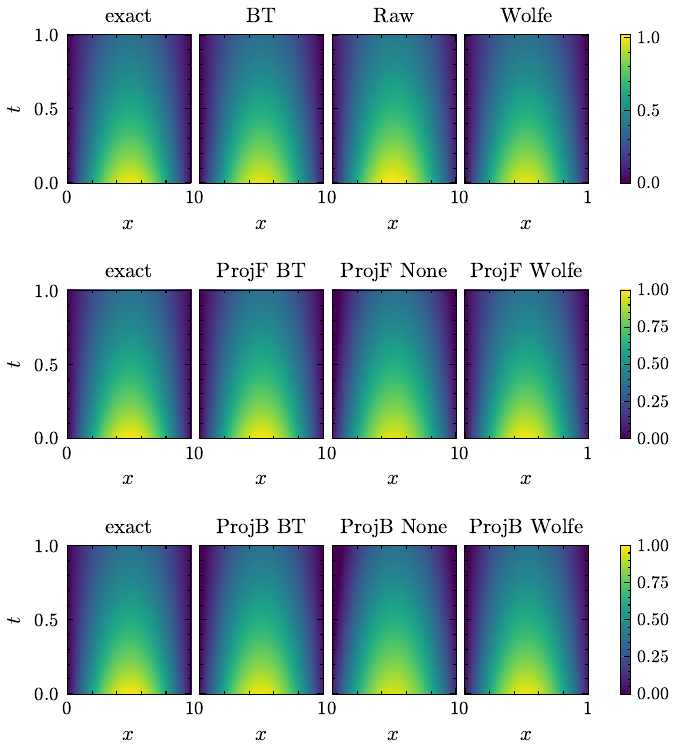}
        \caption{Solution fields \(u(x,t)\).}
        \label{fig:heat-mb-solution-diagnostics-fields}
    \end{subfigure}
    \hfill
    \begin{subfigure}[b]{0.49\linewidth}
        \centering
        \includegraphics[width=\linewidth]{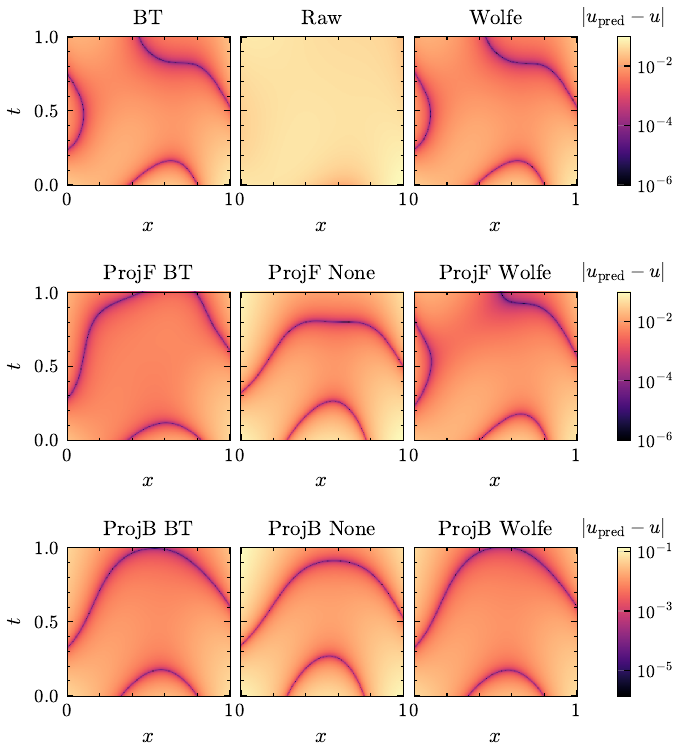}
        \caption{Pointwise error \(|u_\theta-u|\).}
        \label{fig:heat-mb-solution-diagnostics-errors}
    \end{subfigure}

    \vspace{0.6em}

    \begin{subfigure}[b]{0.49\linewidth}
        \centering
        \usebox{\heatstackbox}
        \caption{Pointwise PDE residual \(|\mathcal{R}[u_\theta]|\).}
        \label{fig:heat-mb-solution-diagnostics-residuals}
    \end{subfigure}
    \hfill
    \begin{subfigure}[b]{0.49\linewidth}
        \centering
        \parbox[b][\ht\heatstackbox][c]{\linewidth}{\centering
            \includegraphics[width=\linewidth]{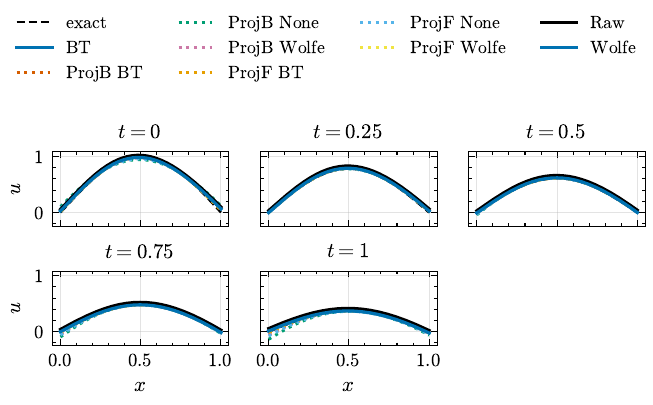}}
        \caption{Snapshots \(u(x,t_k)\) at fixed times.}
        \label{fig:heat-mb-solution-diagnostics-snapshots}
    \end{subfigure}

    \caption{
    Solution diagnostics for the mini-batch first-order heat-1D experiment
    of Figure~\ref{metric-heat}. In every panel group the three
    strips show, top to bottom, the full-space runs, the
    full-gradient-selected DAS-SGD runs (\texttt{projF}), and the
    batch-selected DAS-SGD runs (\texttt{projB}); within each strip the
    panels correspond to the fixed-step, backtracking, and Wolfe variants
    named in the panel titles, preceded by the exact solution where
    applicable.
    }
    \label{fig:heat-mb-solution-diagnostics}
\end{figure}

\begin{table}[!htbp]
\centering
\small
\begin{tabular}{@{}lccc@{}}
\toprule
Step rule & Full space & \texttt{projB} & \texttt{projF} \\
\midrule
\multicolumn{4}{@{}l}{\emph{Final full-batch training loss}} \\
Fixed step   & \(5.1\times10^{-3}\) & \(7.3\times10^{-3}\) & \(4.9\times10^{-3}\) \\
Backtracking & \(2.2\times10^{-3}\) & \(1.9\times10^{-3}\) & \(6.6\times10^{-4}\) \\
Wolfe        & \(2.2\times10^{-3}\) & \(2.1\times10^{-3}\) & \(1.8\times10^{-3}\) \\
\midrule
\multicolumn{4}{@{}l}{\emph{Final relative \(L^2\) error}} \\
Fixed step   & \(0.102\) & \(0.067\) & \(0.045\) \\
Backtracking & \(0.023\) & \(0.031\) & \(0.016\) \\
Wolfe        & \(0.023\) & \(0.033\) & \(0.021\) \\
\bottomrule
\end{tabular}
\caption{Final full-batch training loss and final relative \(L^2\) error
after \(10^{4}\) iterations of mini-batch first-order optimization on
heat-1D (the runs of Figure~\ref{metric-heat}).}
\label{tab:heat-mb-first-order}
\end{table}

Table~\ref{tab:heat-mb-first-order} and Figure~\ref{metric-heat} show that
full-gradient selection helps. In every step rule \texttt{projF} ends below
\texttt{projB}, in full-batch loss and in relative \(L^2\) error. With
backtracking, \texttt{projF} ends three times below full-space SGD in
full-batch loss and has the smallest relative \(L^2\) error of the nine
runs; with Wolfe it ends level with full-space SGD. \texttt{projB} with a
line search ends level with full-space SGD in full-batch loss, with a
somewhat larger relative \(L^2\) error. With the fixed step both DAS
variants have a smaller relative \(L^2\) error than full-space SGD, although
only \texttt{projF} matches its full-batch loss: the loss and the error of
the fixed-step runs do not move together. The DAS stays small, with median
dimensions of \(18\)--\(24\) for \texttt{projB} and \(39\)--\(107\) for
\texttt{projF}. With a line search the full-batch loss never increases, as
the acceptance rule requires, and the search is exhausted in \(16\)--\(27\%\)
of the logged iterations. Figures~\ref{fig:heat-mb-first-order-sol}
and~\ref{fig:heat-mb-solution-diagnostics} show the resulting errors,
residuals and solutions over the space--time domain.

These are single-seed runs, and mini-batch trajectories are sensitive to
small changes in the run, so differences of a factor of two or less, such
as between \texttt{projB} and full-space SGD, should not be read as
significant.

\FloatBarrier
\subsubsection{Gauss--Newton Methods}
\label{sec:mini-batch-gauss-newton}

We now repeat the Gauss--Newton study of
Section~\ref{sec:full-batch-gauss-newton} with minibatches. The method is
the DAS-restricted Gauss--Newton step \eqref{eq:das-gauss-newton} analyzed
there, and we state only what changes. At iteration \(t\) a minibatch
\(\mathcal{B}_t\) containing half of the \(N=1500\) interior collocation
points, \(|\mathcal{B}_t|=750\), is drawn without replacement. The
Jacobian of the residuals on the minibatch,
\(J_{\mathcal{B}_t}\in\mathbb{R}^{|\mathcal{B}_t|\times p}\), defines
the minibatch Gauss--Newton matrix
\(J_{\mathcal{B}_t}^\top J_{\mathcal{B}_t}\) and the gradient
\(g_t=J_{\mathcal{B}_t}^\top r_{\mathcal{B}_t}\) of the minibatch
objective \(\mathcal{L}_{\mathcal{B}_t}\); throughout this section
\(g_t\) denotes this minibatch gradient, and
\(\nabla\mathcal{L}(\theta_t)\) the gradient of the objective on all
\(N\) points.

Both the basis and the step now come from the minibatch. The DAS basis
\(V_t\) is built from the minibatch Gauss--Newton matrix, through the
right singular vectors of \(J_{\mathcal{B}_t}\), and the coverage
condition \eqref{eq:das-gradient-coverage} is imposed on a selection
gradient \(\hat g_t\):
\begin{equation}
\left\lVert
P_t\hat g_t
\right\rVert_2^2
\geq
\tau
\left\lVert
\hat g_t
\right\rVert_2^2 ,
\label{eq:mb-das-gradient-coverage}
\end{equation}
with the same fallback to all retained eigenvectors when they cannot reach
this coverage. We compare two selection gradients: the minibatch gradient,
\(\hat g_t=g_t\) (\emph{batch selection}), and the full-batch gradient,
\(\hat g_t=\nabla\mathcal{L}(\theta_t)\) (\emph{full selection}). They
differ in whether the coverage can be met. The minibatch gradient lies in
the row space of \(J_{\mathcal{B}_t}\), which the retained eigenvectors
span up to the numerical floor of Section~\ref{sec:conditioning}, so under
batch selection the coverage can fail only through that floor. The
full-batch gradient also depends on the residuals outside the minibatch and
in general has a component in the null space of \(J_{\mathcal{B}_t}\),
whose dimension is at least \(p-|\mathcal{B}_t|\); under full selection
the coverage can therefore genuinely fail, and the fallback then applies.
The step is \(p_t=V_ty_t\), where \(y_t\in\mathbb{R}^{k_t}\) solves
\[
\left(
V_t^\top J_{\mathcal{B}_t}^\top J_{\mathcal{B}_t}V_t+\mu_tI
\right)y_t
=
-V_t^\top g_t .
\]
Because \(V_t\) consists of right singular vectors of
\(J_{\mathcal{B}_t}\), the reduced matrix
\(V_t^\top J_{\mathcal{B}_t}^\top J_{\mathcal{B}_t}V_t\) is diagonal,
with the selected squared singular values on its diagonal, as in the
full-batch case.

The properties established in Section~\ref{sec:full-batch-gauss-newton}
carry over with \(J_{\mathcal{B}_t}\) and \(\mathcal{L}_{\mathcal{B}_t}\) in
place of \(J_t\) and \(\mathcal{L}\). The direction lies in
\(\operatorname{span}(V_t)\); the argument of \eqref{eq:reduced-gn-descent}
makes it a strict descent direction for the minibatch objective whenever the
reduced matrix above is positive definite and \(V_t^\top g_t\neq0\); and the
identity \eqref{eq:full-and-projected-derivative}, which uses only
\(p_t=P_tp_t\), holds for the minibatch gradient. Two points differ. With
batch selection, \eqref{eq:mb-das-gradient-coverage} guarantees
\(V_t^\top g_t\neq0\) whenever \(g_t\neq0\), whereas with full selection
this is not implied by the coverage of the full-batch gradient and is an
assumption. And the stationarity implication
\(P_t\hat g_t=0\Rightarrow\hat g_t=0\) now concerns the selection gradient,
and holds only in the iterations where the coverage is met: it concerns the
full objective under full selection, and the minibatch objective under
batch selection.

As in the full-batch experiments, the basis is recomputed at every
iteration, here from a singular value decomposition of the
\(750\times p\) minibatch Jacobian that includes a complete set of right
singular vectors; the minibatch lowers the cost of forming the basis
relative to the full set of points but does not remove it.

Globalization follows Section~\ref{sec:full-batch-gauss-newton} with one
change. The sufficient-decrease and curvature conditions of the
backtracking and Wolfe line searches are evaluated on the minibatch
objective \(\mathcal{L}_{\mathcal{B}_t}\) with the projected gradient, which
by \eqref{eq:full-and-projected-derivative} gives the same directional
derivatives as the unprojected minibatch gradient. Likewise, the
Levenberg--Marquardt runs use the gain ratio \eqref{eq:gain-ratio} with
\(\mathcal{L}_{\mathcal{B}_t}\) in place of \(\mathcal{L}\) and with the
decrease predicted by the reduced minibatch model,
\[
m_t(0)-m_t(p_t)
=
-(V_t^\top g_t)^\top y_t
-\tfrac12\left\lVert J_{\mathcal{B}_t}V_ty_t\right\rVert_2^2 .
\]
The concern raised in Section~\ref{sec:full-batch-gauss-newton}, that a
damping value calibrated on \(\operatorname{span}(V_t)\) may carry little
information about \(\operatorname{span}(V_{t+1})\), applies with more force
here, because the reduced model now also changes with the minibatch. The
experiments below do not show such a penalty. We include DAS-LM runs for
both selection gradients and both coverage levels; trust-region
globalization is again left for future work.

\begin{figure}[!htbp]
    \centering
    \includegraphics[width=\linewidth]{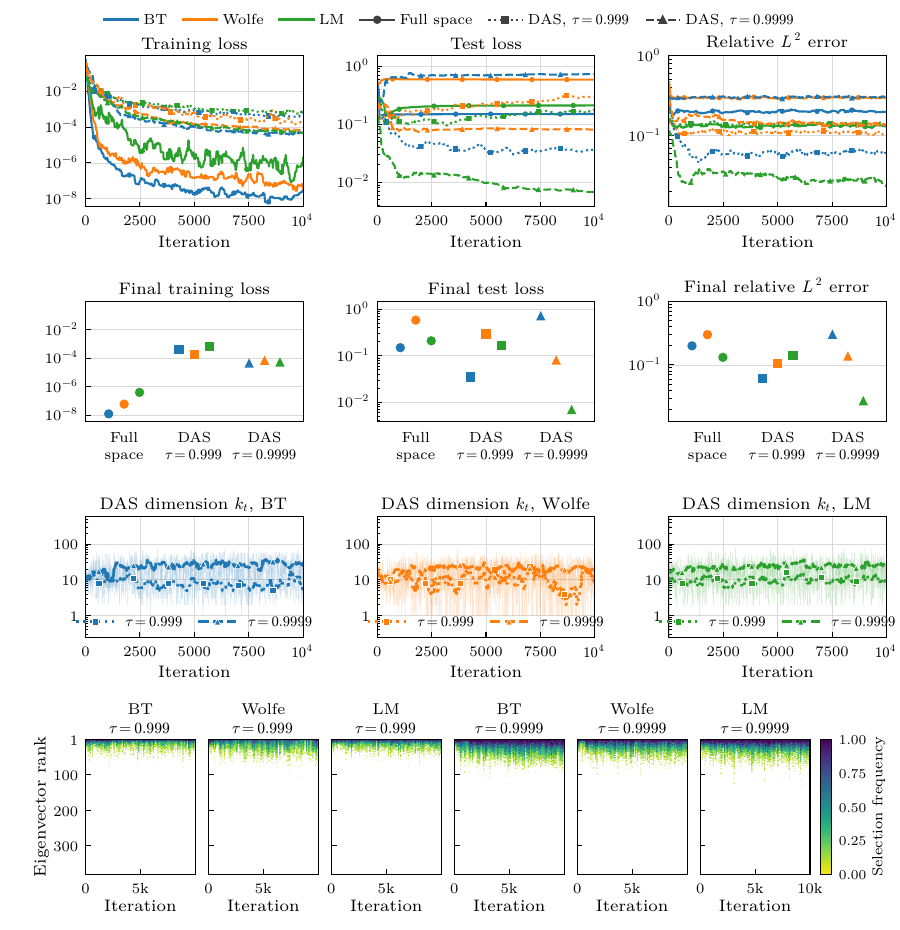}
    \caption{Minibatch Gauss--Newton optimization of Burgers, batch
    selection. First row: training loss on all \(N\) points, test loss, and
    relative \(L^2\) error versus iteration (running median over 250
    iterations) for full-space Gauss--Newton (solid) and DAS-GN with
    \(\tau=0.999\) (dotted, squares) and \(\tau=0.9999\) (dashed,
    triangles); color gives the globalization: backtracking (BT), Wolfe, or
    Levenberg--Marquardt (LM). Second row: the medians of the same three
    quantities over the last \(1000\) iterations, one marker per run. Third
    row: DAS dimension \(k_t\) (thin lines: logged values; thick lines:
    running median over 250 iterations). Fourth row: how often each
    eigenvector of the minibatch Gauss--Newton matrix is selected, per
    window of 100 iterations, with eigenvectors ranked by eigenvalue (rank 1
    is the largest; white: never selected).}
    \label{fig:burgers_loss_batch}
\end{figure}

\begin{figure}[!htbp]
    \centering
    \includegraphics[width=\linewidth]{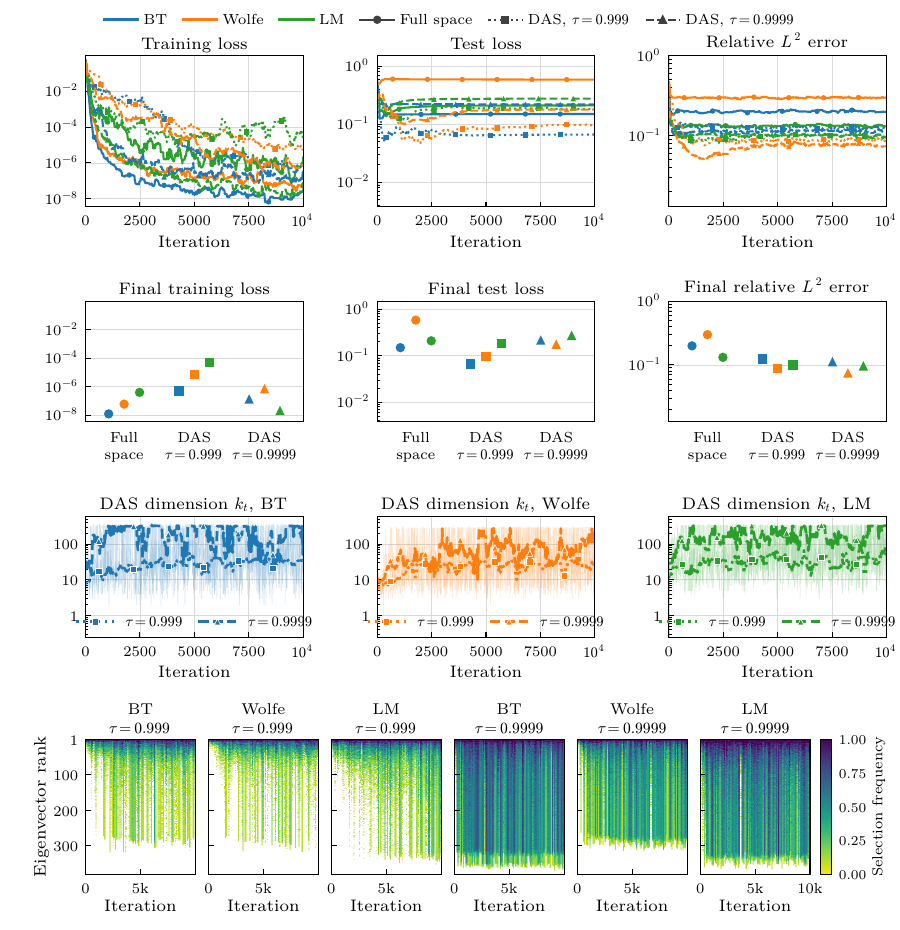}
    \caption{Minibatch Gauss--Newton optimization of Burgers, full
    selection. Rows, line styles, and colors as in
    Figure~\ref{fig:burgers_loss_batch}; the full-space runs are the same.}
    \label{fig:burgers_loss_full}
\end{figure}

\begin{figure}[!htbp]
    \centering
    \includegraphics[width=1\linewidth]{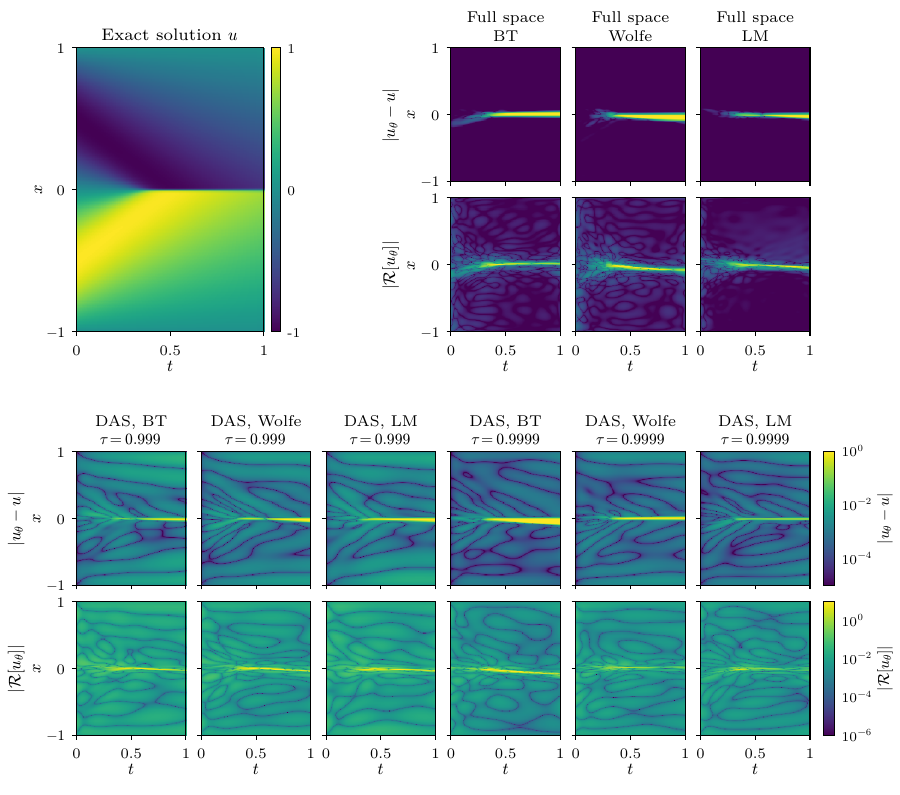}
    \caption{Burgers solutions after minibatch Gauss--Newton training,
    batch selection. Top left: the exact solution. Top right: full-space
    Gauss--Newton. Bottom: DAS-GN. In each block the upper row is the
    pointwise error \(|u_\theta-u|\) and the lower row the pointwise PDE
    residual \(|\mathcal{R}[u_\theta]|\), for the globalization named in the
    panel title; both color scales are logarithmic and shared by all runs,
    so panels are directly comparable.}
    \label{fig:burgers_sol_batch}
\end{figure}

\begin{figure}[!htbp]
    \centering
    \includegraphics[width=1\linewidth]{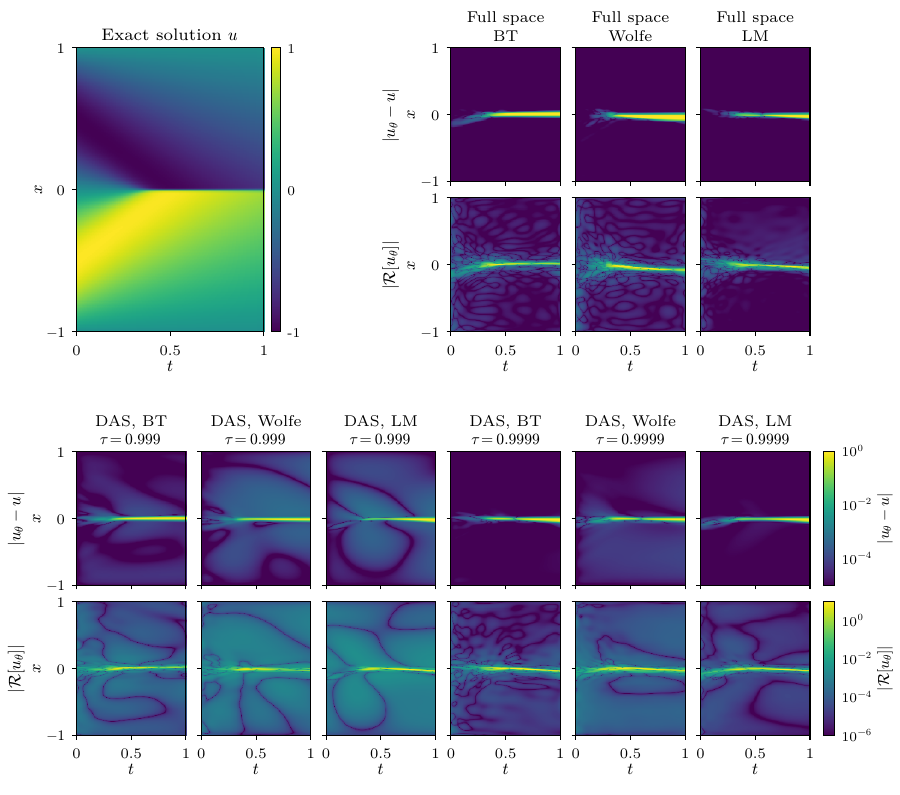}
    \caption{Burgers solutions after minibatch Gauss--Newton training, full
    selection. Layout and color scales as in
    Figure~\ref{fig:burgers_sol_batch}; the full-space runs are the same.}
    \label{fig:burgers_sol_full}
\end{figure}

Figures~\ref{fig:burgers_loss_batch}--\ref{fig:burgers_sol_full} and
Table~\ref{tab:burgers-gn} summarize the minibatch Gauss--Newton experiments
on the scaled Burgers benchmark (\(p=6391\) parameters, hard-constraint
ansatz) for coverage \(\tau=0.999\) and \(\tau=0.9999\). Because the
minibatch objective fluctuates from one iteration to the next, we report the
training loss on all \(N\) points and summarize each run by the median of a
quantity over its last \(1000\) iterations.

The outcome of DAS-GN on this benchmark is governed by the selection
gradient (Figures~\ref{fig:burgers_loss_batch}
and~\ref{fig:burgers_loss_full}, first two rows). With batch selection the
training loss levels off at \(1.9\times10^{-4}\)--\(6.3\times10^{-4}\) for
\(\tau=0.999\) and at \(4.6\times10^{-5}\)--\(7.3\times10^{-5}\) for
\(\tau=0.9999\), three to four orders of magnitude above the full-space runs
(\(1.3\times10^{-8}\)--\(4.2\times10^{-7}\)), and never falls below
\(10^{-5}\). With full selection it reaches
\(5.3\times10^{-7}\)--\(5.2\times10^{-5}\) for \(\tau=0.999\) and
\(2.3\times10^{-8}\)--\(7.4\times10^{-7}\) for \(\tau=0.9999\); the latter
are within a factor of \(2\)--\(60\) of the best full-space run. The
iteration counts follow the same pattern: full-space Gauss--Newton brings
the training loss below \(10^{-3}\) within \(160\)--\(330\) iterations and
below \(10^{-6}\) within \(790\)--\(1680\); DAS-GN needs
\(210\)--\(4230\) iterations for the first threshold, and reaches the second
only with full selection (\(1250\)--\(8840\) iterations in five of the six
runs).

The two selection gradients also produce very different subspaces (third
and fourth rows of the same figures). With batch selection the coverage \(\tau\)
is reached at every logged iteration and the subspace stays small: the
median \(k_t\) is \(8\)--\(25\), the mean \(10.6\)--\(27.8\)
(\(0.17\)--\(0.43\%\) of \(p\)), and \(k_t\) never exceeds \(82\). The
selected directions are the leading ones: \(94\)--\(100\%\) of all
selections fall among the \(50\) eigenvectors with the largest eigenvalues.
With full selection the retained eigenvectors cannot reach the coverage in
\(2.9\)--\(5.6\%\) of the logged iterations for \(\tau=0.999\) and in
\(32\)--\(48\%\) for \(\tau=0.9999\): part of the full-batch gradient then
lies outside the retained eigenvectors, in the null space of the minibatch
Jacobian or below the numerical floor. In those iterations all
retained eigenvectors are used, so that \(k_t\) jumps to about
\(280\)--\(370\) and the selection extends down to the floor; over a whole
trajectory \(9\)--\(36\%\) of all selections lie beyond the \(200\) leading
eigenvectors. The mean \(k_t\) rises accordingly, to \(38\)--\(57\) for
\(\tau=0.999\) and \(117\)--\(187\) for \(\tau=0.9999\)
(\(0.6\)--\(2.9\%\) of \(p\)). At \(\tau=0.9999\), full selection therefore
acts in a third to a half of the iterations as a Gauss--Newton step on the
whole numerically retained subspace, which is consistent with its training
loss approaching that of the full-space runs.

The test loss and the relative \(L^2\) error do not follow the training
loss, and Figures~\ref{fig:burgers_sol_batch}
and~\ref{fig:burgers_sol_full} show why. In every run the error is
concentrated at the shock: more than \(93\%\) of the squared error lies in
the band \(|x|\le0.05\), where the pointwise error reaches
\(0.3\)--\(1.9\). Away from the shock the error does follow the training
loss: its median outside the band is
\(3\times10^{-7}\)--\(8\times10^{-7}\) for the full-space runs,
\(3\times10^{-6}\)--\(8\times10^{-5}\) for DAS-GN with full selection, and
\(4\times10^{-4}\)--\(2\times10^{-3}\) with batch selection. The \(N=1500\)
fixed collocation points do not resolve the shock, so driving the training
loss down fits the equation everywhere except there. The relative \(L^2\)
errors (\(2.8\times10^{-2}\)--\(3.0\times10^{-1}\)) and the test losses
(\(7.0\times10^{-3}\)--\(7.2\times10^{-1}\)) are therefore set by how each
run happens to place the shock, not by how far it reduces the training
loss. They settle early: in \(14\) of the \(15\) runs the test loss at the
end of training is \(1.4\)--\(5.7\) times higher than the lowest value
reached earlier, in most cases during the first \(1100\) iterations. Nine of
the twelve DAS runs end with a smaller relative \(L^2\) error than the best
full-space run (\(1.3\times10^{-1}\)), and the smallest error overall
belongs to DAS-LM with batch selection and \(\tau=0.9999\)
(\(2.8\times10^{-2}\), test loss \(7.0\times10^{-3}\)), whose training loss
is \(5.5\times10^{-5}\). Runs that share a method differ by as much as a
factor of five (batch selection with backtracking: \(6.1\times10^{-2}\) for
\(\tau=0.999\) against \(3.0\times10^{-1}\) for \(\tau=0.9999\)), and each
configuration was run once, so these differences should not be read as a
ranking of the methods by accuracy.

Table~\ref{tab:burgers-gn-rank-savings} quantifies the reduction. With batch
selection the DAS uses on average \(11\)--\(28\) directions, so more than
\(99.5\%\) of the \(p=6391\) parameter directions, and more than \(96\%\) of
the at most \(750\) directions spanned by the minibatch Gauss--Newton
matrix, are left
untouched. The reduced Gauss--Newton matrix then has on average
\(4\times10^{4}\)--\(2\times10^{5}\) times fewer entries than the
\(p\times p\) matrix, and a dense solve, whose cost grows cubically with the
dimension, is on average \(6\times10^{6}\)--\(6\times10^{7}\) times cheaper
per iteration. With full selection the corresponding factors are
\(700\)--\(9500\) and \(1\times10^{4}\)--\(3\times10^{5}\). This outweighs
the larger iteration count: summing \(k_t^3\) over the iterations needed to
bring the training loss below \(10^{-3}\) (estimated from the logged values)
gives \(2\times10^{4}\)--\(6\times10^{6}\) times less solve work than the
corresponding sum of \(p^3\) for full-space Gauss--Newton with the same
globalization. These figures concern the reduced solve only. Because every
DAS iteration also forms the singular value decomposition of the
minibatch Jacobian, with a complete set of \(p\) right singular vectors,
to build the basis, they do not translate into shorter run
times: the DAS runs took \(2.6\)--\(3.1\) seconds per iteration against
about \(0.9\) for the full-space runs. Realizing the savings requires a
cheaper construction of \(V_t\).

The LM runs do not show the penalty anticipated above. In the full space LM
ends with the highest training loss of the three globalizations
(\(4.2\times10^{-7}\), against \(1.3\times10^{-8}\) and
\(6.3\times10^{-8}\)). In the DAS at \(\tau=0.9999\) it is competitive with
the line searches. With full selection DAS-LM reaches the lowest
training loss of all DAS runs, \(2.3\times10^{-8}\), which is \(6\) times
lower than DAS-GN with backtracking, \(19\) times lower than full-space LM,
and within a factor of two of the best full-space run. With batch selection
its training loss (\(5.5\times10^{-5}\)) lies between those of the two line
searches, and it has the smallest relative \(L^2\) error of the study. At
\(\tau=0.999\) LM ends with the highest training loss of the three DAS
globalizations: by a factor of \(1.6\)--\(3.4\) with batch selection, and by
a factor of \(7\)--\(98\) with full selection, where its training loss is
also not monotone over the second half of the run. Along the four DAS-LM
trajectories \(k_t\) ranges from \(1\) to \(82\) with batch selection and
from \(3\) to \(364\) with full selection, changes by as much as \(349\)
between consecutive logged iterations, and the ratio of the largest to the
smallest selected eigenvalue ranges from \(1\) to \(10^{14}\). Consecutive
reduced models are thus far from comparable, yet the carried-over damping
does not systematically harm the method: the ratio of the DAS training loss
to that of the full-space run with the same globalization is smaller for LM
than for both line searches in three of the four configurations, in part
because full-space LM is itself the weakest of the full-space runs.

\begin{table}[!htbp]
\centering
\footnotesize
\setlength{\tabcolsep}{2.7pt}
\begin{tabular}{llllcccccrr}
\toprule
 & & & & \multicolumn{2}{c}{\(k_t\)} & Training & Test & Relative & \multicolumn{2}{c}{Iterations to loss} \\
Method & Selection & Global. & \(\tau\) & median & mean & loss & loss & \(L^2\) error & \(<10^{-3}\) & \(<10^{-6}\) \\
\midrule
Full-space & --- & BT & --- & --- & --- & \(1.3\times10^{-8}\) & \(1.5\times10^{-1}\) & \(2.0\times10^{-1}\) & 160 & 790 \\
Full-space & --- & Wolfe & --- & --- & --- & \(6.3\times10^{-8}\) & \(5.8\times10^{-1}\) & \(3.0\times10^{-1}\) & 260 & 790 \\
Full-space & --- & LM & --- & --- & --- & \(4.2\times10^{-7}\) & \(2.1\times10^{-1}\) & \(1.3\times10^{-1}\) & 330 & 1680 \\
\midrule
DAS & batch & BT & 0.999 & 8 & 10.6 & \(4.0\times10^{-4}\) & \(3.5\times10^{-2}\) & \(6.1\times10^{-2}\) & 2650 & --- \\
DAS & batch & Wolfe & 0.999 & 10 & 13.6 & \(1.9\times10^{-4}\) & \(2.9\times10^{-1}\) & \(1.1\times10^{-1}\) & 1780 & --- \\
DAS & batch & LM & 0.999 & 10 & 11.8 & \(6.3\times10^{-4}\) & \(1.7\times10^{-1}\) & \(1.4\times10^{-1}\) & 4230 & --- \\
\midrule
DAS & batch & BT & 0.9999 & 25 & 27.8 & \(4.6\times10^{-5}\) & \(7.2\times10^{-1}\) & \(3.0\times10^{-1}\) & 1020 & --- \\
DAS & batch & Wolfe & 0.9999 & 20 & 21.5 & \(7.3\times10^{-5}\) & \(8.2\times10^{-2}\) & \(1.4\times10^{-1}\) & 1800 & --- \\
DAS & batch & LM & 0.9999 & 23 & 25.7 & \(5.5\times10^{-5}\) & \(7.0\times10^{-3}\) & \(2.8\times10^{-2}\) & 1410 & --- \\
\midrule
DAS & full & BT & 0.999 & 26 & 48.6 & \(5.3\times10^{-7}\) & \(6.6\times10^{-2}\) & \(1.2\times10^{-1}\) & 1130 & 5020 \\
DAS & full & Wolfe & 0.999 & 23 & 37.7 & \(7.8\times10^{-6}\) & \(9.7\times10^{-2}\) & \(8.9\times10^{-2}\) & 1470 & 8840 \\
DAS & full & LM & 0.999 & 29 & 57.3 & \(5.2\times10^{-5}\) & \(1.8\times10^{-1}\) & \(9.9\times10^{-2}\) & 590 & --- \\
\midrule
DAS & full & BT & 0.9999 & 193 & 186.5 & \(1.4\times10^{-7}\) & \(2.2\times10^{-1}\) & \(1.1\times10^{-1}\) & 220 & 1250 \\
DAS & full & Wolfe & 0.9999 & 51 & 117.4 & \(7.4\times10^{-7}\) & \(1.8\times10^{-1}\) & \(7.5\times10^{-2}\) & 1270 & 5470 \\
DAS & full & LM & 0.9999 & 152 & 179.6 & \(2.3\times10^{-8}\) & \(2.8\times10^{-1}\) & \(9.7\times10^{-2}\) & 210 & 1250 \\
\bottomrule
\end{tabular}
\caption{Burgers, minibatch Gauss--Newton (\(750\) of the \(N=1500\) interior
points per iteration, \(10^4\) iterations). Selection is the gradient used
to choose the DAS: of the minibatch (batch) or of all \(N\) points (full).
Training loss (on all \(N\) points), test loss, and relative \(L^2\) error
are medians over the last \(1000\) iterations; \(k_t\) is summarized over
the whole trajectory. The last two columns give the first logged iteration
at which the training loss falls below the threshold (logged every ten
iterations; ---: never). The network has \(p=6391\) parameters.}
\label{tab:burgers-gn}

\vspace{1em}

\centering
\footnotesize
\setlength{\tabcolsep}{5pt}
\begin{tabular}{lllcccccc}
\toprule
 & & & & \multicolumn{3}{c}{Directions saved} & Matrix & Dense \\
\cmidrule(lr){5-7}
Selection & Global. & \(\tau\) & Mean \(k_t\) & mean & median & worst & entries saved & solve saved \\
\midrule
batch & BT & 0.999 & 10.6 & 98.6\% & 98.9\% & 93.3\% & 99.97\% & 99.999\% \\
batch & Wolfe & 0.999 & 13.6 & 98.2\% & 98.7\% & 92.9\% & 99.94\% & 99.998\% \\
batch & LM & 0.999 & 11.8 & 98.4\% & 98.7\% & 94.0\% & 99.96\% & 99.999\% \\
\midrule
batch & BT & 0.9999 & 27.8 & 96.3\% & 96.7\% & 90.1\% & 99.82\% & 99.990\% \\
batch & Wolfe & 0.9999 & 21.5 & 97.1\% & 97.3\% & 90.8\% & 99.89\% & 99.995\% \\
batch & LM & 0.9999 & 25.7 & 96.6\% & 96.9\% & 89.1\% & 99.85\% & 99.993\% \\
\midrule
full & BT & 0.999 & 48.6 & 93.5\% & 96.5\% & 57.9\% & 98.77\% & 99.628\% \\
full & Wolfe & 0.999 & 37.7 & 95.0\% & 96.9\% & 58.1\% & 99.23\% & 99.783\% \\
full & LM & 0.999 & 57.3 & 92.4\% & 96.1\% & 53.9\% & 98.43\% & 99.490\% \\
\midrule
full & BT & 0.9999 & 186.5 & 75.1\% & 74.3\% & 51.1\% & 89.96\% & 95.643\% \\
full & Wolfe & 0.9999 & 117.4 & 84.3\% & 93.2\% & 58.4\% & 94.97\% & 98.177\% \\
full & LM & 0.9999 & 179.6 & 76.1\% & 79.7\% & 51.5\% & 91.03\% & 96.247\% \\
\midrule
\multicolumn{3}{l}{Batch selection pooled} & 18.5 & 97.5\% & 97.9\% & 89.1\% & 99.91\% & 99.996\% \\
\multicolumn{3}{l}{Full selection pooled} & 104.5 & 86.1\% & 94.8\% & 51.1\% & 95.40\% & 98.161\% \\
\multicolumn{3}{l}{All runs pooled} & 61.5 & 91.8\% & 97.1\% & 51.1\% & 97.65\% & 99.078\% \\
\bottomrule
\end{tabular}
\caption{Savings of DAS-GN on Burgers relative to the largest possible rank
of the minibatch Gauss--Newton matrix from which the basis is built,
\(\operatorname{rank}(J_{\mathcal{B}_t}^\top J_{\mathcal{B}_t})\le|\mathcal{B}_t|=750\).
Directions saved is
\(1-k_t/750\), at the mean, median, and largest \(k_t\) of each trajectory;
matrix entries and dense solve cost compare the trajectory averages of
\(k_t^2\) and \(k_t^3\) with \(750^2\) and \(750^3\).}
\label{tab:burgers-gn-rank-savings}
\end{table}

\FloatBarrier
\section{Discussion}
\label{sec:discussion}

\subsection{Scope of the Contribution}
\label{sec:discussion-scope}

This work introduces the dominant aligned subspace (DAS) as a tool for
studying, and potentially exploiting, the concentration of gradient energy in
a small subset of Hessian eigendirections. Our goal is not to claim that the
current implementation is already an efficient large-scale optimizer for
first-order or Gauss--Newton methods. Instead, the primary contribution
is to identify a structural distinction that is often obscured in
curvature-based optimization: directions ranked by Hessian eigenvalue
magnitude need not be the same directions that explain the optimization
gradient.

This distinction is increasingly relevant as machine-learning models continue
to grow in parameter count and computational cost. Large-scale curvature
analysis, repeated Hessian-vector products, and second-order optimization are
often inaccessible without substantial computational resources. In this
context, DAS provides a possible route toward lower-dimensional
curvature-aware optimization. If a small, dynamically selected subspace can
retain most of the optimization-relevant gradient signal, then some
computationally expensive operations may be carried out in that subspace
rather than in the full space.

The experiments in this paper should therefore be interpreted as a proof of
concept. They show that, on the considered PINN benchmarks, the gradient
energy can often be concentrated in a small adaptive subspace of Hessian
eigendirections. The resulting DAS dimension changes throughout training:
the subspace expands when gradient energy is spread across many directions and
contracts when the energy is more concentrated. This behavior differs from a
fixed top-\(k\) Hessian eigenspace, which selects directions only according to
curvature magnitude.

\subsection{Interpretation of the Results}
\label{sec:discussion-interpretation}

The question this paper asks is whether optimization of a PINN can be
confined, at every iteration, to a subspace far smaller than the parameter
space without losing the solution. The experiments answer it
affirmatively: the DAS is such a subspace. On Poisson-2D, DAS-GN with a
line search reaches relative \(L^2\) errors of
\(0.8\times10^{-6}\)--\(2.3\times10^{-6}\), against
\(3.8\times10^{-7}\)--\(1.8\times10^{-6}\) in the full space, with a median
of \(10\)--\(16\) directions out of \(p=1981\)
(Table~\ref{tab:poisson2d-gn}). On wave-1D, DAS-GD with a line search ends
within \(6\)--\(21\%\) of the full-space relative \(L^2\) error
(Table~\ref{tab:wave1d-sgd-error-norms}). On the scaled Burgers benchmark
the DAS runs are at least as accurate as the full-space ones
(Table~\ref{tab:burgers-gn}): the full-space errors are
\(1.3\times10^{-1}\)--\(3.0\times10^{-1}\), all six runs with full-gradient
selection (\texttt{projF}) lie between \(7.5\times10^{-2}\) and
\(1.2\times10^{-1}\), and the runs with mini-batch selection
(\texttt{projB}) between \(2.8\times10^{-2}\) and \(3.0\times10^{-1}\). The
smallest of these subspaces, \texttt{projB}, uses \(11\)--\(28\) directions
on average, between \(0.2\) and \(0.4\%\) of the \(p=6391\) parameters, and
is no less accurate than the full space. In each case more than \(97\%\)
of the parameter directions are left untouched at every iteration, and the
solution is matched.

The training and test losses are secondary to this question, but they show
how the match is obtained. The DAS runs end at a higher training loss and
take more iterations: on Poisson-2D the DAS-GN training loss ends
\(4\)--\(22\) times above the full-space line-search runs, and on Burgers
\texttt{projF} at \(\tau=0.9999\) comes within a factor of \(2\)--\(60\) of
the best full-space training loss while \texttt{projB} levels off three to
four orders of magnitude above it. This gap does not carry over to the
solution. The \texttt{projB} runs, with the smallest subspaces and the
highest training losses, are as accurate as the full-space runs, and the
most accurate run of the study, DAS-LM with \texttt{projB} and
\(\tau=0.9999\), has a training loss more than three orders of magnitude
above that of the best-optimized full-space run. The directions the DAS
leaves out are needed to drive the sampled residual to its floor, but not
to reach the solution: beyond a certain point, lowering the training loss
fits the collocation set without improving the solution error.

The scope of this conclusion should be stated precisely. The DAS is not a
single fixed subspace; it is rebuilt at every iteration, and what the
experiments establish is that a small subspace suffices at each step. The
match holds for the line-search methods on all three benchmarks and for
Levenberg--Marquardt on Burgers; on Poisson-2D, full-space
Levenberg--Marquardt remains \(24\)--\(33\) times more accurate than
DAS-LM. On Burgers the match is at a modest accuracy, since no run,
full-space or DAS, resolves the shock, where more than \(93\%\) of the
squared error is concentrated
(Figures~\ref{fig:burgers_sol_batch} and~\ref{fig:burgers_sol_full}); the
experiments therefore show that the restriction does not degrade the
solution, and with one run per configuration they do not rank the DAS
variants against each other. A possible explanation for the match is that
the DAS acts as a spectral filter, discarding directions that carry little
of the descent signal; testing this, and the result itself, across more
architectures, PDEs, and random seeds is left for future work.

\subsection{Limitations}
\label{sec:discussion-limitations}

The primary limitation of the current implementation is scalability. The
experiments use relatively small neural networks and PDE benchmarks because
the DAS construction relies on full Hessian information. Explicitly forming a
dense Hessian and computing its eigendecomposition at every iteration is not
feasible for modern large-scale neural networks.

Although the selected DAS subspace can be much smaller than the ambient
parameter space, this does not by itself imply lower wall-clock cost. The
computational cost of constructing the basis, computing curvature
information, and refreshing the subspace must be included in any practical
comparison. In particular, a reduced \(k_t \times k_t\) Gauss--Newton solve
may be inexpensive when \(k_t \ll p\), but the cost of obtaining the DAS basis
and forming the reduced curvature operator can still dominate the total
runtime.

A second limitation is that DAS currently selects individual Hessian
eigenvectors. This choice is natural for analysis, but individual
eigenvectors can be numerically unstable when eigenvalues are clustered or
nearly degenerate. In such cases, the invariant subspace associated with a
group of eigenvalues may be stable even though the individual eigenvectors
are not. Consequently, future versions of DAS may need to select spectral
bands, invariant subspaces, or general gradient-informed bases rather than
individual eigendirections.

Finally, the present work does not prove that the DAS-selected subspace is the
smallest useful optimization subspace. The energy threshold guarantees that
the selected projector satisfies
\[
\left\lVert
P_tg_t
\right\rVert_2^2
\geq
\tau
\left\lVert
g_t
\right\rVert_2^2,
\]
but this criterion does not distinguish useful gradient components from
components dominated by stochastic noise. It is possible that a smaller
subspace could yield better optimization behavior if it excludes directions
that are poorly conditioned, noisy, or weakly relevant to long-term descent.
The spurious-alignment collapse of Appendix~\ref{app:spurious} is an
instance of this, and it is why a better selection criterion is needed.

\subsection{Scalable Approximations}
\label{sec:discussion-scalability}

A central direction for future work is to replace explicit Hessian formation
and full eigendecomposition with matrix-free approximations
\citep{saad2003iterative}. The first-order
experiments in this paper use the full dense Hessian, of the full
training objective in the full-batch experiments and of the current
mini-batch in the mini-batch experiment. The Gauss--Newton
experiments already build the DAS from the eigenvectors of the
Gauss--Newton (Gramian) matrix \(J_t^\top J_t\), which shares the
Hessian's dominant spectral structure near a small-residual solution and is
cheaper to form, and on Burgers from its minibatch version alone; doing the
same for first-order methods is a first step toward lower cost. Even there,
however, the basis comes from a full singular value decomposition at every
iteration. Krylov methods are particularly promising, because they build
low-dimensional subspaces from Hessian-vector products without
materializing the Hessian. They still assume that a basis exists in that
representation, and even with a gradient seed they tend to favor the
leading eigenmodes over the aligned ones.

For a symmetric Hessian \(H_t\) and starting vector \(q_t\), a Krylov subspace
of order \(m\) is
\[
\mathcal{K}_m(H_t,q_t)
=
\operatorname{span}
\left\{
q_t,
H_tq_t,
H_t^2q_t,
\ldots,
H_t^{m-1}q_t
\right\}.
\]
When \(q_t=g_t\), the resulting subspace is directly seeded by the current
gradient:
\[
\mathcal{K}_m(H_t,g_t)
=
\operatorname{span}
\left\{
g_t,
H_tg_t,
H_t^2g_t,
\ldots,
H_t^{m-1}g_t
\right\}.
\]
This construction may provide an efficient approximation to a
gradient-aligned curvature subspace without requiring the full Hessian
eigenbasis; it underlies Krylov subspace descent \citep{vinyals2012krylov}
and, in randomized form, subspace Newton methods \citep{gower2019rsn}. It is
also the construction followed by \citet{hoang2026adaptive,hoang2026subspace}; given where the selected
eigenvectors lie in our experiments (at both ends of the spectrum), we
expect that the Krylov construction may not be be the best for PINNs because of its extreme ill conditioning and Krylov might struggle to provide a truley seeded basis without eigenvalue bias.

Block Krylov methods provide another possible extension. Rather than starting
from a single vector, one may use a block containing the current gradient,
previous gradients, stochastic gradients from multiple mini-batches, or
approximate Newton directions. Such a basis may be more robust to mini-batch
noise and may better capture directions that remain relevant across multiple
optimization steps.

An extended Krylov construction \citep{druskin1998extended} may be useful
when both high- and low-curvature directions are important. For example, a regularized expanded
basis could include both Hessian and inverse-Hessian actions:
\[
\mathcal{K}^{\mathrm{exp}}_m(H_t,g_t)
=
\operatorname{span}
\left\{
g_t,\,
H_tg_t,\,
H_t^2g_t,\,
(H_t+\lambda I)^{-1}g_t,\,
(H_t+\lambda I)^{-2}g_t,\,
\ldots
\right\}.
\]
Such spaces may capture both stiff directions and weak-curvature directions
without requiring an explicit eigenvalue decomposition. A key question is
whether these approximate bases preserve the central DAS property:
\[
\left\lVert
P_{\mathcal{S}_t}g_t
\right\rVert_2^2
\geq
\tau
\left\lVert
g_t
\right\rVert_2^2,
\]
where \(P_{\mathcal{S}_t}\) is the projector onto an efficiently computed
approximate subspace \(\mathcal{S}_t\).

\subsection{Open Questions}
\label{sec:discussion-open-questions}

Several questions must be addressed before DAS can become a practical
large-scale optimization tool.

\paragraph{Does alignment concentration persist at scale?}
The current experiments use small PINNs biggest being around 9k parameters because full Hessian analysis is
tractable only in modest parameter dimensions. Larger networks may have
different spectral structure, weaker gradient-energy concentration, or
greater sensitivity to stochastic gradients. It remains necessary to test
whether a small aligned subspace continues to capture most of the useful
descent signal in modern large-scale models.

\paragraph{Should high-curvature directions be excluded?}
The dominant Hessian eigendirections may correspond to stiff directions that
limit stable learning rates or introduce undesirable curvature variation. A
possible extension is to exclude a set of extreme high-curvature directions
and construct the DAS only in the remaining complement, in the spirit of
recent methods that treat sharp and flat directions differently
\citep{zakarin2025accelerating,nicolas2025dome}. Such a method would
combine alignment-based selection with explicit spectral stabilization.

\paragraph{Are individual eigenvectors the appropriate objects to select?}
When the Hessian spectrum contains clusters of near-equal eigenvalues,
individual eigenvectors may rotate substantially under small perturbations.
Selecting spectral blocks, invariant subspaces, or Krylov basis vectors may
be more stable than selecting individual eigenvectors.

\paragraph{How small can the selected subspace be?}
DAS selects the smallest exact eigenspace that satisfies the prescribed
energy threshold. However, some retained gradient energy may be
noisy or unhelpful. It may be possible to choose a smaller subspace using
criteria based on predicted decrease, local conditioning, curvature
regularization, or the consistency of directions across several iterations.

\paragraph{How should the threshold be selected?}
The energy threshold \(\tau\) controls a tradeoff between gradient coverage,
DAS dimension, and curvature variation. A fixed threshold such as
\(\tau=0.95\) or \(\tau=0.99\) may not be optimal throughout training. An
adaptive threshold based on line-search behavior, estimated conditioning, or
predicted reduction may yield a more efficient method.

\paragraph{How frequently should the basis be refreshed?}
The full-gradient-selected basis improved several solution metrics in the
mini-batch experiments, but it requires additional computation. Periodically
refreshing the basis, reusing previous bases, or using an exponentially
averaged gradient may reduce this cost while retaining much of the
stabilization benefit.

\paragraph{Does DAS improve generalization?}
Some experiments suggest that DAS can improve solution-level error, possibly
because it filters poorly conditioned or noisy directions. However, this
hypothesis requires systematic validation across random seeds, network
architectures, PDE classes, and non-PINN machine-learning tasks, and
against the fixed-subspace baselines of
\citet{li2018intrinsic,gressmann2020improving,li2022low}.

\subsection{Summary}
\label{sec:discussion-summary}

DAS highlights that the Hessian eigendirections with the largest curvature
need not be the directions that explain the current optimization gradient.
The present experiments provide initial evidence that exploiting this
gradient--Hessian alignment structure can be useful, particularly when the
basis-selection procedure is stabilized against mini-batch noise.

The main challenge is now computational rather than conceptual. A practical
large-scale DAS method must approximate the relevant aligned subspace using
matrix-free curvature information, inexpensive basis updates, and robust
selection rules. Krylov, block-Krylov, expanded-Krylov might work initially but with the degeneracy of the hessian for PINNs wouldn't make a robust optimizer so stochastic
subspace-tracking methods provide promising directions for extending DAS
beyond the small-scale Hessian analyses that could be robust in this context.

\acks{%
A.~AliAhmad gratefully acknowledges financial support from Professor George
Em Karniadakis and thanks Professor Khemraj Shukla for his guidance and
support throughout this work. Both authors acknowledge support from the
U.S. Department of Energy, Office of Science, Advanced Scientific Computing
Research program, through the Scientific Discovery through Advanced
Computing (SciDAC) Institute ``LEADS: LEarning-Accelerated Domain Science.''
A.~AliAhmad also thanks the friends and family who inspired the n-DAS
acronym, which stands for more than n-Dominant Aligned Subspace. This research
was conducted using computational resources and services at the Center for
Computation and Visualization, Brown University (the Oscar cluster). The
authors declare no competing interests.}

\newpage

\bibliography{cite}

@article{alain2019negative,
  title={Negative eigenvalues of the {Hessian} in deep neural networks},
  author={Alain, Guillaume and Le Roux, Nicolas and Manzagol, Pierre-Antoine},
  journal={arXiv preprint arXiv:1902.02366},
  year={2019}
}

@article{raissi2019physics,
  title={Physics-informed neural networks: A deep learning framework for solving forward and inverse problems involving nonlinear partial differential equations},
  author={Raissi, Maziar and Perdikaris, Paris and Karniadakis, George E.},
  journal={Journal of Computational Physics},
  volume={378},
  pages={686--707},
  year={2019},
  doi={10.1016/j.jcp.2018.10.045}
}

@article{karniadakis2021physics,
  title={Physics-informed machine learning},
  author={Karniadakis, George E. and Kevrekidis, Ioannis G. and Lu, Lu and Perdikaris, Paris and Wang, Sifan and Yang, Liu},
  journal={Nature Reviews Physics},
  volume={3},
  number={6},
  pages={422--440},
  year={2021},
  doi={10.1038/s42254-021-00314-5}
}

@article{wang2021pinnfail,
  title={When and why {PINNs} fail to train: A neural tangent kernel perspective},
  author={Wang, Sifan and Teng, Yujun and Perdikaris, Paris},
  journal={Journal of Computational Physics},
  volume={449},
  pages={110768},
  year={2022},
  doi={10.1016/j.jcp.2021.110768}
}

@article{rathore2024challenges,
  title={Challenges in training {PINNs}: A loss landscape perspective},
  author={Rathore, Pratik and Lei, Weimu and Frangella, Zachary and Lu, Lu and Udell, Madeleine},
  journal={arXiv preprint arXiv:2402.01868},
  year={2024},
  url={https://arxiv.org/abs/2402.01868}
}

@article{jnini2026curvature,
  title={Curvature-aware optimization for high-accuracy physics-informed neural networks},
  author={Jnini, Anas and Kiyani, Elham and Shukla, Khemraj and Urb{\'a}n, Jorge F. and Daryakenari, Nazanin Ahmadi and M{\"u}ller, Johannes and Zeinhofer, Marius and Karniadakis, George Em},
  journal={arXiv preprint arXiv:2604.05230},
  year={2026},
  doi={10.48550/arXiv.2604.05230},
  url={https://arxiv.org/abs/2604.05230}
}

@article{cohen2021gradient,
  title={Gradient descent on neural networks typically occurs at the edge of stability},
  author={Cohen, Jeremy M. and Kaur, Simran and Li, Yuanzhi and Kolter, J. Zico and Talwalkar, Ameet},
  journal={arXiv preprint arXiv:2103.00065},
  year={2021},
  url={https://arxiv.org/abs/2103.00065}
}

@article{cohen2022adaptive,
  title={Adaptive gradient methods at the edge of stability},
  author={Cohen, Jeremy M. and Ghorbani, Behrooz and Krishnan, Shankar and Agarwal, Naman and Medapati, Sourabh and Badura, Michal and Suo, Daniel and Cardoze, David E. and Nado, Zachary and Dahl, George E. and Gilmer, Justin},
  journal={arXiv preprint arXiv:2207.14484},
  year={2022},
  url={https://arxiv.org/abs/2207.14484}
}

@inproceedings{song2024edge,
  title={Trajectory alignment: Understanding the edge of stability phenomenon via bifurcation theory},
  author={Song, Minhak and Yun, Chulhee},
  booktitle={Advances in Neural Information Processing Systems},
  volume={36},
  pages={71632--71682},
  year={2023},
  url={https://arxiv.org/abs/2307.04204}
}

@inproceedings{song2025does,
  title={Does {SGD} really happen in tiny subspaces?},
  author={Song, Minhak and Ahn, Kwangjun and Yun, Chulhee},
  booktitle={The Thirteenth International Conference on Learning Representations},
  year={2025}
}

@article{gur2018gradient,
  title={Gradient descent happens in a tiny subspace},
  author={Gur-Ari, Guy and Roberts, Daniel A. and Dyer, Ethan},
  journal={arXiv preprint arXiv:1812.04754},
  year={2018},
  url={https://arxiv.org/abs/1812.04754}
}

@inproceedings{ghorbani2019investigation,
  title={An investigation into neural net optimization via {Hessian} eigenvalue density},
  author={Ghorbani, Behrooz and Krishnan, Shankar and Xiao, Ying},
  booktitle={Proceedings of the 36th International Conference on Machine Learning},
  series={Proceedings of Machine Learning Research},
  volume={97},
  pages={2232--2241},
  year={2019},
  publisher={PMLR},
  url={https://proceedings.mlr.press/v97/ghorbani19b.html}
}

@article{sagun2017empirical,
  title={Empirical analysis of the {Hessian} of over-parametrized neural networks},
  author={Sagun, Levent and Evci, Utku and G{\"u}ney, V. U{\u{g}}ur and Dauphin, Yann and Bottou, L{\'e}on},
  journal={arXiv preprint arXiv:1706.04454},
  year={2017},
  url={https://arxiv.org/abs/1706.04454}
}

@inproceedings{li2018visualizing,
  title={Visualizing the loss landscape of neural nets},
  author={Li, Hao and Xu, Zheng and Taylor, Gavin and Studer, Christoph and Goldstein, Tom},
  booktitle={Advances in Neural Information Processing Systems},
  volume={31},
  year={2018}
}

@article{wilamowski2001ill,
  title={Ill-conditioning in neural network training problems},
  author={Saarinen, Sirpa and Bramley, Randall and Cybenko, George},
  journal={SIAM Journal on Scientific Computing},
  volume={14},
  number={3},
  pages={693--714},
  year={1993}
}

@inproceedings{botev2017practical,
  title={Practical {Gauss--Newton} optimisation for deep learning},
  author={Botev, Aleksandar and Ritter, Hippolyt and Barber, David},
  booktitle={Proceedings of the 34th International Conference on Machine Learning},
  series={Proceedings of Machine Learning Research},
  volume={70},
  pages={557--565},
  year={2017},
  publisher={PMLR}
}

@inproceedings{zhao2024theoretical,
  title={Theoretical characterisation of the {Gauss--Newton} conditioning in neural networks},
  author={Zhao, Jim and Singh, Sidak Pal and Lucchi, Aur{\'e}lien},
  booktitle={Advances in Neural Information Processing Systems},
  volume={37},
  pages={114965--115000},
  year={2024}
}

@inproceedings{kingma2015adam,
  title={{Adam}: A method for stochastic optimization},
  author={Kingma, Diederik P. and Ba, Jimmy},
  booktitle={International Conference on Learning Representations},
  year={2015},
  url={https://arxiv.org/abs/1412.6980}
}

@article{robbins1951stochastic,
  title={A stochastic approximation method},
  author={Robbins, Herbert and Monro, Sutton},
  journal={The Annals of Mathematical Statistics},
  volume={22},
  number={3},
  pages={400--407},
  year={1951},
  doi={10.1214/aoms/1177729586}
}

@book{nocedal2006numerical,
  title={Numerical Optimization},
  author={Nocedal, Jorge and Wright, Stephen J.},
  edition={Second},
  publisher={Springer},
  address={New York, NY},
  year={2006},
  doi={10.1007/978-0-387-40065-5}
}

@article{armijo1966minimization,
  title={Minimization of functions having {Lipschitz} continuous first partial derivatives},
  author={Armijo, Larry},
  journal={Pacific Journal of Mathematics},
  volume={16},
  number={1},
  pages={1--3},
  year={1966},
  doi={10.2140/pjm.1966.16.1}
}

@article{wolfe1969convergence,
  title={Convergence conditions for ascent methods},
  author={Wolfe, Philip},
  journal={SIAM Review},
  volume={11},
  number={2},
  pages={226--235},
  year={1969},
  doi={10.1137/1011036}
}

@inproceedings{karimi2016linear,
  title={Linear convergence of gradient and proximal-gradient methods under the {P}olyak-{\L}ojasiewicz condition},
  author={Karimi, Hamed and Nutini, Julie and Schmidt, Mark},
  booktitle={Machine Learning and Knowledge Discovery in Databases (ECML PKDD 2016)},
  series={Lecture Notes in Computer Science},
  volume={9851},
  pages={795--811},
  year={2016},
  publisher={Springer},
  doi={10.1007/978-3-319-46128-1_50}
}

@article{dembo1982inexact,
  title={Inexact {N}ewton methods},
  author={Dembo, Ron S. and Eisenstat, Stanley C. and Steihaug, Trond},
  journal={SIAM Journal on Numerical Analysis},
  volume={19},
  number={2},
  pages={400--408},
  year={1982},
  doi={10.1137/0719025}
}

@inproceedings{nutini2015coordinate,
  title={Coordinate descent converges faster with the {G}auss-{S}outhwell rule than random selection},
  author={Nutini, Julie and Schmidt, Mark and Laradji, Issam H. and Friedlander, Michael and Koepke, Hoyt},
  booktitle={Proceedings of the 32nd International Conference on Machine Learning},
  series={Proceedings of Machine Learning Research},
  volume={37},
  pages={1632--1641},
  year={2015},
  publisher={PMLR}
}

@article{nutini2022let,
  title={Let's make block coordinate descent converge faster: Faster greedy rules, message-passing, active-set complexity, and superlinear convergence},
  author={Nutini, Julie and Laradji, Issam and Schmidt, Mark},
  journal={Journal of Machine Learning Research},
  volume={23},
  number={131},
  pages={1--74},
  year={2022},
  url={https://jmlr.org/papers/v23/18-541.html}
}

@article{hoang2026subspace,
  title={Subspace {Levenberg--Marquardt} algorithms in training neural networks},
  author={Hoang, M. Duc},
  journal={arXiv preprint arXiv:2609.00789},
  year={2026},
  url={https://arxiv.org/abs/2609.00789}
}

@article{hoang2026adaptive,
  title={Adaptive hybrid subspace {Levenberg--Marquardt} algorithm with adequacy monitor for large-scale least squares problems},
  author={Hoang, M. Duc and Lewis, Timothy J.},
  journal={arXiv preprint arXiv:2608.25524},
  year={2026},
  url={https://arxiv.org/abs/2608.25524}
}

@book{saad2003iterative,
  title={Iterative Methods for Sparse Linear Systems},
  author={Saad, Yousef},
  edition={Second},
  publisher={Society for Industrial and Applied Mathematics},
  address={Philadelphia, PA},
  year={2003},
  doi={10.1137/1.9780898718003}
}

@article{druskin1998extended,
  title={Extended {Krylov} subspaces: Approximation of the matrix square root and related functions},
  author={Druskin, Vladimir and Knizhnerman, Leonid},
  journal={SIAM Journal on Matrix Analysis and Applications},
  volume={19},
  number={3},
  pages={755--771},
  year={1998},
  doi={10.1137/S0895479895292400}
}

@article{pearson1901lines,
  title={On lines and planes of closest fit to systems of points in space},
  author={Pearson, Karl},
  journal={The London, Edinburgh, and Dublin Philosophical Magazine and Journal of Science},
  volume={2},
  number={11},
  pages={559--572},
  year={1901},
  doi={10.1080/14786440109462720}
}

@article{hotelling1933analysis,
  title={Analysis of a complex of statistical variables into principal components},
  author={Hotelling, Harold},
  journal={Journal of Educational Psychology},
  volume={24},
  number={6},
  pages={417--441},
  year={1933},
  doi={10.1037/h0071325}
}

@article{zakarin2025accelerating,
  title={Accelerating neural network training along sharp and flat directions},
  author={Zakarin, Daniyar and Singh, Sidak Pal},
  journal={arXiv preprint arXiv:2505.11972},
  year={2025},
  url={https://arxiv.org/abs/2505.11972}
}

@inproceedings{deng2026suspicious,
  title={Suspicious alignment of {SGD}: A fine-grained step size condition analysis},
  author={Deng, Shenyang and Liao, Boyao and Ouyang, Zhuoli and Pang, Tianyu and Song, Minhak and Yang, Yaoqing},
  booktitle={Proceedings of the 37th International Conference on Algorithmic Learning Theory},
  series={Proceedings of Machine Learning Research},
  volume={313},
  pages={1--66},
  year={2026},
  publisher={PMLR},
  url={https://arxiv.org/abs/2601.11789}
}

@article{so2026minibatch,
  title={Mini-batch noise lowers sharpness via dominant-subspace fluctuations},
  author={So, Junho and Shin, Dongwook},
  journal={arXiv preprint arXiv:2607.23012},
  note={HiLD 2026: 4th Workshop on High-dimensional Learning Dynamics},
  year={2026},
  url={https://arxiv.org/abs/2607.23012}
}

@article{benarous2023spectral,
  title={Spectral alignment of stochastic gradient descent for high-dimensional classification tasks},
  author={Ben Arous, G{\'e}rard and Gheissari, Reza and Huang, Jiaoyang and Jagannath, Aukosh},
  journal={arXiv preprint arXiv:2310.03010},
  year={2023},
  url={https://arxiv.org/abs/2310.03010}
}

@article{nicolas2025dome,
  title={{DOME}: Improving signal-to-noise in stochastic gradient descent via sharp-direction subspace filtering},
  author={Nicolas, Julien and Maouche, Mohamed and Ben Mokhtar, Sonia and Coates, Mark},
  journal={arXiv preprint arXiv:2507.03545},
  year={2025},
  url={https://arxiv.org/abs/2507.03545}
}

@inproceedings{paszke2019pytorch,
  title={{PyTorch}: An Imperative Style, High-Performance Deep Learning Library},
  author={Paszke, Adam and Gross, Sam and Massa, Francisco and Lerer, Adam and Bradbury, James and Chanan, Gregory and Killeen, Trevor and Lin, Zeming and Gimelshein, Natalia and Antiga, Luca and Desmaison, Alban and K{\"o}pf, Andreas and Yang, Edward and DeVito, Zach and Raison, Martin and Tejani, Alykhan and Chilamkurthy, Sasank and Steiner, Benoit and Fang, Lu and Bai, Junjie and Chintala, Soumith},
  booktitle={Advances in Neural Information Processing Systems},
  volume={32},
  year={2019}
}

@inproceedings{li2018intrinsic,
  title={Measuring the Intrinsic Dimension of Objective Landscapes},
  author={Li, Chunyuan and Farkhoor, Heerad and Liu, Rosanne and Yosinski, Jason},
  booktitle={International Conference on Learning Representations},
  year={2018}
}

@inproceedings{gressmann2020improving,
  title={Improving Neural Network Training in Low Dimensional Random Bases},
  author={Gressmann, Frithjof and Eaton-Rosen, Zach and Luschi, Carlo},
  booktitle={Advances in Neural Information Processing Systems},
  volume={33},
  year={2020}
}

@article{li2022low,
  title={Low Dimensional Trajectory Hypothesis is True: {DNNs} Can Be Trained in Tiny Subspaces},
  author={Li, Tao and Tan, Lei and Huang, Zhehao and Tao, Qinghua and Liu, Yipeng and Huang, Xiaolin},
  journal={IEEE Transactions on Pattern Analysis and Machine Intelligence},
  volume={45},
  number={3},
  pages={3411--3420},
  year={2023}
}

@inproceedings{vinyals2012krylov,
  title={Krylov Subspace Descent for Deep Learning},
  author={Vinyals, Oriol and Povey, Daniel},
  booktitle={Proceedings of the 15th International Conference on Artificial Intelligence and Statistics},
  series={Proceedings of Machine Learning Research},
  volume={22},
  pages={1261--1268},
  year={2012},
  publisher={PMLR}
}

@inproceedings{gower2019rsn,
  title={{RSN}: Randomized Subspace {Newton}},
  author={Gower, Robert M. and Kovalev, Dmitry and Lieder, Felix and Richt{\'a}rik, Peter},
  booktitle={Advances in Neural Information Processing Systems},
  volume={32},
  year={2019}
}

@inproceedings{arora2022understanding,
  title={Understanding Gradient Descent on the Edge of Stability in Deep Learning},
  author={Arora, Sanjeev and Li, Zhiyuan and Panigrahi, Abhishek},
  booktitle={Proceedings of the 39th International Conference on Machine Learning},
  series={Proceedings of Machine Learning Research},
  volume={162},
  pages={948--1024},
  year={2022},
  publisher={PMLR}
}

@inproceedings{damian2023self,
  title={Self-Stabilization: The Implicit Bias of Gradient Descent at the Edge of Stability},
  author={Damian, Alex and Nichani, Eshaan and Lee, Jason D.},
  booktitle={International Conference on Learning Representations},
  year={2023}
}

@inproceedings{muller2023achieving,
  title={Achieving High Accuracy with {PINNs} via Energy Natural Gradient Descent},
  author={M{\"u}ller, Johannes and Zeinhofer, Marius},
  booktitle={Proceedings of the 40th International Conference on Machine Learning},
  series={Proceedings of Machine Learning Research},
  volume={202},
  pages={25471--25485},
  year={2023},
  publisher={PMLR}
}

@article{wang2023expert,
  title={An Expert's Guide to Training Physics-informed Neural Networks},
  author={Wang, Sifan and Sankaran, Shyam and Wang, Hanwen and Perdikaris, Paris},
  journal={arXiv preprint arXiv:2308.08468},
  year={2023}
}

@article{kiyani2025which,
  title={Optimizing the Optimizer for Physics-Informed Neural Networks and {Kolmogorov--Arnold} Networks},
  author={Kiyani, Elham and Shukla, Khemraj and Urb{\'a}n, Jorge F. and Darbon, J{\'e}r{\^o}me and Karniadakis, George Em},
  journal={arXiv preprint arXiv:2501.16371},
  year={2025}
}

@book{golub2013matrix,
  author    = {Golub, Gene H. and Van Loan, Charles F.},
  title     = {Matrix Computations},
  edition   = {Fourth},
  publisher = {Johns Hopkins University Press},
  address   = {Baltimore, MD},
  year      = {2013}
}

@book{higham2002accuracy,
  author    = {Higham, Nicholas J.},
  title     = {Accuracy and Stability of Numerical Algorithms},
  edition   = {Second},
  publisher = {SIAM},
  address   = {Philadelphia, PA},
  year      = {2002}
}

@article{davis1970rotation,
  author  = {Davis, Chandler and Kahan, W. M.},
  title   = {The Rotation of Eigenvectors by a Perturbation. {III}},
  journal = {SIAM Journal on Numerical Analysis},
  volume  = {7},
  number  = {1},
  pages   = {1--46},
  year    = {1970}
}

@book{stewart1990matrix,
  author    = {Stewart, G. W. and Sun, Ji-guang},
  title     = {Matrix Perturbation Theory},
  publisher = {Academic Press},
  address   = {Boston, MA},
  year      = {1990}
}

@techreport{nielsen1999damping,
  author      = {Nielsen, Hans Bruun},
  title       = {Damping Parameter in {Marquardt's} Method},
  institution = {Informatics and Mathematical Modelling, Technical University of Denmark},
  address     = {Kongens Lyngby, Denmark},
  year        = {1999}
}

@book{madsen2004methods,
  author    = {Madsen, Kaj and Nielsen, Hans Bruun and Tingleff, Ole},
  title     = {Methods for Non-Linear Least Squares Problems},
  edition   = {Second},
  publisher = {Informatics and Mathematical Modelling, Technical University of Denmark},
  address   = {Kongens Lyngby, Denmark},
  year      = {2004}
}

\newpage

\appendix
\section{Experimental Details}
\label{app:details}

\subsection{Experimental Setup}
\label{sec:details}

\paragraph{Architecture.}
All networks are \(\tanh\) multilayer perceptrons with input dimension 2
(space and time, or two spatial coordinates) and a scalar output. The
experiments of Section~\ref{sec:experiments} use the networks of
Section~\ref{sec:setup}: \(2\)--\(30\)--\(30\)--\(30\)--\(1\) with
\(p=1981\) parameters for Poisson-2D, heat-1D, and wave-1D, and four hidden
layers of
width 45 with \(p=6391\) for the scaled Burgers benchmark. The
spurious-alignment study of Appendix~\ref{app:spurious} uses a smaller
\(2\)--\(20\)--\(20\)--\(20\)--\(1\) network with \(p=921\) parameters.

\paragraph{Data.}
Collocation points are sampled uniformly at random from the domain of each
PDE with the budgets of Table~\ref{tab:collocation} and then held fixed for
the whole run; they are never resampled. The test set used for the test loss
is drawn once in the same way, with four times as many points of each type,
and is likewise held fixed.

\paragraph{Optimizer settings.}
Unless stated otherwise, the following learning rates and initial step
lengths were used:
\begin{enumerate}
    \item fixed-step GD/SGD: learning rate \(\eta=0.01\);
    \item backtracking line search: initial trial step length
      \(\alpha_t^{(0)}=1.0\);
    \item Wolfe line search: initial trial step length
      \(\alpha_t^{(0)}=1.0\).
\end{enumerate}
The energy threshold is \(\tau=0.95\) or \(\tau=0.99\) for first-order
experiments and \(\tau=0.999\) or \(\tau=0.9999\) for Gauss--Newton
experiments, as stated with each experiment (Section~\ref{sec:setup}).
\begin{enumerate}
    \item Each experiment was run for \(10{,}000\) iterations, except the
      \(20{,}000\)-iteration runs of Appendix~\ref{app:spurious}.
    \item All experiments were run in double precision.
    \item The line-search Gauss--Newton runs use the fixed damping
      \(\mu_t=10^{-6}\); the Levenberg--Marquardt runs adapt it
      (Section~\ref{sec:setup}).
\end{enumerate}
\paragraph{Software and hardware.}
Our experiments were conducted using PyTorch \citep{paszke2019pytorch}. For
the edge-of-stability diagnostics we followed the setup of
\citet{cohen2021gradient}, using the publicly available
implementation,\footnote{\url{https://github.com/locuslab/edge-of-stability}}
which is also the code base used by \citet{song2025does}. All experiments
were performed on NVIDIA RTX~3090 GPUs.

\section{Spurious Alignment and the Energy Threshold}
\label{app:spurious}

\subsection{Overcoming Spurious Alignment}
\label{OSA}

\begin{figure}[H]
    
    \begin{subfigure}[b]{\linewidth}
        \centering
        \includegraphics[width=1\linewidth]{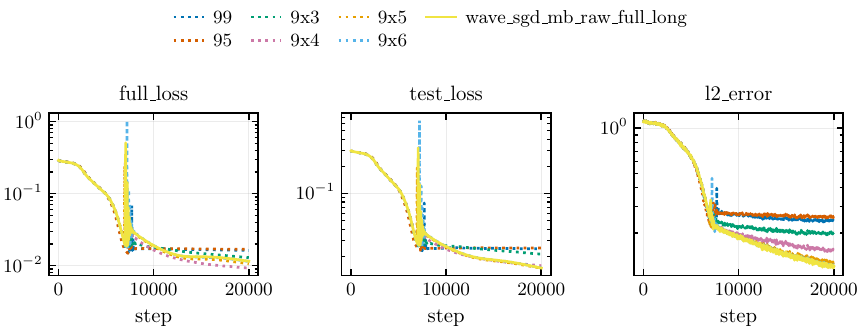}
    \end{subfigure}
    
    \vspace{0.60em}
    
    \begin{subfigure}[b]{\linewidth}
        \centering
        \includegraphics[width=0.49\linewidth]{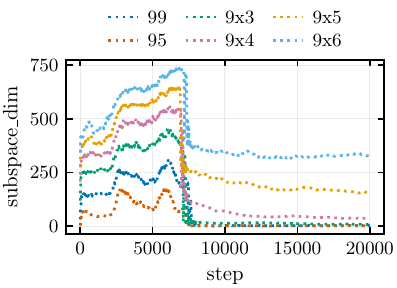}
        \includegraphics[width=0.49\linewidth]{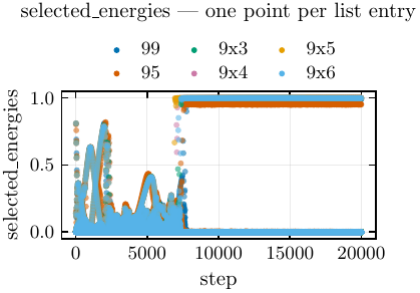}
    \end{subfigure}
    \caption{Spurious alignment in fixed-step DAS-GD on wave-1D. Top:
    training loss, test loss, and relative \(L^2\) error versus iteration
    for DAS-GD with energy thresholds \(\tau\in\{0.95, 0.99, 0.999, 0.9999,
    0.99999, 0.999999\}\) (legend labels ``95'', ``99'', ``9x3'',
    \ldots, ``9x6'') and for the full-space fixed-step run (solid). After
    the edge-of-stability event near iteration 7500, every run with
    \(\tau\le0.9999\) stalls, while \(\tau=0.99999\) and
    \(\tau=0.999999\) continue to track the full-space run. Bottom left:
    DAS dimension \(k_t\); the two thresholds that track the full-space run
    require \(k_t\approx400\)--\(700\) of \(p=921\) directions during the
    main phase of training. Bottom right: gradient-energy scores \(a_i\) of
    the selected eigenvectors; after the collapse a single direction carries
    essentially all of the captured energy.}
    \label{fig:spurious-alignment}
\end{figure}

As reported by \citet{song2025does}, and as seen in our own fixed-step
GD and DAS-GD runs, the spurious-alignment phenomenon persists whenever no
line search is used: once the iterate reaches the edge of stability, the
gradient becomes dominated by the oscillating component along the sharpest
eigendirection, the energy-threshold rule selects that direction alone, and
the DAS-GD run stalls (Figure~\ref{fig:spurious-alignment}). The runs of
this appendix use the smaller \(2\)--\(20\)--\(20\)--\(20\)--\(1\) network
(\(p=921\)) of Appendix~\ref{app:details} and \(20{,}000\) iterations.
With the present criterion, the only remedy is to raise the energy
threshold, \(\tau=0.95, 0.99, 0.999, 0.9999,\ldots\); only at
\(\tau=0.99999\) and \(\tau=0.999999\) does DAS-GD track the full-space
run. The drawback is that such a threshold over-captures during the early
phase of training, where far less energy is needed. This motivates future
work on a threshold schedule, on a per-direction energy cap (no single
direction may contribute more than a prescribed share of the captured
energy), and on smarter selection criteria in general.

\section{Convergence of DAS-Restricted Methods}
\label{app:convergence}

\subsection{Scope}
\label{app:scope}

This appendix establishes convergence guarantees for gradient,
Gauss--Newton, and Newton iterations whose search directions are restricted
to the dominant aligned subspace (DAS). The guarantees are of three
different kinds and are kept separate throughout.
\begin{enumerate}
\item[(i)] \emph{Global convergence to stationarity} for deterministic
iterations equipped with a backtracking (Armijo) line search, a Wolfe line
search, or a fixed step length (Section~\ref{app:global}).
\item[(ii)] \emph{Local convergence} of the DAS-Newton and DAS-Gauss--Newton
(DAS-GN) iterations, analysed as inexact Newton methods whose subspace
changes at every iteration (Section~\ref{app:local}). For DAS-GN the
zero-residual and nonzero-residual regimes are treated separately.
\item[(iii)] The \emph{boundary} of these results
(Section~\ref{app:boundary}). Every statement below concerns deterministic
iterations in which the gradient, the Hessian, and the DAS are computed
from the full training loss. The mini-batch experiments reported in the
main text are not covered by any theorem of this appendix;
Section~\ref{app:boundary} states what a stochastic extension would
require and why it is not immediate; the mini-batch results are empirical and have not been formally proven.
\end{enumerate}
The global analysis follows the framework of \citet{hoang2026adaptive} for
subspace Levenberg--Marquardt methods (see also \citealp{hoang2026subspace});
Section~\ref{app:hoang} records the correspondence result by result and
identifies what is specific to the DAS.

\subsection{Setting and standing assumptions}
\label{app:setting}

Let $\mathcal{L}:\mathbb{R}^p\to\mathbb{R}$ be twice continuously
differentiable, and write $g_t=\nabla\mathcal{L}(\theta_t)$ and
$H_t=\nabla^2\mathcal{L}(\theta_t)$ at the current iterate $\theta_t$. If
$g_t=0$ the iteration stops at a stationary point; all statements below
concern iterations that have not stopped.

\begin{definition}[Dominant aligned subspace]\label{def:das}
Let $H_t=\sum_{i=1}^p\lambda_{t,i}\,v_{t,i}v_{t,i}^\top$ be an
eigendecomposition with orthonormal eigenvectors, and let
$c_{t,i}=\langle v_{t,i},g_t\rangle$, so that
$\sum_{i=1}^pc_{t,i}^2=\|g_t\|^2$. The DAS index set
$\mathcal{A}_t\subseteq\{1,\dots,p\}$ is chosen by one of two rules:
\begin{enumerate}
\item[(a)] (\emph{top-$k$}) $\mathcal{A}_t$ consists of $k$ indices with
the largest values of $c_{t,i}^2$;
\item[(b)] (\emph{energy threshold} $\tau\in(0,1]$) $\mathcal{A}_t$ is the
smallest set of indices, taken in decreasing order of $c_{t,i}^2$, with
$\sum_{i\in\mathcal{A}_t}c_{t,i}^2\ge\tau\|g_t\|^2$;
\end{enumerate}
ties being broken arbitrarily. Set $k_t=|\mathcal{A}_t|$,
$V_t=[v_{t,i}]_{i\in\mathcal{A}_t}\in\mathbb{R}^{p\times k_t}$,
$P_t=V_tV_t^\top$,
$\Lambda_{\mathcal{A}_t}=\operatorname{diag}(\lambda_{t,i})_{i\in\mathcal{A}_t}$,
and
\begin{equation}
\rho_t=\frac{\|P_tg_t\|^2}{\|g_t\|^2}
=\frac{\sum_{i\in\mathcal{A}_t}c_{t,i}^2}{\|g_t\|^2}\in[0,1],
\label{eq:app-rho}
\end{equation}
the captured energy fraction of the DAS.
\end{definition}

Three consequences of Definition~\ref{def:das} are used repeatedly: $V_t$
has orthonormal columns; $H_tV_t=V_t\Lambda_{\mathcal{A}_t}$, hence
$H_tP_t=P_tH_tP_t$ and $(I-P_t)H_tP_t=0$; and
$\|V_t^\top g_t\|=\|P_tg_t\|=\sqrt{\rho_t}\,\|g_t\|$.

\paragraph{Variable subspaces.}
The DAS is recomputed at every iterate from $(H_t,g_t)$ alone. No
assumption relating $\mathcal{A}_t$ to $\mathcal{A}_{t+1}$ is made
anywhere in this appendix: the dimension $k_t$ and the span of $V_t$ may
differ between consecutive iterations without restriction.
Remark~\ref{rem:variable} identifies, for each result, exactly which
properties of $V_t$ its proof uses.

\paragraph{Search directions.}
All DAS-restricted methods considered generate
\begin{equation}
p_t=-V_tM_tV_t^\top g_t\in\operatorname{range}(P_t),
\label{eq:app-direction}
\end{equation}
with $M_t\in\mathbb{R}^{k_t\times k_t}$ symmetric positive definite:
\begin{align}
M_t&=I &&\text{(DAS-GD)},\notag\\
M_t&=G_t^{-1},\qquad G_t:=V_t^\top J_t^\top J_tV_t+\mu_tI,\quad\mu_t>0,
&&\text{(DAS-GN)},\label{eq:app-gn}\\
M_t&=\operatorname{diag}\bigl(1/\max\{|\lambda_{t,i}|,\delta\}\bigr)_{i\in\mathcal{A}_t},\quad\delta>0,
&&\text{(DAS-Newton)}.\notag
\end{align}
Here $\mathcal{L}(\theta)=\tfrac12\|r(\theta)\|^2$ for DAS-GN, with
residual Jacobian $J_t=\partial r(\theta_t)/\partial\theta$ and
residual curvature $S_t:=H_t-J_t^\top J_t=\sum_ir_i(\theta_t)\nabla^2r_i(\theta_t)$.
DAS-GN uses isotropic (Levenberg) damping, as in our implementation.
DAS-Newton is diagonal in the DAS basis because
$V_t^\top H_tV_t=\Lambda_{\mathcal{A}_t}$.

\paragraph{Step-length rules.}
Given a descent direction $p_t$ (i.e., $g_t^\top p_t<0$), the next iterate
is $\theta_{t+1}=\theta_t+\alpha_tp_t$ with $\alpha_t>0$ chosen by one of
the following rules \citep[Chapter~3]{nocedal2006numerical}.
\begin{enumerate}
\item[(LS-A)] \emph{Backtracking.} Given $\alpha_t^{(0)}>0$,
$\beta\in(0,1)$, and $c_1\in(0,1)$, set $\alpha_t=\alpha_t^{(0)}\beta^{j_t}$
with $j_t$ the smallest nonnegative integer for which
\begin{equation}
\mathcal{L}(\theta_t+\alpha_tp_t)\le\mathcal{L}(\theta_t)+c_1\alpha_t\,g_t^\top p_t.
\label{eq:app-armijo}
\end{equation}
\item[(LS-W)] \emph{Wolfe.} $\alpha_t$ satisfies \eqref{eq:app-armijo}
and, for some $c_2\in(c_1,1)$,
\begin{equation}
\nabla\mathcal{L}(\theta_t+\alpha_tp_t)^\top p_t\ge c_2\,g_t^\top p_t.
\label{eq:app-wolfe}
\end{equation}
\item[(FS)] \emph{Fixed step.} $\alpha_t=\alpha$ for a constant $\alpha>0$.
\end{enumerate}

\begin{assumption}[Lower bound]\label{ass:lb}
$\mathcal{L}_{\inf}:=\inf_\theta\mathcal{L}(\theta)>-\infty$.
\end{assumption}

\begin{assumption}[Lipschitz gradient]\label{ass:lip}
$\nabla\mathcal{L}$ is $L$-Lipschitz on $\mathbb{R}^p$.
\end{assumption}

\begin{assumption}[Bounded preconditioners]\label{ass:prec}
There are constants $0<m\le M<\infty$ such that $mI\preceq M_t\preceq MI$
for all $t$.
\end{assumption}

\begin{assumption}[Uniform alignment]\label{ass:align}
There is $\rho_{\min}\in(0,1]$ such that $\rho_t\ge\rho_{\min}$ for all $t$.
\end{assumption}

Assumption~\ref{ass:lip} can be weakened to a convex neighbourhood of the
level set $\{\theta:\mathcal{L}(\theta)\le\mathcal{L}(\theta_0)\}$ that
contains all trial points examined by the step-length rule.
Assumption~\ref{ass:prec} holds for DAS-GD with $m=M=1$; for DAS-GN with
$\mu_t\in[\mu_{\min},\mu_{\max}]$ and $\|J_t\|\le\bar\sigma$, with
$m=1/(\bar\sigma^2+\mu_{\max})$ and $M=1/\mu_{\min}$; and for DAS-Newton
with $m=1/\max\{L,\delta\}$ and $M=1/\delta$, since $|\lambda_{t,i}|\le L$
under Assumption~\ref{ass:lip}. Assumption~\ref{ass:align} is the
deterministic adequacy condition of \citet[Assumption~3]{hoang2026adaptive},
which their algorithm enforces by enriching the subspace; Lemma~\ref{lem:alignment}
shows that for the DAS it holds by construction.

\subsection{Global convergence}
\label{app:global}

\begin{lemma}[Alignment is automatic for the DAS]\label{lem:alignment}
Under rule (a) of Definition~\ref{def:das}, $\rho_t\ge k/p$; under rule
(b), $\rho_t\ge\tau$. In either case Assumption~\ref{ass:align} holds with
$\rho_{\min}=k/p$ or $\rho_{\min}=\tau$, and $P_tg_t\neq0$ whenever
$g_t\neq0$. Neither bound holds for subspaces selected by eigenvalue
magnitude: for the top-$k$ (dominant) eigenspace and for its orthogonal
complement (bulk), $\rho_t$ may equal zero while $g_t\neq0$.
\end{lemma}

\begin{proof}
(a) The $k$ largest of $p$ nonnegative numbers sum to at least $k/p$ times
their total. (b) is immediate. For the last claim take $g_t$ orthogonal to
the top-$k$ eigenvectors (respectively, in their span).
\end{proof}

\begin{lemma}[Descent]\label{lem:descent}
Let $V_t$ have orthonormal columns and let $M_t$ satisfy
Assumption~\ref{ass:prec}. If $P_tg_t\neq0$, then the direction
\eqref{eq:app-direction} satisfies
\begin{equation}
g_t^\top p_t\le-m\|P_tg_t\|^2<0,\qquad
\|p_t\|\le M\|P_tg_t\|,\qquad
\frac{-g_t^\top p_t}{\|g_t\|\,\|p_t\|}\ge\frac{m}{M}\sqrt{\rho_t}.
\label{eq:app-descent}
\end{equation}
\end{lemma}

\begin{proof}
With $\tilde g_t=V_t^\top g_t$ we have $\|\tilde g_t\|=\|P_tg_t\|$,
$g_t^\top p_t=-\tilde g_t^\top M_t\tilde g_t\le-m\|\tilde g_t\|^2$, and
$\|p_t\|=\|V_tM_t\tilde g_t\|=\|M_t\tilde g_t\|\le M\|\tilde g_t\|$.
Dividing gives the third inequality. The first two are
\citet[Lemma~2 and proof of Theorem~1]{hoang2026adaptive} in the present
notation.
\end{proof}

\begin{remark}[DAS condition number]\label{rem:kappa}
For DAS-Newton with $\lambda_{t,i}>0$ on $\mathcal{A}_t$ and the floor
inactive, $M_t=\Lambda_{\mathcal{A}_t}^{-1}$ and
$m/M=1/\kappa_{\mathcal{A}}(H_t)$, where
$\kappa_{\mathcal{A}}(H_t)=\max_{i\in\mathcal{A}_t}\lambda_{t,i}/\min_{i\in\mathcal{A}_t}\lambda_{t,i}$
is the DAS condition number \eqref{eq:das-condition-number} of the main text.
Thus \eqref{eq:app-descent} reads
$\cos\angle(p_t,-g_t)\ge\sqrt{\rho_t}/\kappa_{\mathcal{A}}(H_t)$, and
Assumption~\ref{ass:prec} for DAS-Newton amounts to
$\sup_t\kappa_{\mathcal{A}}(H_t)<\infty$ together with the floor $\delta$.
\end{remark}

\begin{theorem}[Global convergence with a line search]\label{thm:linesearch}
Let Assumptions~\ref{ass:lb}--\ref{ass:prec} hold, let $p_t$ be given by
\eqref{eq:app-direction}, and let $\theta_{t+1}=\theta_t+\alpha_tp_t$.
\begin{enumerate}
\item[(a)] \textup{(Wolfe)} If $\alpha_t$ satisfies (LS-W), then
\[
\mathcal{L}(\theta_{t+1})\le\mathcal{L}(\theta_t)
-\frac{c_1(1-c_2)\,m^2}{L\,M^2}\,\|P_tg_t\|^2.
\]
\item[(b)] \textup{(Backtracking)} If $\alpha_t$ is produced by (LS-A) with
$\alpha_t^{(0)}\ge\alpha^{(0)}>0$, then the backtracking loop terminates
with
\[
\alpha_t\ \ge\ \alpha_{\min}:=\min\Bigl\{\alpha^{(0)},\ \frac{2\beta(1-c_1)\,m}{L\,M^2}\Bigr\}
\qquad\text{and}\qquad
\mathcal{L}(\theta_{t+1})\le\mathcal{L}(\theta_t)-c_1\alpha_{\min}\,m\,\|P_tg_t\|^2 .
\]
\end{enumerate}
Consequently
$\sum_{t\ge0}\|P_tg_t\|^2\le C\bigl(\mathcal{L}(\theta_0)-\mathcal{L}_{\inf}\bigr)$,
with $C=LM^2/(c_1(1-c_2)m^2)$ in case (a) and $C=1/(c_1\alpha_{\min}m)$ in
case (b); in particular $\|P_tg_t\|\to0$. If in addition
Assumption~\ref{ass:align} holds, then $\|g_t\|\to0$ and
\begin{equation}
\min_{0\le t\le T}\|g_t\|^2\ \le\
\frac{C\,\bigl(\mathcal{L}(\theta_0)-\mathcal{L}_{\inf}\bigr)}{\rho_{\min}\,(T+1)}
\qquad\text{for every }T\ge0.
\label{eq:app-rate}
\end{equation}
\end{theorem}

\begin{proof}
By Lemma~\ref{lem:descent}, $p_t$ is a descent direction; hence Wolfe
steps exist \citep[Lemma~3.1]{nocedal2006numerical}, and termination of
backtracking is shown in (b).

(a) Subtracting $g_t^\top p_t$ from both sides of \eqref{eq:app-wolfe}
and applying Assumption~\ref{ass:lip},
\[
(c_2-1)\,g_t^\top p_t\le\bigl(\nabla\mathcal{L}(\theta_{t+1})-g_t\bigr)^\top p_t
\le L\alpha_t\|p_t\|^2,
\]
so $\alpha_t\ge(1-c_2)(-g_t^\top p_t)/(L\|p_t\|^2)$. Substituting into
\eqref{eq:app-armijo} and using \eqref{eq:app-descent},
\[
\mathcal{L}(\theta_{t+1})\le\mathcal{L}(\theta_t)-\frac{c_1(1-c_2)}{L}\,\frac{(g_t^\top p_t)^2}{\|p_t\|^2}
\le\mathcal{L}(\theta_t)-\frac{c_1(1-c_2)}{L}\cdot\frac{m^2\|P_tg_t\|^4}{M^2\|P_tg_t\|^2}.
\]

(b) Assumption~\ref{ass:lip} gives, for every $\alpha>0$,
$\mathcal{L}(\theta_t+\alpha p_t)\le\mathcal{L}(\theta_t)+\alpha g_t^\top p_t+\tfrac{L}{2}\alpha^2\|p_t\|^2$.
Hence \eqref{eq:app-armijo} holds for every
$\alpha\le\bar\alpha_t:=2(1-c_1)(-g_t^\top p_t)/(L\|p_t\|^2)$
\citep[Lemma~3]{hoang2026adaptive}, and $\bar\alpha_t\ge2(1-c_1)m/(LM^2)$
by \eqref{eq:app-descent}. The loop therefore stops at the first trial step
not exceeding $\bar\alpha_t$: either $\alpha_t=\alpha_t^{(0)}$, or the
preceding trial $\alpha_t/\beta$ was rejected, which forces
$\alpha_t/\beta>\bar\alpha_t$. Thus
$\alpha_t\ge\min\{\alpha^{(0)},\beta\bar\alpha_t\}\ge\alpha_{\min}$, and
\eqref{eq:app-armijo} with \eqref{eq:app-descent} gives the stated
decrease.

Summing the per-step decrease over $t=0,\dots,T$ and using
Assumption~\ref{ass:lb} bounds $\sum_{t\le T}\|P_tg_t\|^2$ uniformly in $T$.
Under Assumption~\ref{ass:align}, $\|g_t\|^2\le\|P_tg_t\|^2/\rho_{\min}$,
so $\sum_t\|g_t\|^2<\infty$, and \eqref{eq:app-rate} follows from
$(T+1)\min_{t\le T}\|g_t\|^2\le\sum_{t\le T}\|g_t\|^2$.
\end{proof}

A natural initial trial step for (LS-A) is
$\alpha_t^{(0)}=1/\max_{i\in\mathcal{A}_t}|\lambda_{t,i}|$ for DAS-GD
(capped above when the selected eigenvalues are close to zero) and
$\alpha_t^{(0)}=1$ for DAS-GN and DAS-Newton; both satisfy
$\alpha_t^{(0)}\ge\min\{1,1/L\}$, as Theorem~\ref{thm:linesearch}(b)
requires.

\begin{theorem}[Global convergence with a fixed step]\label{thm:fixedstep}
Let Assumptions~\ref{ass:lb}--\ref{ass:prec} hold, let $p_t$ be given by
\eqref{eq:app-direction}, and let $\theta_{t+1}=\theta_t+\alpha p_t$ with
a constant $\alpha\in\bigl(0,2m/(LM^2)\bigr)$. Then
\[
\mathcal{L}(\theta_{t+1})\le\mathcal{L}(\theta_t)-\alpha\Bigl(m-\frac{LM^2\alpha}{2}\Bigr)\|P_tg_t\|^2,
\qquad
\sum_{t\ge0}\|P_tg_t\|^2\le\frac{\mathcal{L}(\theta_0)-\mathcal{L}_{\inf}}{\alpha\,(m-LM^2\alpha/2)} .
\]
For DAS-GD the admissible range is $\alpha\in(0,2/L)$, and with
$\alpha=1/L$,
\[
\min_{0\le t\le T}\|P_tg_t\|^2\le\frac{2L\,(\mathcal{L}(\theta_0)-\mathcal{L}_{\inf})}{T+1};
\qquad\text{under Assumption~\ref{ass:align},}\quad
\min_{0\le t\le T}\|g_t\|^2\le\frac{2L\,(\mathcal{L}(\theta_0)-\mathcal{L}_{\inf})}{\rho_{\min}\,(T+1)} .
\]
\end{theorem}

\begin{proof}
By Assumption~\ref{ass:lip} and \eqref{eq:app-descent},
$\mathcal{L}(\theta_{t+1})\le\mathcal{L}(\theta_t)+\alpha g_t^\top p_t+\tfrac{L}{2}\alpha^2\|p_t\|^2
\le\mathcal{L}(\theta_t)-\alpha m\|P_tg_t\|^2+\tfrac{L}{2}\alpha^2M^2\|P_tg_t\|^2$.
The remaining claims follow by summation as in
Theorem~\ref{thm:linesearch}.
\end{proof}

\begin{corollary}[Linear rate under a Polyak--\L{}ojasiewicz condition]\label{cor:pl}
In the setting of Theorem~\ref{thm:fixedstep} with DAS-GD, suppose
Assumption~\ref{ass:align} holds and $\mathcal{L}$ satisfies
$\tfrac12\|\nabla\mathcal{L}(\theta)\|^2\ge\mu\,(\mathcal{L}(\theta)-\mathcal{L}^\star)$
for all $\theta$, where $\mathcal{L}^\star=\min\mathcal{L}$
\citep{karimi2016linear}. Then, with $\alpha=1/L$,
\[
\mathcal{L}(\theta_t)-\mathcal{L}^\star\le\Bigl(1-\frac{\rho_{\min}\mu}{L}\Bigr)^t\bigl(\mathcal{L}(\theta_0)-\mathcal{L}^\star\bigr)
\qquad(t\ge0),
\]
to be compared with the factor $1-\mu/L$ for full-space gradient descent
under the same assumptions.
\end{corollary}

\begin{proof}
Theorem~\ref{thm:fixedstep} with $\alpha=1/L$ gives
$\mathcal{L}(\theta_{t+1})-\mathcal{L}^\star\le\mathcal{L}(\theta_t)-\mathcal{L}^\star-\tfrac{1}{2L}\|P_tg_t\|^2$,
and $\|P_tg_t\|^2\ge\rho_{\min}\|g_t\|^2\ge2\rho_{\min}\mu\,(\mathcal{L}(\theta_t)-\mathcal{L}^\star)$.
\end{proof}

\begin{assumption}[Restricted smoothness]\label{ass:restricted}
There is a constant $L_{\mathcal{A}}>0$ such that
\begin{equation}
\bigl\|P_t\,\nabla^2\mathcal{L}(\theta_t+u\alpha p_t)\,P_t\bigr\|_2\le L_{\mathcal{A}}
\label{eq:app-restricted}
\end{equation}
for every $t$, every trial step $\alpha$ examined by the step-length rule,
and every $u\in[0,1]$.
\end{assumption}

Under Assumption~\ref{ass:lip}, \eqref{eq:app-restricted} holds with
$L_{\mathcal{A}}\le L$, since $\|PHP\|\le\|H\|$. At $u=0$ the left-hand
side equals $\|P_tH_tP_t\|=\max_{i\in\mathcal{A}_t}|\lambda_{t,i}|$, the
sharpest \emph{aligned} eigenvalue, whereas $\|H_t\|=\max_i|\lambda_{t,i}|$.

\begin{proposition}[Restricted constants]\label{prop:restricted}
Theorems~\ref{thm:linesearch} and~\ref{thm:fixedstep} and
Corollary~\ref{cor:pl} remain valid, with the same proofs, when
Assumption~\ref{ass:lip} is replaced by Assumption~\ref{ass:restricted} and
$L$ is replaced by $L_{\mathcal{A}}$ in every statement. In particular
DAS-GD with the fixed step $\alpha=1/L_{\mathcal{A}}$ satisfies
$\mathcal{L}(\theta_{t+1})\le\mathcal{L}(\theta_t)-\|P_tg_t\|^2/(2L_{\mathcal{A}})$,
and under the Polyak--\L{}ojasiewicz condition of Corollary~\ref{cor:pl}
the contraction factor becomes $1-\rho_{\min}\mu/L_{\mathcal{A}}$.
\end{proposition}

\begin{proof}
Assumption~\ref{ass:lip} enters the proofs at exactly two places, and in
both the quantity bounded is a quadratic form in $p_t\in\operatorname{range}(P_t)$.
In Theorem~\ref{thm:linesearch}(a),
$(\nabla\mathcal{L}(\theta_{t+1})-g_t)^\top p_t
=\alpha_t\int_0^1p_t^\top\nabla^2\mathcal{L}(\theta_t+u\alpha_tp_t)\,p_t\,du
=\alpha_t\int_0^1p_t^\top P_t\nabla^2\mathcal{L}(\theta_t+u\alpha_tp_t)P_t\,p_t\,du
\le\alpha_tL_{\mathcal{A}}\|p_t\|^2$.
In Theorem~\ref{thm:linesearch}(b) and Theorem~\ref{thm:fixedstep}, for
every trial $\alpha$,
$\mathcal{L}(\theta_t+\alpha p_t)-\mathcal{L}(\theta_t)-\alpha g_t^\top p_t
=\alpha^2\int_0^1(1-u)\,p_t^\top P_t\nabla^2\mathcal{L}(\theta_t+u\alpha p_t)P_t\,p_t\,du
\le\tfrac12L_{\mathcal{A}}\alpha^2\|p_t\|^2$.
Existence of Wolfe steps uses only continuity of $\nabla\mathcal{L}$ and
Assumption~\ref{ass:lb} \citep[Lemma~3.1]{nocedal2006numerical}.
\end{proof}

\begin{remark}[Interpretation; not a theorem]\label{rem:interpretation}
Proposition~\ref{prop:restricted} says that the step-length restriction and
the rate constants for DAS-GD are governed by the curvature along the
aligned directions, not by the largest Hessian eigenvalue. Treating the
Hessian as locally constant along a step, this suggests a stability
threshold $\alpha<2/\max_{i\in\mathcal{A}_t}|\lambda_{t,i}|$ for DAS-GD in
place of $\alpha<2/\lambda_{\max}$ for full-space GD, and, comparing the
contraction factors of Proposition~\ref{prop:restricted} and
Corollary~\ref{cor:pl}, that DAS-GD is at least as fast as full-space GD in
these bounds precisely when
\begin{equation}
\rho_{\min}\ \ge\ L_{\mathcal{A}}/L .
\label{eq:app-das-vs-full}
\end{equation}
The same bookkeeping offers a reading of the observation of
\citet{song2025does} that updates projected onto the dominant eigenspace
make little progress while updates projected onto its complement train
normally: with $\alpha=1/L_{\mathcal{A}}$ the guaranteed decrease per step is
$\|P_tg_t\|^2/(2L_{\mathcal{A}})$; for the dominant subspace
$\|P_tg_t\|^2\approx\|g_t\|^2$ but $L_{\mathcal{A}}=\lambda_1$, while for the
bulk $\|P_tg_t\|^2$ is small but $L_{\mathcal{A}}\le\lambda_{k+1}\ll\lambda_1$.
Neither statement is a theorem about the actual dynamics: the bounds are
lower bounds on guaranteed decrease, and \eqref{eq:app-restricted} is
assumed rather than verified along the trajectory. They are the form in
which the analysis makes the empirical comparison of
Section~\ref{sec:experiments} quantitative.
\end{remark}

\begin{remark}[Which properties of the subspace sequence are used]\label{rem:variable}
The proofs of Theorems~\ref{thm:linesearch} and~\ref{thm:fixedstep} use
$V_t$ only through Lemma~\ref{lem:descent}, which needs orthonormal
columns and nothing else; the eigenvector property of
Definition~\ref{def:das} is not used, and no relation between $V_t$ and
$V_{t+1}$ is used. This is not an oversight. Each proof is a per-iteration
sufficient-decrease argument in which every quantity refers to the current
$(\theta_t,V_t)$; the summation that yields $\sum_t\|P_tg_t\|^2<\infty$
never compares consecutive subspaces; the step-length rules (LS-A) and
(LS-W) carry no state from one iteration to the next; and acceptance is
decided on the full loss, so $\mathcal{L}(\theta_t)$ is monotone
regardless of how $V_t$ changes. Consequently the global results hold for
\emph{any} sequence of orthonormal bases, in particular for a DAS whose
dimension and orientation change at every iteration, and in particular
for the sequence produced by the hybrid construction of
\citet{hoang2026adaptive}, whose global analysis rests on the same three
facts. The eigenvector property is used in exactly two places: to
identify $\|P_tH_tP_t\|$ with $\max_{i\in\mathcal{A}_t}|\lambda_{t,i}|$
after Assumption~\ref{ass:restricted}, and in Lemma~\ref{lem:residual}.
Assumption~\ref{ass:align} is the only requirement linking the subspace to
the gradient, and Lemma~\ref{lem:alignment} discharges it for the DAS.
\end{remark}

\subsection{Local convergence}
\label{app:local}

Throughout this subsection $\theta^\star$ is a local minimizer with
$g(\theta^\star)=0$, $H^\star:=\nabla^2\mathcal{L}(\theta^\star)\succ0$, and
$\nabla^2\mathcal{L}$ continuous in a neighbourhood $U$ of $\theta^\star$.
Unit steps $\theta_{t+1}=\theta_t+p_t$ are taken (the unit step is
eventually accepted by (LS-A) with $\alpha_t^{(0)}=1$ for Newton-type
directions near a nondegenerate minimizer; we do not reprove this standard
fact and simply assume unit steps). Since $H_t\succ0$ near $\theta^\star$,
every $\lambda_{t,i}$ is positive there, and if $\delta<\lambda_{\min}(H^\star)/2$
the floor in the DAS-Newton direction is inactive in a neighbourhood of
$\theta^\star$, so $M_t=\Lambda_{\mathcal{A}_t}^{-1}$.

The local analysis rests on the theory of inexact Newton methods
\citep{dembo1982inexact}: if the steps satisfy
$\|H_tp_t+g_t\|\le\zeta_t\|g_t\|$ with $\zeta_t\le\bar\zeta<1$ for all $t$,
then for $\theta_0$ close enough to $\theta^\star$ the iterates converge to
$\theta^\star$ and, for any $\zeta\in(\bar\zeta,1)$,
$\|H^\star(\theta_{t+1}-\theta^\star)\|\le\zeta\|H^\star(\theta_t-\theta^\star)\|$
for all $t$ (their Theorem~2.3); the convergence is superlinear if and
only if $\|H_tp_t+g_t\|/\|g_t\|\to0$ (their Theorem~3.3). These results
constrain only the residual $H_tp_t+g_t$ at each iterate and make no
assumption on how $p_t$ is generated; in particular they apply verbatim to
a subspace that changes with $t$.

\begin{lemma}[Newton residual of the DAS steps]\label{lem:residual}
\begin{enumerate}
\item[(i)] If $\lambda_{t,i}\neq0$ for $i\in\mathcal{A}_t$ and
$p_t=-V_t\Lambda_{\mathcal{A}_t}^{-1}V_t^\top g_t$ (DAS-Newton, floor
inactive), then
\begin{equation}
H_tp_t+g_t=(I-P_t)\,g_t,\qquad\|H_tp_t+g_t\|=\sqrt{1-\rho_t}\,\|g_t\| .
\label{eq:app-newton-residual}
\end{equation}
\item[(ii)] If $p_t=-V_tG_t^{-1}V_t^\top g_t$ is the DAS-GN step
\eqref{eq:app-gn}, then
\begin{align}
& H_tp_t+g_t=(I-P_t)\,g_t-\mu_tp_t+P_tS_tp_t, \\
& \|H_tp_t+g_t\|\le\zeta_t\|g_t\|, \label{eq:app-gn-residual} \\
& \zeta_t:=\sqrt{1-\rho_t}+\bigl(\mu_t+\|P_tS_tP_t\|\bigr)\|G_t^{-1}\|.
\end{align}
Moreover, since $(I-P_t)g_t\perp p_t$,
\begin{equation}
\|H_tp_t+g_t\|\ \ge\ \sqrt{(1-\rho_t)\|g_t\|^2+\mu_t^2\|p_t\|^2}\;-\;\|S_t\|\,\|p_t\| .
\label{eq:app-gn-residual-lower}
\end{equation}
\end{enumerate}
\end{lemma}

\begin{proof}
By Definition~\ref{def:das}, $H_tV_t=V_t\Lambda_{\mathcal{A}_t}$, hence
$H_tP_t=V_t\Lambda_{\mathcal{A}_t}V_t^\top$ and $(I-P_t)H_tP_t=0$.

(i) $H_tp_t=-V_t\Lambda_{\mathcal{A}_t}\Lambda_{\mathcal{A}_t}^{-1}V_t^\top g_t=-P_tg_t$,
and $\|(I-P_t)g_t\|^2=\|g_t\|^2-\|P_tg_t\|^2$.

(ii) Let $y_t=-G_t^{-1}V_t^\top g_t$, so $p_t=V_ty_t$ and
$(G_t-\mu_tI)y_t=V_t^\top J_t^\top J_tV_ty_t$. Then
$P_tJ_t^\top J_tp_t=V_t(G_t-\mu_tI)y_t=-P_tg_t-\mu_tp_t$, while
$(I-P_t)J_t^\top J_tp_t=(I-P_t)(H_t-S_t)P_tp_t=-(I-P_t)S_tp_t$.
Adding these and $S_tp_t+g_t$ gives
$H_tp_t+g_t=(I-P_t)g_t-\mu_tp_t+S_tp_t-(I-P_t)S_tp_t=(I-P_t)g_t-\mu_tp_t+P_tS_tp_t$.
The upper bound uses $P_tS_tp_t=P_tS_tP_tp_t$ and
$\|p_t\|=\|y_t\|\le\|G_t^{-1}\|\,\|P_tg_t\|\le\|G_t^{-1}\|\,\|g_t\|$.
For the lower bound, $(I-P_t)g_t$ and $\mu_tp_t\in\operatorname{range}(P_t)$
are orthogonal, so $\|(I-P_t)g_t-\mu_tp_t\|^2=(1-\rho_t)\|g_t\|^2+\mu_t^2\|p_t\|^2$,
and $\|P_tS_tp_t\|\le\|S_t\|\|p_t\|$.
\end{proof}

\begin{proposition}[DAS-Newton: local convergence with a variable subspace]\label{prop:newton-local}
Let Assumption~\ref{ass:align} hold and let $\theta_{t+1}=\theta_t+p_t$
with the DAS-Newton direction, $\delta<\lambda_{\min}(H^\star)/2$. Then for
$\theta_0$ sufficiently close to $\theta^\star$ the iterates converge to
$\theta^\star$, and for any $\zeta\in\bigl(\sqrt{1-\rho_{\min}},1\bigr)$,
\[
\|H^\star(\theta_{t+1}-\theta^\star)\|\le\zeta\,\|H^\star(\theta_t-\theta^\star)\|
\qquad\text{for all }t\ge0 .
\]
The convergence is superlinear if and only if $\rho_t\to1$.
\end{proposition}

\begin{proof}
In a neighbourhood of $\theta^\star$ the floor is inactive and
Lemma~\ref{lem:residual}(i) gives $\|H_tp_t+g_t\|=\sqrt{1-\rho_t}\|g_t\|$
with $\sqrt{1-\rho_t}\le\sqrt{1-\rho_{\min}}<1$. Theorems~2.3 and~3.3 of
\citet{dembo1982inexact} apply with $\bar\zeta=\sqrt{1-\rho_{\min}}$, the
neighbourhood being chosen small enough that the iterates, which remain
within a ball around $\theta^\star$ by the contraction, stay where the floor
is inactive.
\end{proof}

\begin{proposition}[DAS-GN: zero-residual and nonzero-residual regimes]\label{prop:gn-local}
Let $\mathcal{L}=\tfrac12\|r\|^2$ with $r$ twice continuously
differentiable near $\theta^\star$, let Assumption~\ref{ass:align} hold,
and let $\theta_{t+1}=\theta_t+p_t$ with the DAS-GN direction
\eqref{eq:app-gn} and damping $\mu_t\in[0,\bar\mu]$.
\begin{enumerate}
\item[(a)] \textup{(Zero residual.)} Suppose $r(\theta^\star)=0$. For every
$\zeta\in\bigl(\sqrt{1-\rho_{\min}},1\bigr)$ there exist $\bar\mu>0$ and
$\epsilon>0$ such that, if $\mu_t\le\bar\mu$ for all $t$ and
$\|\theta_0-\theta^\star\|\le\epsilon$, then $\theta_t\to\theta^\star$ and
$\|H^\star(\theta_{t+1}-\theta^\star)\|\le\zeta\|H^\star(\theta_t-\theta^\star)\|$
for all $t\ge0$. The convergence is superlinear if and only if
$\rho_t\to1$ and $\mu_t\to0$.
\item[(b)] \textup{(Nonzero residual.)} Suppose $r(\theta^\star)\neq0$, and
let $\zeta_t$ be as in \eqref{eq:app-gn-residual}. If
$\sup_t\zeta_t\le\bar\zeta<1$, then for $\theta_0$ sufficiently close to
$\theta^\star$ the iterates converge linearly with any rate
$\zeta\in(\bar\zeta,1)$ in the norm $\|H^\star\,\cdot\,\|$. No superlinear
rate is asserted: since $S(\theta^\star)\neq0$, the term $P_tS_tp_t$ in
\eqref{eq:app-gn-residual} does not in general become negligible relative
to $\|g_t\|$, and the condition $\sup_t\zeta_t<1$ may fail when the
residual curvature restricted to the DAS is large relative to
$\lambda_{\min}(J_t^\top J_t)$. This is the large-residual limitation shared
by all Gauss--Newton and Levenberg--Marquardt methods
\citep[Section~10.3]{nocedal2006numerical}; cf.\
\citet[Remark~2]{hoang2026adaptive}.
\end{enumerate}
\end{proposition}

\begin{proof}
(a) Since $r(\theta^\star)=0$, $S(\theta^\star)=0$, and by continuity
$J^\top J=H-S\to H^\star\succ0$ as $\theta\to\theta^\star$. Fix
$\zeta\in(\sqrt{1-\rho_{\min}},1)$ and pick $\bar\zeta$ with
$\sqrt{1-\rho_{\min}}<\bar\zeta<\zeta$. Choose a ball $U'\subseteq U$ around
$\theta^\star$ and $\bar\mu>0$ such that, for $\theta\in U'$,
$\lambda_{\min}(J^\top J)\ge\tfrac12\lambda_{\min}(H^\star)$,
$\|S\|\le\sigma$, and
$2(\bar\mu+\sigma)/\lambda_{\min}(H^\star)\le\bar\zeta-\sqrt{1-\rho_{\min}}$.
For $\theta_t\in U'$, $\|G_t^{-1}\|\le1/\lambda_{\min}(V_t^\top J_t^\top J_tV_t)\le2/\lambda_{\min}(H^\star)$,
so \eqref{eq:app-gn-residual} gives $\zeta_t\le\bar\zeta$. Theorem~2.3 of
\citet{dembo1982inexact} is proved by induction and uses the residual bound
only at the current iterate; its iterates satisfy
$\|\theta_t-\theta^\star\|\le\kappa(H^\star)\,\|\theta_0-\theta^\star\|$, so
choosing $\epsilon$ with $\kappa(H^\star)\epsilon$ smaller than the radius
of $U'$ keeps every iterate in $U'$, where the bound is available. This
gives convergence and the linear rate. For the superlinear
characterization, Theorem~3.3 of \citet{dembo1982inexact} states that the
convergence is superlinear if and only if $\|H_tp_t+g_t\|/\|g_t\|\to0$.
By \eqref{eq:app-gn-residual} and \eqref{eq:app-gn-residual-lower}, and
since $\|S_t\|\to0$ while $\|p_t\|\le\|G_t^{-1}\|\|g_t\|$ with
$\|G_t^{-1}\|$ bounded, this ratio tends to zero if and only if
$(1-\rho_t)+\mu_t^2\|p_t\|^2/\|g_t\|^2\to0$. Finally
$\|p_t\|\ge\|P_tg_t\|/\|G_t\|\ge\sqrt{\rho_{\min}}\|g_t\|/(\|J_t\|^2+\bar\mu)$
with $\|J_t\|$ bounded near $\theta^\star$, so
$\mu_t^2\|p_t\|^2/\|g_t\|^2\to0$ if and only if $\mu_t\to0$.

(b) Under $\sup_t\zeta_t\le\bar\zeta<1$, Theorem~2.3 of
\citet{dembo1982inexact} applies directly. The remaining statements are
explanations of why its hypothesis may fail and why Theorem~3.3 gives no
superlinear rate; they are not claims of divergence.
\end{proof}

\begin{remark}[Subspace rotation on the quadratic model]\label{rem:gs}
Propositions~\ref{prop:newton-local} and~\ref{prop:gn-local} show that a
DAS step reduces the error only along the current DAS and leaves the
component $(I-P_t)g_t$ untouched; error components outside the current
subspace are reduced only when later subspaces select them. For a
\emph{quadratic} loss with fixed eigenbasis and eigencoordinates $z_i$, the
selection rule guarantees this: the coefficients $c_i=\lambda_iz_i$ of
unselected coordinates are unchanged by a step while those of selected
coordinates shrink, so unselected coordinates rise in the alignment ranking
and are eventually selected. In eigencoordinates, DAS-GD and DAS-Newton on
a quadratic are exactly block coordinate descent and block coordinate
minimization with the greedy Gauss--Southwell rule
\citep{nutini2015coordinate,nutini2022let}. We use this only as intuition
for why a changing subspace is the mechanism of convergence rather than an
obstacle to it; we make no such claim for general nonconvex losses, for
which the guarantees are those of Sections~\ref{app:global}
and~\ref{app:local} only.
\end{remark}

\subsection{Correspondence with \citet{hoang2026adaptive}}
\label{app:hoang}

The global argument of Section~\ref{app:global} is that of
\citet{hoang2026adaptive} with their reduced basis $V_k$ and reduced damped
matrix $B_k$ replaced by $V_t$ and $M_t=B_k^{-1}$ (up to the orthogonal
factor $Z_k$ of their reduced SVD, so that $m=1/M_B$ and $M=1/m_B$).
Specifically: Assumptions~\ref{ass:lb}, \ref{ass:lip} and~\ref{ass:prec}
are their Assumptions~1, 2 and~4; Assumption~\ref{ass:align} is their
Assumption~3 and $\rho_t$ is their $\eta_k^{\mathrm{sub}}$ (their eq.~(2));
Lemma~\ref{lem:descent} is their Lemma~2 together with the bounds in the
proof of their Theorem~1; the step lower bound in
Theorem~\ref{thm:linesearch}(b) is their Lemma~3, and with $\alpha^{(0)}=1$
the constant $\alpha_{\min}$ coincides with their $\min\{1,\beta\bar t\}$,
$\bar t=2(1-\alpha)m_B^2/(M_BL_1)$. They conclude $\liminf_k\|g_k\|=0$ by
contradiction; retaining $\|P_tg_t\|^2$ rather than its lower bound
$\rho_{\min}\|g_k\|^2$ yields the limit and the rate in
\eqref{eq:app-rate}. Theorem~\ref{thm:linesearch}(a),
Theorem~\ref{thm:fixedstep}, Corollary~\ref{cor:pl} and
Proposition~\ref{prop:restricted} have no counterpart there.

Their local analysis (their Lemma~4 and Theorem~2) bounds the exact Newton
residual of a general subspace LM step by four terms: the omitted gradient
$\|(I-P_k)g_k\|$; the curvature non-invariance $\|H_kP_k-P_kH_kP_k\|\|s_k\|$,
where their $H_k=J_k^\top J_k$ is the Gauss--Newton matrix; the
regularization $\|R_k\|\|y_k\|$; and the residual-curvature term
$\|S_k\|\|s_k\|$. Linear convergence requires the sum to stay below one and
superlinear convergence requires it to vanish, which is imposed through
their Assumptions~6 and~7. Lemma~\ref{lem:residual} is the specialization
to the DAS. Two things change. First, the DAS is invariant under the true
Hessian, so the non-invariance term with respect to $J_k^\top J_k$ equals
$\|(I-P_t)S_tP_t\|\|p_t\|$ and is no longer an independent source of error:
their Assumption~6 on the non-invariance is implied by their Assumption~7
on $S$. Second, the bound becomes an identity, which is what allows the
forcing term of the inexact Newton iteration to be read off exactly as
$\sqrt{1-\rho_t}$ for DAS-Newton and to be characterized as in
Proposition~\ref{prop:gn-local}(a) for DAS-GN. The adequacy condition,
which their monitor enforces, is for the DAS the content of
Lemma~\ref{lem:alignment}.

\subsection{Boundary of the results}
\label{app:boundary}

The following limitations are intrinsic to the analysis above and should
be read together with it.

\paragraph{Full-batch iterations only.}
Every result takes $g_t=\nabla\mathcal{L}(\theta_t)$ and
$H_t=\nabla^2\mathcal{L}(\theta_t)$ to be exact, and
Definition~\ref{def:das} selects $\mathcal{A}_t$ from these exact
quantities. The mini-batch experiments of
Section~\ref{sec:mini-batch-optimization} replace $g_t$ by an estimate $\hat g_t$ with
$\mathbb{E}[\hat g_t\mid\theta_t]=g_t$ and, in some variants, select the
subspace from mini-batch information as well. Two separate issues arise,
and we resolve neither here.
\begin{enumerate}
\item[(1)] If $\mathcal{A}_t$, and hence $P_t$, is computed from the
full-batch $(H_t,g_t)$ while only the step uses $\hat g_t$, then $P_t$ is
determined by $\theta_t$ and $\mathbb{E}[P_t\hat g_t\mid\theta_t]=P_tg_t$.
The argument of Theorem~\ref{thm:fixedstep} then goes through in
expectation with a diminishing step length and a bounded-variance condition,
in the manner of \citet{robbins1951stochastic}, with $\|P_tg_t\|^2$ in place
of $\|g_t\|^2$ and an additional variance term of order $\alpha_t^2$ per
iteration. We state this as an expectation, not a theorem.
\item[(2)] If $\mathcal{A}_t$ is selected from $\hat g_t$, then $P_t$ is a
random projector correlated with the estimation error, and in general
$\mathbb{E}[P_t\hat g_t\mid\theta_t]\neq P_tg_t$: the selected coefficients
are precisely those the noise happened to inflate. Lemma~\ref{lem:descent}
then fails in expectation, and none of the arguments above applies without
a separate treatment of this selection bias.
\end{enumerate}
The line-search results additionally assume that the acceptance tests
\eqref{eq:app-armijo}--\eqref{eq:app-wolfe} are evaluated on the exact loss.
The theorems of this appendix are therefore guarantees for the full-batch
variants of DAS-GD, DAS-GN and DAS-Newton, and the mini-batch behaviour
reported in the main text is an empirical finding that they do not
explain.

\paragraph{Guarantees, not speed.}
Theorems~\ref{thm:linesearch} and~\ref{thm:fixedstep} show that restriction
to a DAS recomputed at every iterate cannot destroy convergence and that
the constants depend on the subspace only through $\rho_{\min}$,
$L_{\mathcal{A}}$, and $m/M$. Whether DAS-restricted training is as fast as
full-space training is the empirical question of the main text;
Remark~\ref{rem:interpretation} is interpretation, not proof.

\paragraph{Methods not analysed.}
No claim is made for trust-region or quasi-Newton (BFGS) methods, nor for
adaptive optimizers such as Adam; none of these is used in our experiments.

\paragraph{Exact eigenvectors.}
Definition~\ref{def:das} assumes an exact eigendecomposition, as used in
all of our experiments. Computed
eigenvectors satisfy $H_tV_t=V_t\Lambda_{\mathcal{A}_t}+E_t$ with $\|E_t\|$ at
the level of the eigensolver's backward error. The global results are
unaffected, since they use only orthonormality of $V_t$
(Remark~\ref{rem:variable}); in Lemma~\ref{lem:residual} the residual
acquires an additional term bounded by $\|E_t\|\,\|M_t\|\,\|g_t\|$, which
enters $\zeta_t$ additively and is negligible at the accuracy of a full
eigendecomposition.

\end{document}